\documentclass[10pt,twocolumn,letterpaper]{article}

\usepackage{cvpr}              %

\usepackage{graphicx}
\usepackage{amsmath}
\usepackage{amssymb}
\usepackage{booktabs}
\usepackage{float}
\usepackage[accsupp]{axessibility}
\usepackage{pifont}%
\usepackage{listings}

\usepackage{gensymb}

\usepackage[pagebackref,breaklinks,colorlinks]{hyperref}

\usepackage[capitalize]{cleveref}
\crefname{section}{Sec.}{Secs.}
\Crefname{section}{Section}{Sections}
\Crefname{table}{Table}{Tables}
\crefname{table}{Tab.}{Tabs.}

\def\cvprPaperID{5403} %
\def\confName{CVPR}
\def\confYear{2022}

\newcommand{\cmark}{\ding{51}}%
\newcommand{\xmark}{\ding{55}}%
\newif\ifdrafting
\draftingtrue %
\draftingfalse %
\ifdrafting
    \newcommand{\ds}[1]{{\leavevmode\color[rgb]{1,0,0}[Deqing: #1]}}
    \newcommand{\ks}[1]{{\leavevmode\color[rgb]{0,0,1}[Kyle: #1]}}
    \newcommand{\dk}[1]{{\leavevmode\color[rgb]{1,0,1}[Dilip: #1]}}
    \newcommand{\cih}[1]{{\leavevmode\color[rgb]{0,0.4,0}[Charles: #1]}}
    \newcommand{\lu}[1]{{\leavevmode\color[rgb]{0.5,0.5,1,0}[Lu: #1]}}
\else
    \newcommand{\ds}[1]{}
    \newcommand{\ks}[1]{}
    \newcommand{\dk}[1]{}
    \newcommand{\cih}[1]{}
    \newcommand{\lu}[1]{}
\fi

\newcommand{\ourmethod}{pyramid adversarial training}
\newcommand{\OurMethod}{Pyramid Adversarial Training}

\newcommand{\PixelMethod}{Adversarial Training}

\newcommand{\ignore}[1]{}
\makeatletter
\newcommand{\printfnsymbol}[1]{%
        \textsuperscript{\@fnsymbol{#1}}%
}
\makeatother

\begin{document}

\title{\OurMethod{} Improves ViT Performance}

\author{Charles Herrmann\thanks{Equal contribution, ordered alphabetically.}\;\quad Kyle Sargent\printfnsymbol{1}\quad Lu Jiang\quad Ramin Zabih\quad\\Huiwen Chang\quad Ce Liu\thanks{Currently affiliated with Microsoft Azure AI.}\quad Dilip Krishnan\quad  Deqing Sun\\
\vspace{1.2ex}
{Google Research}
}

\maketitle

\begin{abstract}
Aggressive data augmentation is a key component of the strong generalization capabilities of Vision Transformer (ViT). One such data augmentation technique is adversarial training (AT); however, many prior works\cite{tradeoff_robustness_accuracy:2020, adversarial_machine_learning_at_scale:2016} have shown that this often results in poor clean accuracy. In this work, we present \ourmethod{} (PyramidAT), a simple and effective technique to improve ViT's overall performance. %
We pair it with a ``matched'' Dropout and stochastic depth regularization, which adopts the same Dropout and stochastic depth configuration for the clean and adversarial samples. %
Similar to the improvements on CNNs by AdvProp~\cite{advprop:2020} (not directly applicable to ViT), our \ourmethod{} breaks the trade-off between in-distribution accuracy and out-of-distribution robustness for ViT and related architectures. 
It leads to $1.82\%$ absolute improvement on ImageNet clean accuracy for the ViT-B model when trained only on ImageNet-1K data, while simultaneously boosting performance on $7$ ImageNet robustness metrics, by absolute numbers ranging from $1.76\%$ to $15.68\%$. We set a new state-of-the-art for ImageNet-C (41.42 mCE), ImageNet-R (53.92\%), and ImageNet-Sketch (41.04\%) without extra data, using only the ViT-B/16 backbone and our \ourmethod{}. 
Our code is publicly available at \url{pyramidat.github.io}. 
\end{abstract}

\vspace{-4mm}
\section{Introduction}
\label{sec:introduction}
One fascinating aspect of human intelligence is the ability to generalize from limited experiences to new environments~\cite{lake2017building}. While deep learning has made remarkable progress in emulating or ``surpassing'' humans on classification tasks, deep models have difficulty generalizing to out-of-distribution data~\cite{ood_image}. Convolutional neural networks (CNNs) may fail to classify images with challenging contexts~\cite{imagenet-a:2021}, unusual colors and textures~\cite{imagenet-r:2021, imagenet-sketch:2019, stylized-imagenet:2019} and common or adversarial corruptions~\cite{imagenet-c:2019, explaining_and_harnessing_adversarial_examples:2015}.
To reliably deploy neural networks on diverse tasks in the real world, we must improve their robustness to out-of-distribution data. %

\begin{figure}[t!]
    \centering
	\newcommand{\Figwidth}{\linewidth}
    \includegraphics[width=\Figwidth]{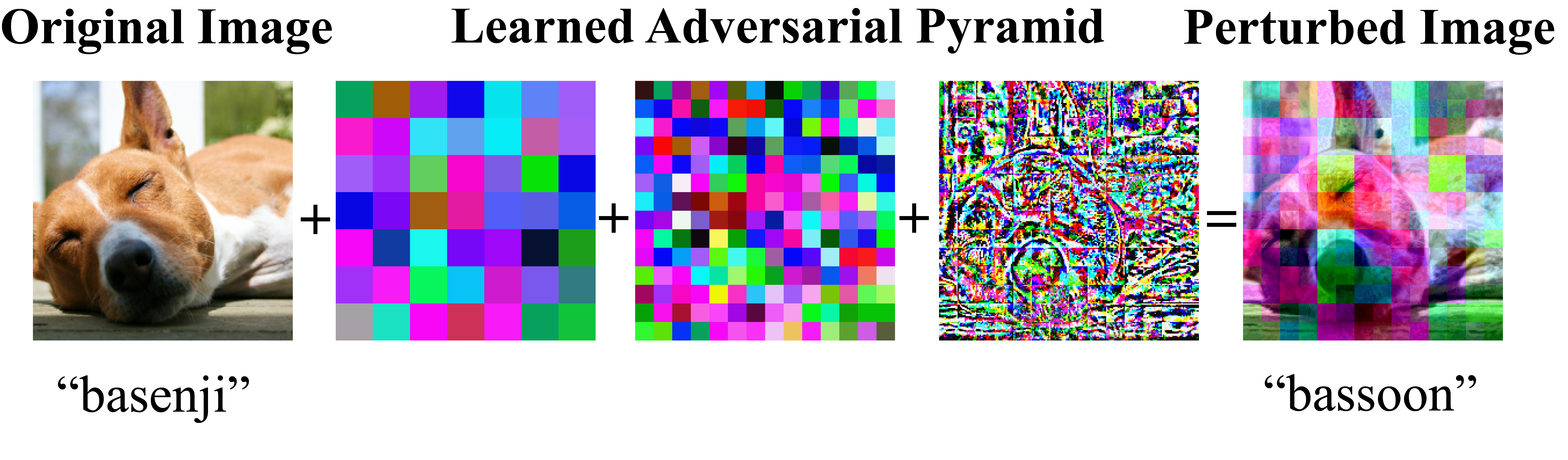}
    \includegraphics[width=\Figwidth]{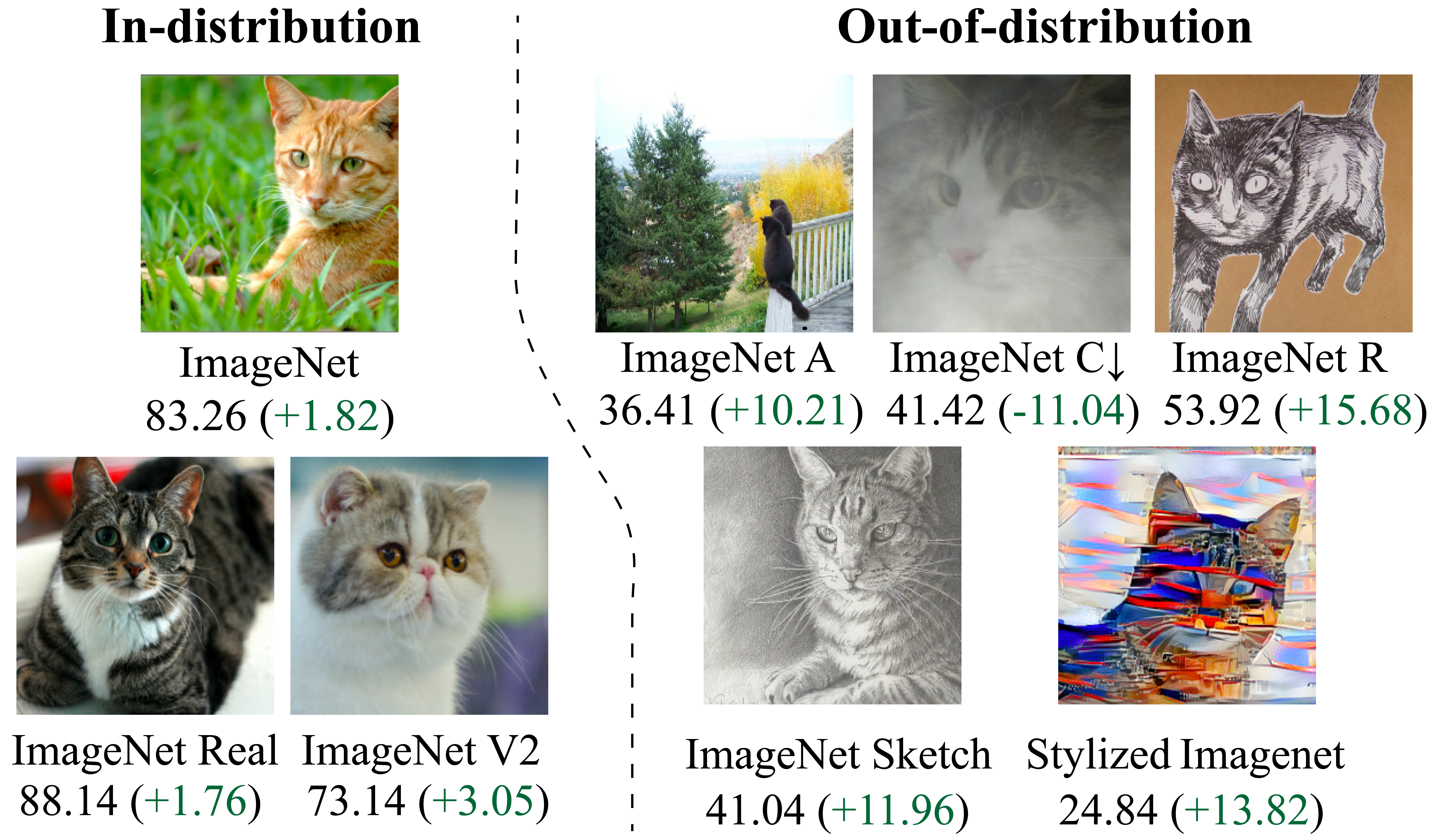}
    \caption{Top: Visualization of our learned multi-scale pyramid perturbations. We show the original image, multiple scales of a perturbation pyramid, and the perturbed image. Bottom: We show thumbnails of in-distribution and out-of-distribution datasets, and the gains from applying our technique on each dataset. (Note that lower is better for ImageNet-C.)}
    \label{fig:multi_scale_perturbation}
\end{figure}

One major line of research focuses on network design. Recently the Vision Transformer (ViT)~\cite{vit:2020} and its variants~\cite{deit:2021, swin:2021, dpt:2021, beit:2021} have advanced the state of the art on a variety of computer vision tasks. In particular, ViT models are more robust than comparable CNN architectures~\cite{towards_robust_vit:2021, vit_are_robust:2021, vit_are_robust:2021, on_the_robustness_of_vision_transformers_to_adversarial_examples:2021}. 
With a weak inductive bias and powerful model capacity, ViT relies heavily on strong data augmentation and regularization to achieve better generalization~\cite{how_to_train_your_vit:2021,deit:2021}. 
To further push this envelope, we explore using adversarial training~\cite{kurakin2016adversarial,theoretically_principled_tradeoff:2019} as a powerful regularizer to improve the performance of ViT models.

Prior work~\cite{robustness_at_odds_with_accuracy:2019} suggests that there exists a performance trade-off between in-distribution generalization and robustness  to adversarial examples. Similar trade-offs have been observed between in-distribution and out-of-distribution generalization~\cite{tradeoff_robustness_accuracy:2020, theoretically_principled_tradeoff:2019}. These trade-offs have primarily been observed in the context of CNNs \cite{unlabeled_improves_robustness:2019, tradeoff_robustness_accuracy:2020}. However, recent work has demonstrated the trade-off can be broken. AdvProp~\cite{advprop:2020} achieves this via adversarial training (abbreviated AT) with a ``split'' variant of Batch Normalization \cite{ioffe2015batch} for EfficientNet \cite{efficientnet:2019}. In our work, we demonstrate that the trade-off can be broken for the newly introduced vision transformer architecture \cite{vit:2020}.

We introduce \emph{\ourmethod{}} (abbreviated as PyramidAT) that trains the model with input images altered at multiple spatial scales, as illustrated in Fig.~\ref{fig:multi_scale_perturbation}; the pyramid attack is designed to make large edits to the image in a structured, controlled manner (similar to augmenting brightness) and small edits to the image in a flexible manner (similar to pixel adversaries). Using these structured, multi-scale adversarial perturbations leads to significant performance gains compared to both baseline and standard pixel-wise adversarial perturbations. Interestingly, we see these gains for both clean (in-distribution) and robust (out-of-distribution) accuracy. We further enhance the pyramid attack with additional regularization techniques: ``matched'' Dropout and stochastic depth. Matched Dropout uses the same Dropout configuration for both the regular and adversarial samples in a mini-batch (hence the word matched). Stochastic depth \cite{stochastic_depth:2016,how_to_train_your_vit:2021} randomly drops layers in the network and provides a further boost when  matched and paired with matched Dropout and multi-scale perturbations. %

Our ablation studies confirm the importance of matched Dropout when used in conjunction with the \ourmethod{}. They also reveal a complicated interplay between adversarial training, the attack being used, and network capacity.  We additionally show that our approach is applicable to datasets of various scales (ImageNet-1K and ImageNet-21K) and for a variety of network architectures such as ViT~\cite{vit:2020}, Discrete ViT \cite{anonymous2022discrete}, ResNet\cite{resnet:2015}, and MLP-Mixer \cite{tolstikhin2021mlp}. Our contributions are summarized below:

\begin{itemize}
    \itemsep0.5mm
    \item To our knowledge, we appear to be the first to demonstrate that adversarial training improves ViT model performance on both ImageNet~\cite{imagenet:2009} and out-of-distribution ImageNet robustness datasets~\cite{imagenet-a:2021,imagenet-c:2019,imagenet-r:2021,imagenet-sketch:2019,stylized-imagenet:2019}.
    \item We demonstrate the importance of matched Dropout and stochastic depth for the adversarial training of ViT.  
    \item We design \ourmethod{}  to generate multi-scale, structured  adversarial perturbations, which achieve significant performance gains over non-adversarial baseline and adversarial training with pixel perturbations.
    \item We establish a new state of the art for ImageNet-C, ImageNet-R, and ImageNet-Sketch without extra data, using only our \ourmethod{} and the standard ViT-B/16 backbone. We further improve our results by incorporating extra ImageNet-21K data. 
    \item We perform numerous ablations which highlight several elements critical to the performance gains.%
\end{itemize}

\section{Related Work}
\label{sec:related}
There exists a large body of work on measuring and improving the robustness of deep learning models, in the context of adversarial examples and generalization to non-adversarial but shifted distributions. We define \emph{out-of-distribution accuracy/robustness} to explicitly refer to performance of a model on non-adversarial distribution shifts, and \emph{adversarial accuracy/robustness} to refer to the special case of robustness on adversarial examples. When the evaluation is performed on a dataset drawn from the same distribution, we call this \emph{clean accuracy}. 

\vspace{-3mm}
\paragraph{Adversarial training and robustness}  The discovery of adversarial examples~\cite{szegedy2013intriguing} has stimulated a large body of literature on adversarial attacks and defenses~\cite{kurakin2016adversarial,towards_deep_learning_models_resistant_to_adversarial_attacks:2018,moosavi2019robustness,qin2019adversarial,athalye2018obfuscated,carlini2017towards,papernot2016distillation,xiao2018spatially}. Of the many proposed defenses, adversarial training \cite{kurakin2016adversarial,towards_deep_learning_models_resistant_to_adversarial_attacks:2018} has emerged as a simple, effective, albeit expensive approach to make networks adversarially robust. Although some work~\cite{robustness_at_odds_with_accuracy:2019, theoretically_principled_tradeoff:2019} has suggested a tradeoff between adversarial and out-of-distribution robustness or clean accuracy, other analysis~\cite{unlabeled_improves_robustness:2019, tradeoff_robustness_accuracy:2020} has suggested simultaneous improvement is achievable. In~\cite{tradeoff_robustness_accuracy:2020, virtual_adversarial:2018}, the authors note improved accuracy on both clean and adversarially perturbed data, though only on smaller datasets such as CIFAR-10 \cite{cifar:2009} and SVHN \cite{svhn:2011}, and only through the use of additional data extending the problem to the semi-supervised setting. Similarly in NLP, adversarial training leads to improvement of clean accuracy for machine translation~\cite{cheng2020advaug,cheng2019robust}.

Most closely related to our work is the technique of \cite{advprop:2020}, which demonstrates the potential of adversarial training to improve both clean accuracy and out-of-distribution robustness. They focus primarily on CNNs and propose split batch norms to separately capture the statistics of clean and adversarially perturbed samples in a mini-batch. At inference time, the batch norms associated with adversarially perturbed samples are discarded, and all data (presumed clean or out-of-distribution) flows through the batch norms associated with clean samples. Their results are demonstrated on EfficientNet\cite{efficientnet:2019} and ResNet \cite{resnet:2015} architectures. However, their approach is not directly applicable to ViT where batch norms do not exist. In our work, we propose novel approaches, and find that properly constructed adversarial training helps clean accuracy and out-of-distribution robustness for ViT models.

\vspace{-3mm}
\paragraph{Robustness of ViT} ViT models have been found to be more adversarially robust than CNNs~~\cite{vit_are_robust:2021, intriguing_properties_of_vit:2021}, and more importantly,  generalize better than CNNs with similar model capacity on ImageNet out-of-distribution robustness benchmarks~\cite{vit_are_robust:2021}. While existing works focus on analyzing the cause of ViT's superior generalizability, this work aims at further improving the strong out-of-distribution robustness of the ViT model. A promising approach to this end is data augmentation; as shown recently ~\cite{how_to_train_your_vit:2021,deit:2021}, ViT benefits from strong data augmentation. However, the data augmentation techniques used in ViT~\cite{how_to_train_your_vit:2021,deit:2021} are optimized for clean accuracy on ImageNet, and knowledge about robustness is still limited. Different from prior works, this paper focuses on improving both the clean accuracy and robustness for ViT. We show that our technique can effectively complement strong ViT augmentation as in~\cite{how_to_train_your_vit:2021}. We additionally verify that our proposed augmentation can benefit three other  architectures: ResNet\cite{resnet:2015}, MLP-Mixer~\cite{tolstikhin2021mlp}, and Discrete ViT~\cite{anonymous2022discrete}.

\vspace{-3mm}
\paragraph{Data augmentation}
Existing data augmentation techniques, although mainly developed for CNNs, transfer reasonably well to ViT models \cite{randaugment:2020,wang2021augmax,hendrycks2019augmix}. Other work has studied larger structured attacks \cite{xiao2018spatially}. Our work is different from prior work in that we utilize adversarial training to augment ViT and tailor our design to the ViT architecture. To our knowledge, we appear to be the first to demonstrate that adversarial training substantially improves ViT performance in both clean and out-of-distribution accuracies.

\section{Approach}
\label{sec:approach}

We work in the supervised learning setting where we are given a training dataset $\mathcal{D}$ consisting of clean images, represented as $x$ and their labels $y$. The loss function considered is a cross-entropy loss $L(\theta, x, y)$, where $\theta$ are the parameters of the ViT model, with weight regularization $f$. The baseline models minimize the following loss:
\begin{equation}
    \mathbb{E}_{(x,y)\sim\mathcal{D}} \Big[L(\theta, \tilde{x}, y) + f(\theta) \Big],
\end{equation}
where $\tilde{x}$ refers to a data-augmented version of the clean sample $x$, and we adopt the standard data augmentations as in \cite{how_to_train_your_vit:2021}, such as RandAug~\cite{randaugment:2020}.

\subsection{\PixelMethod{}}

The overall training objective for adversarial training~\cite{wald1945statistical} is given as follows:
\begin{equation}
    \mathbb{E}_{(x,y)\sim\mathcal{D}}  \Big[\max_{\delta\in\mathcal{P}} L(\theta, \tilde{x} + \delta, y) + f(\theta) \Big],
\label{eq:adv_objective}
\end{equation}
where $\delta$ a per-pixel, per-color-channel additive perturbation, and $\mathcal{P}$ is the perturbation distribution. Note that the adversarial image, $x^{a}$, is given by $\tilde{x}+\delta$, and we use these two interchangeably below.  The perturbation, $\delta$, is computed by optimizing the objective inside the maximization of Eqn.~\ref{eq:adv_objective}. This objective tries to improve the worst-case performance of the network \wrt the perturbation; subsequently, the resulting model has lower clean accuracy.

To remedy this, we can train on both clean and adversarial images~\cite{explaining_and_harnessing_adversarial_examples:2015, kurakin2016adversarial, advprop:2020} using  the following objective:
\begin{equation}
\mathbb{E}_{(x,y)\sim\mathcal{D}}  \Big[L(\theta, \tilde{x}, y) + \lambda  \max_{\delta\in\mathcal{P}} L(\theta, \tilde{x} + \delta, y) + f(\theta) \Big],
\label{eqn:mix_adv}
\end{equation}
This objective uses adversarial images as a form of regularization or data augmentation, to force the network towards certain representations that perform well on out-of-distribution data. These networks exhibit some degree of robustness but still have good clean accuracy. More recently, \cite{advprop:2020} proposes a split batch norm that leads to performance gains for CNNs on both clean and robust ImageNet test datasets. Note that they do not concern themselves with adversarial robustness, and neither do we in this paper.

\subsection{\OurMethod{}}
\noindent Pixel-wise adversarial images are defined\cite{kurakin2016adversarial} as  $x^{a} = x + \delta$ where the perturbation distribution $P$ consists of a clipping function $C_{\mathcal{B}_\epsilon}$  that clips the perturbation at each pixel location to be inside the specified ball ($\mathcal{B}_\epsilon$) for a specified $l_p$-norm \cite{towards_deep_learning_models_resistant_to_adversarial_attacks:2018}, with maximal radius $\epsilon$ for the perturbation. 
\vspace{-3mm}
\paragraph{Motivation}
For pixel-wise adversarial images, increasing the value of $\epsilon$ or the number of steps of the inner loop in Eqn.~\ref{eqn:mix_adv} eventually causes a drop in clean accuracy (Fig~\ref{fig:strength_visualization}). Conceptually, pixel attacks are very flexible and, if given the ability to make large changes (in $L_2$ distance), can destroy the object being classified; training with these images may harm the network. In contrast, augmentations, like brightness, can lead to large $L_2$ distances but will preserve the object because they are structured. Our main motivation is to design an attack which has the best of both worlds: a low-magnitude flexible component and a high-magnitude structured component; this attack can lead to large image differences while still preserving the class identity.

\vspace{-3mm}
\paragraph{Approach}
We propose \ourmethod{} (PyramidAT) which generates adversarial examples by perturbing the input image at multiple scales. This attack is more flexible and yet also more structured, since it consists of multiple scales, but the perturbations are constrained at each scale.

\begin{equation}
    x^{a} = C_{\mathcal{B}_1}\Big( \tilde{x} + \sum_{s\in S} m_s \cdot C_{\mathcal{B}_{\epsilon_s}}(\delta_s)\Big),
    \label{equation:multi_scale}
\end{equation}
where $C_{\mathcal{B}_1}$ is the clipping function that keeps the image within the normal range, $S$ is the set of scales,  $m_s$ is the multiplicative constant for scale $s$, $\delta_s$ is the learned perturbation (with the same shape as $x$). For scale $s$, the weights in $\delta_s$ are shared for pixels in square regions of size $s \times s$ with top left corner $[s\cdot i, s\cdot j]$ for all discrete $i\in [0, \textrm{width}/s]$ and $j\in [0,\textrm{height}/s]$, as shown in Fig.~\ref{fig:multi_scale_perturbation}. Note that, similar to pixel AT, each channel of the image is perturbed independently. More details of the parameter settings are given in Section \ref{sec:experiment} and pseudocode is included in the supplementals.

\vspace{-4mm}
\paragraph{Setting up the attack} %
For both the pixel and pyramid attacks, we use Projected Gradient Descent (PGD) on a random label using multiple steps~\cite{towards_deep_learning_models_resistant_to_adversarial_attacks:2018}. 
With regards to the loss, we observe that for ViT, maximizing the negative loss of the true label leads to aggressive label leaking \cite{kurakin2016adversarial}, \ie, the network learns to predict the adversarial attack and performs better on the perturbed image.  To avoid this, we pick a random label and then minimize the softmax cross-entropy loss towards that random label as described in~\cite{kurakin2016adversarial}.

\subsection{``Matched'' Dropout and Stochastic Depth} 
Standard training for ViT models uses both Dropout~\cite{dropout:2014} and stochastic depth~\cite{stochastic_depth:2016} as regularizers. During adversarial training, we have both the clean samples and adversarial samples in a mini-batch. 
This poses a question about Dropout treatment during adversarial training (either pixel or pyramid).
In the adversarial training literature, the usual strategy is to run the adversarial attack (to generate adversarial samples) without using Dropout or stochastic depth. However, this leads to a training mismatch between the clean and adversarial training paths when both are used in the loss (Eqn.~\ref{eqn:mix_adv}), with the clean samples trained with Dropout and the adversarial samples without Dropout. For each training instance in the mini-batch, the clean branch will only  update subsets of the network while the adversarial branch updates the entire network. The adversarial branch updates are therefore more closely aligned with the model performance during evaluation, thereby leading to an improvement of adversarial accuracy at the expense of clean accuracy. This objective function is given below:

\begin{equation}
\mathbb{E}_{(x,y)\sim\mathcal{D}}  \Big[L( \mathcal{M}(\theta), \tilde{x}, y) + \lambda  \max_{\delta\in\mathcal{P}} L(\theta, x^{a}, y) + f(\theta) \Big],
\end{equation}
where, with a slight abuse of notation, $\mathcal{M}(\theta)$ denotes a network with a random Dropout mask and a stochastic depth configuration. To address the  issue above, we propose adversarial training of ViT with ``matched'' Dropout, \ie, using the same Dropout configuration for both clean and adversarial training branches (as well as for the generation of adversarial samples). 
We show through ablation in Section \ref{sec:experiment} that using the same Dropout configuration leads to the best overall performance for both the clean and robust datasets.

\section{Experiments}
\label{sec:experiment}
In this section, we compare the effectiveness of our proposed PyramidAT to non-AT models, and PixelAT models. 

\subsection{Experimental Setup}

\begin{table*}\centering
\small
\begin{tabular}{l|cc|cccc|ccc}
\toprule
&\multicolumn{2}{c|}{}&\multicolumn{7}{c}{Out of Distribution Robustness Test} \\
 Method & ImageNet & Real & A & C$\downarrow$ & ObjectNet & V2 & Rendition & Sketch & Stylized \\
\midrule
 ViT~\cite{vit:2020} & 72.82 & 78.28 & 8.03 & 74.08 & 17.36 & 58.73 & 27.07 & 17.28 & 6.41 \\
 ViT+CutMix~\cite{yun2019cutmix} & 75.49 & 80.53 & 14.75 & 64.07 & 21.61 & 62.37 & 28.47 & 17.15 & 7.19 \\
 ViT+Mixup~\cite{zhang2017mixup} & 77.75 & 82.93 & 12.15 & 61.76 & 25.65 & 64.76 & 34.90 & 25.97 & 9.84 \\
 RegViT (RandAug)~\cite{how_to_train_your_vit:2021} & 79.92 & 85.14 & 17.48 & 52.46 & 29.30 & 67.49 & 38.24 & 29.08 & 11.02 \\
 +Random Pixel & 79.72 & 84.72 & 17.81 & 52.83 & 28.72 & 67.17 & 39.01 & 29.26 & 12.11 \\
 +Random Pyramid & 80.06 & 85.02 & 19.15 & 52.49 & 29.41 & 67.81 & 39.78 & 30.30 & 11.64 \\
 +PixelAT & 80.42 & 85.78 & 19.15 & 47.68 & 30.11 & 68.78 & 45.39 & 34.40 & 18.28 \\
 +PyramidAT (Ours) & \textbf{81.71} & \textbf{86.82} & \textbf{22.99} & \textbf{44.99} & \textbf{32.92} & \textbf{70.82} & \textbf{47.66} & \textbf{36.77} & \textbf{19.14} \\
\midrule
 RegViT~\cite{how_to_train_your_vit:2021} on 384x384  & 81.44 & 86.38 & 26.20 & 58.19 & 35.59 & 70.09 & 38.15 & 28.13 & 8.36 \\
 +Random Pixel & 81.32 & 86.18 & 25.95 & 58.69 & 34.12 & 69.50 & 37.66 & 28.79 & 9.77 \\
 +Random Pyramid & 81.42 & 86.30 & 27.55 & 57.31 & 34.83 & 70.53 & 38.12 & 29.16 & 9.61 \\
 +PixelAT & 82.24 & 87.35 & 31.23 & 48.56 & 37.41 & 71.67 & 44.07 & 33.68 & 13.52 \\
 +PyramidAT (Ours) & \textbf{83.26} & \textbf{88.14} & \textbf{36.41} & \textbf{47.76} & \textbf{39.79} & \textbf{73.14} & \textbf{46.68} & \textbf{36.73} & \textbf{15.00} \\
\end{tabular}
\vspace{-1mm}
\caption{Main results on ImageNet-1k. All columns reports top-1 accuracy except ImageNet-C which reports mean Corruption Error (mCE) where lower is better. All models are ViT-B/16. The first set of rows show the performance on train and testing on $224 \times 224$ images. The second set of rows shows performance by fine-tuning on $384 \times 384$ images.}
\label{tab:imagenet_1k}
\end{table*}

\begin{table}
\small
\begin{tabular}{l|cc}
\toprule
 Method & Extra Data & IM-C mCE $\downarrow$ \\
 \midrule
 DeepAugment+AugMix \cite{imagenet-r:2021} & \xmark & 53.60 \\
 AdvProp \cite{advprop:2020} & \xmark & 52.90 \\
 Robust ViT \cite{towards_robust_vit:2021} & \xmark & 46.80 \\
 Discrete ViT \cite{anonymous2022discrete} & \xmark & 46.20 \\
 QualNet \cite{Kim_2021_CVPR} & \xmark & 42.50 \\
 Ours (ViT-B/16 + PyramidAT) &\xmark & \textbf{41.42} \\ 
 \midrule
 Discrete ViT \cite{anonymous2022discrete} & \cmark & 38.74 \\
  Ours (ViT-B/16 + PyramidAT) & \cmark & \textbf{36.80}
\end{tabular}
\vspace{-1mm}
\caption{Comparison to state of the art for mean Corruption Error (mCE) on ImageNet-C. Extra data is IM-21k.}
\label{tab:sota_c}
\end{table}

\begin{table}
\small
\begin{tabular}{l|cc}
\toprule
 Method & Extra Data & IM-Rendition \\
 \midrule
 Faces of Robustness \cite{imagenet-r:2021} & \xmark & 46.80 \\
 Robust ViT \cite{towards_robust_vit:2021}& \xmark & 48.70 \\
 Discrete ViT \cite{anonymous2022discrete}& \xmark & 48.82 \\
 Ours (ViT-B/16 + PyramidAT)& \xmark & \textbf{53.92} \\ 
 \midrule
 Discrete ViT \cite{anonymous2022discrete} & \cmark & 55.26 \\
 Ours (ViT-B/16 + PyramidAT) & \cmark& \textbf{57.84} \\ 
\end{tabular}
\caption{Comparison to state of the art for Top-1 on ImageNet-R. Extra data is IM-21k.}
\label{tab:sota_r}
\end{table}

\begin{table}
\small
\begin{tabular}{l|cc}
\toprule
 Method & Extra Data & IM-Sketch \\
 \midrule
ConViT-B\cite{d2021convit} & \xmark &35.70 \\
Swin-B\cite{swin:2021} & \xmark & 32.40 \\
 Robust ViT \cite{towards_robust_vit:2021} & \xmark & 36.00 \\
 Discrete ViT \cite{anonymous2022discrete} & \xmark & 39.10 \\
 Ours (ViT-B/16 + PyramidAT) & \xmark & \textbf{41.04} \\ 
 \midrule
 Discrete ViT \cite{anonymous2022discrete}  & \cmark & 44.72 \\
 Ours (ViT-B/16 + PyramidAT) & \cmark & \textbf{46.03} \\ 
\end{tabular}
\vspace{-1mm}
\caption{Comparison to state of the art for Top-1 on ImageNet-Sketch. Extra data is IM-21k.}
\label{tab:sota_sketch}
\end{table}

\vspace{-1mm}
\paragraph{Models} We focus primarily on ViT-B/16  \cite{vit:2020}, the baseline ViT with a patch size of 16. We also demonstrate our technique on other network architectures, such as ViT-Ti/16, ResNet~\cite{resnet:2015}, MLP-Mixer~\cite{tolstikhin2021mlp}, and the recent  Discrete ViT~\cite{anonymous2022discrete}.

\vspace{-4mm}
\paragraph{Datasets} %
We train models on both ImageNet-1K and ImageNet-21K \cite{ilsvrc:2015, imagenet:2009}. We evaluate in-distribution performance on 2 additional variants: ImageNet-ReaL \cite{imagenet_real:2020}  which relabels the validation set of the original ImageNet in order to correct labeling errors; and ImageNet-V2\cite{recht2019imagenet} which collects another version of ImageNet's evaluation set. We evaluate out-of-distribution robustness on 6 datasets: ImageNet-A~\cite{imagenet-a:2021} which places the ImageNet objects in unusual contexts or orientations;  ImageNet-C~\cite{imagenet-c:2019}  which applies a series of corruptions (e.g. motion blur, snow, JPEG, etc.); ImageNet-Rendition~\cite{imagenet-r:2021} which contains abstract or rendered versions of the object; ObjectNet~\cite{barbu2019objectnet} which consists of a large real-world set from a large number of different backgrounds, rotations, and imaging view points; ImageNet-Sketch~\cite{imagenet-sketch:2019} which contains artistic sketches of the objects; and Stylized ImageNet~\cite{stylized-imagenet:2019} which processes the ImageNet images with style transfer from an unrelated source image. For brevity, we may abbreviate ImageNet as IM. For all datasets except IM-C, we report top-1 accuracy (where higher is better). For IM-C, we report the standard ``Mean corruption error'' (mCE) (where lower is better).

\vspace{-4mm}
\paragraph{Implementation details} \hspace{-1mm} Following \cite{how_to_train_your_vit:2021},  we use a batch size of 4096, a cosine decay learning rate schedule (0.001 magnitude) with linear warmup for the first 10k steps, \cite{stochastic_gradient_descent_with_warm_restarts:2017}, and the AdamW optimizer \cite{adam:2014} in all our experiments. Augmentations and regularizations include RandAug\cite{randaugment:2020} with the default setting of $(2,15)$, Dropout \cite{dropout:2014} at probability $0.1$, and stochastic depth \cite{stochastic_depth:2016} at probability $0.1$. We train with Scenic\cite{dehghani2021scenic}, a Jax\cite{jax2018github} library, on DragonFish TPUs.

To generate the pixel adversarial attack, we follow~\cite{advprop:2020}. We use a learning rate of $1/255$, $\epsilon=4/255$, and attack for $5$ steps with SGD. We use PGD \cite{towards_deep_learning_models_resistant_to_adversarial_attacks:2018} to generate the adversarial perturbations. We also experiment with using more recent optimizers~\cite{zhuang2020adabelief} to construct the attacks (results are provided in the supplementals). For pyramid attacks, we find using stronger perturbations at coarser scales is more effective than equal perturbation strengths across all scales. By default, we use a 3-level pyramid and use perturbation scale factors $S = [32, 16, 1]$ (a scale of $1$ means that each pixel has one learned parameter, a scale of $16$ means that each $[16,16]$ patch has one learned parameter) with multiplicative terms of $m_s = [20, 10, 1]$ (see Eqn.~\ref{equation:multi_scale}). We use a clipping value of $\epsilon_s = 6/255$ for all levels of the pyramid. 

\subsection{Experimental Results on ViT-B/16}

\begin{table*}\centering\small
\begin{tabular}{l|cc|cccc|ccc}
\toprule
&\multicolumn{2}{c|}{}&\multicolumn{7}{c}{Out of Distribution Robustness Test} \\
 Method & ImageNet & Real & A & C$\downarrow$ & ObjectNet & V2 & Rendition & Sketch & Stylized \\
\midrule
 ViT-B/16 (512x512) & 84.42 & 88.74 & 55.77 & 46.69 & 46.68 & 74.88 & 51.26 & 36.79 & 13.44 \\
 +PixelAT & 84.82 & 89.10 & 57.39 & 43.31 & 47.53 & 75.42 & 53.35 & 39.07 & 17.66 \\
 +PyramidAT (Ours) & \textbf{85.35} & \textbf{89.43} & \textbf{62.44} & \textbf{40.85} & \textbf{49.39} & \textbf{76.39} & \textbf{56.15} & \textbf{43.95} & \textbf{19.84} \\
\end{tabular}
\caption{
Main results from pre-training on ImageNet-21K, fine-tuning on ImageNet-1K. We pre-train with the adversarial technique mentioned (pixel or pyramid), but fine-tune on clean data only.
}
\label{tab:imagenet_21k_ft}
\end{table*}

\vspace{-1mm}
\paragraph{ImageNet-1K} Table \ref{tab:imagenet_1k} shows results on ImageNet-1K and robustness datasets for ViT-B/16 models without adversarial training, with pixel adversarial attacks and with pyramid adversarial attacks. Both adversarial training attacks use matched Dropout and stochastic depth, and optimize the random target loss. The pyramid attack provides consistent improvements, on both clean and robustness accuracies, over the baseline and pixel adversaries. In Table~\ref{tab:imagenet_1k}, we also compare against CutMix~\cite{yun2019cutmix} augmentation. We find that CutMix improves performance over the ViT baseline but cannot improve performance when combined with RandAug. Similar to~\cite{patch_gaussian:2019}, we find that CutOut~\cite{cutout:2017} does not boost performance on ImageNet for our models.

The robustness gains of our technique are preserved through fine-tuning on clean data at higher resolution (384x384), as shown in the second set of rows of Table~\ref{tab:imagenet_1k}.  Further, adversarial perturbations are consistently better than random perturbations on either pre-training or fine-tuning, for both pixel and pyramid models.

\ignore{
\cih{this belongs in the supplemental (maybe ablation?)}
We experiment with fine-tuning with adversarial examples as well, but this leads to equal or worse performance. Roughly speaking, this matches the standard intuition regarding adversarial training, which suggests that a robust representations must be robust at every level (including the initial feature extraction) which is mostly unchanged during fine-tuning.
}

\vspace{-3mm}
\paragraph{State of the art} Our model trained on IM-1K sets a new overall state of the art for IM-C~\cite{imagenet-c:2019}, IM-Rendition~\cite{imagenet-r:2021}, and IM-Sketch~\cite{imagenet-sketch:2019}, as shown in Tables~\ref{tab:sota_c},~\ref{tab:sota_r}, and~\ref{tab:sota_sketch}. While we compare all our models under a unified framework in our main experiments, we select the optimal pre-processing, fine-tuning, and Dropout setting for the given dataset when comparing against the state-of-the-art. We also compare against~\cite{anonymous2022discrete} on IM-21K and find that our results still compare favorably.

\vspace{-3mm}
\paragraph{ImageNet-21K} In table \ref{tab:imagenet_21k_ft}, we show that our technique maintains gains over the baseline Reg-ViT and pixel-wise attack on the larger dataset IM-21K. 
Following~\cite{how_to_train_your_vit:2021}, we pre-train on IM-21K and fine-tune on IM-1K at a higher resolution (in our case, 512x512). We apply adversarial training during the pre-training stage only.

\begin{table*}[!htbp]\centering\small
\begin{tabular}{l|cc|cccc|ccc}
\toprule
&\multicolumn{2}{c|}{}&\multicolumn{7}{c}{Out of Distribution Robustness Test} \\
 Method & ImageNet & Real & A & C$\downarrow$ & ObjectNet & V2 & Rendition & Sketch & Stylized \\
\midrule
 ResNet-50\cite{resnet:2015} (our run) & 76.70 & 83.11 & 4.49 & 74.90 & 26.47 & 64.31 & 36.24 & 23.44 & 6.41 \\
 +PixelAT & 77.37 & 84.11 & 6.03 & 66.88 & 27.80 & 65.59 & 41.75 & 27.04 & 8.13 \\
 +PyramidAT & \textbf{77.48} & \textbf{84.22} & \textbf{6.24} & \textbf{66.77} & \textbf{27.91} & \textbf{65.96} & \textbf{43.32} & \textbf{28.55} & \textbf{8.83} \\
\midrule
 MLP-Mixer~\cite{tolstikhin2021mlp} (our run) & 78.27 & 83.64 & 10.84 & 58.50 & 25.90 & 64.97 & 38.51 & 29.00 & 10.08 \\
 +PixelAT & 77.17 & 82.99 & 9.93 & 57.68 & 24.75 & 64.03 & 44.43 & 33.68 & \textbf{15.31} \\
 +PyramidAT & \textbf{79.29} & \textbf{84.78} & \textbf{12.97} & \textbf{52.88} & \textbf{28.60} & \textbf{66.56} & \textbf{45.34} & \textbf{34.79} & 14.77 \\
\midrule
 Discrete ViT~\cite{anonymous2022discrete} (our run) & 79.88 & 84.98 & 18.12 & 49.43 & 29.95 & 68.13 & 41.70 & 31.13 & 15.08 \\
 +PixelAT & 80.08 & 85.37 & 16.88 & 48.93 & \textbf{30.98} & 68.63 & \textbf{48.00} & \textbf{37.42} & \textbf{22.34} \\
 +PyramidAT & \textbf{80.43} & \textbf{85.67} & \textbf{19.55} & \textbf{47.30} & 30.28 & \textbf{69.04} & 46.72 & 37.21 & 19.14 \\
\end{tabular}
\vspace{-1mm}
\caption{
Pyramid adversarial training improves the performance of ResNet, MLP-Mixer, and Discrete ViT.
On MLPMixer, pixel attacks degrade clean performance but improve robustness, similar to the traditionally observed effect of adversarial training.
}
\label{tab:backbones}
\end{table*}

\subsection{Ablations}

\vspace{-1mm}
\paragraph{ImageNet-1k on other backbones}
We explore the effects of adversarial training on three other backbones: ResNet~\cite{resnet:2015}, Discrete ViT~\cite{anonymous2022discrete}, and MLP-Mixer~\cite{tolstikhin2021mlp}. As shown in Table~\ref{tab:backbones}, we find slightly different results. For ResNet, we use the split BN from~\cite{advprop:2020} and show improved performance from  PyramidAT. Other ResNet variants (-101, -200) show the same trend and are included in the supplementals.  For Discrete ViT, we show that AT with both pixel and pyramid leads to general improvements, though the gain from pyramid over pixel is less consistent than with ViT-B/16. For MLP-Mixer, we observe decreases in clean accuracy but gains in the robustness datasets for PixelAT, similar to what has traditionally been observed from AT on ConvNets. However, with PyramidAT, we observe improvements for all evaluation datasets.

\vspace{-3mm}
\paragraph{Matched Dropout and Stochastic Depth}
We study the impact of handling Dropout and stochastic depth for the clean and adversarial update  in Table~\ref{tab:dropout_ablation}. We find that applying matched Dropout for the clean and adversarial update is crucial for achieving simultaneous gains in clean and robust performance. When we eliminate Dropout in the adversarial update (``without Dropout'' rows in~\ref{tab:dropout_ablation}), we observe significant decreases in performance on clean, IM-ReaL, and IM-A; and increases in performance on IM-Sketch and IM-Stylized. This result appears similar to the usual trade-off suggested in~\cite{tradeoff_robustness_accuracy:2020,theoretically_principled_tradeoff:2019}. By contrast, carefully handling Dropout and stochastic depth can lead to performance gains in both clean and out-of-distribution datasets.

\begin{table*}\centering
\small
\begin{tabular}{l|cc|cccc|ccc}
\toprule
&\multicolumn{2}{c|}{}&\multicolumn{7}{c}{Out of Distribution Robustness Test} \\
 Method & ImageNet & Real & A & C$\downarrow$ & ObjectNet & V2 & Rendition & Sketch & Stylized \\
\midrule
 PixelAT with matched Dropout & \textbf{80.42} & \textbf{85.78} & \textbf{19.15} & \textbf{47.68} & \textbf{30.11} & \textbf{68.78} & 45.39 & 34.40 & 18.28 \\
 PixelAT without Dropout & 79.35 & 84.67 & 15.27 & 51.45 & 29.46 & 67.01 & \textbf{47.83} & \textbf{35.77} & \textbf{18.75} \\
\midrule
 PyramidAT with matched Dropout & \textbf{81.71} & \textbf{86.82} & \textbf{22.99} & \textbf{44.99} & \textbf{32.92} & \textbf{70.82} & 47.66 & 36.77 & 19.14 \\
 PyramidAT without Dropout & 79.43 & 85.13 & 14.13 & 54.70 & 29.67 & 67.40 & \textbf{52.34} & \textbf{40.25} & \textbf{22.34} \\
\end{tabular}
\vspace{-1mm}
\caption{Matched Dropout leads to better performance on in-distribution datasets than AT without Dropout.}
\label{tab:dropout_ablation}
\end{table*}

\vspace{-3mm}
\paragraph{Pyramid attack setup} In Table~\ref{tab:pyramid_ablation}, we ablate the pyramid attacks. Pyramid attacks are consistently better than pixel or patch attacks, while the 3-level pyramid attack tends to have the best overall performance. Note that a 2-level pyramid attack consists  of both the pixel and patch attacks. Please refer to the supplementals for comparison on all the metrics.

\begin{table}\centering\small
\begin{tabular}{l|ccccc}
\toprule
 Method & IM & A & C$\downarrow$ & Rend. & Sketch \\
\midrule
 Pixel & 80.42 & 19.15 & 47.68 & 45.39 & 34.40 \\
 Patch & 81.20 & 21.33 & 50.30 & 42.87 & 33.75 \\
 2-level Pyramid & 81.65 & 22.79 & 45.27 & 47.00 & 36.71 \\
 3-level Pyramid & \textbf{81.71} & 22.99 & \textbf{44.99} & 47.66 & 36.77 \\
 4-level Pyramid & 81.66 & \textbf{23.21} & 45.29 & \textbf{47.68} & \textbf{37.41} \\
\end{tabular}
\vspace{-1mm}
\caption{
Pyramid structure ablation. This shows the effect of the layers of the pyramid. Adding coarser layers with larger magnitudes typically improves performance. Patch attack is a 1-level pyramid with shared parameters across a patch of size $16 \times 16$.
}
\label{tab:pyramid_ablation}
\end{table}

\vspace{-3mm}
\paragraph{Network capacity and random augmentation}
We test the effect of network capacity on adversarial training and, consistent with existing literature~\cite{adversarial_machine_learning_at_scale:2016,towards_deep_learning_models_resistant_to_adversarial_attacks:2018}, find that large capacity is critical to effectively utilizing PixelAT. Specifically, low-capacity networks, like ViT-Ti/16, which already struggle to represent the dataset, can be made worse through PixelAT. Table~\ref{tab:tiny} shows that PixelAT  hurts in-distribution performance of the RandAugment 0.4 model but improves out-of-distribution performance.\ignore{, leading to decreases of: 3\% in clean, 2\% in Real, and 15\% in IM-A. This model also exhibits improvements in robustness, with gains to out-of-distributions of: 19\% to Rendition, 21\% to Sketch, and 37\% to Stylized.} Unlike prior work, we note that this effect depends on both the network capacity and the random augmentation applied to the dataset.

Table~\ref{tab:tiny} shows that a low-capacity network can benefit from adversarial training if the random augmentation is of a small magnitude. Standard training with RandAugment~\cite{randaugment:2020} magnitude of 0.4 (abbreviated as RAm=0.4) provides a better clean accuracy than standard training with RAm=0.1; however, PixelAT with the weaker augmentation, RAm=0.1, performs better than either standard training or PixelAT at RAm=0.4. This suggests that the augmentation should be tuned for adversarial training and not fixed based on standard training. %

Table~\ref{tab:tiny} also shows that PyramidAT acts differently than PixelAT and can provide in-distribution gains despite being used with stronger augmentation. For these models, we find that for the robustness datasets, PixelAT tends to marginally outperform PyramidAT.

\begin{table}\centering\small
\begin{tabular}{l|ccccc}
\toprule
 Method & IM & A & C$\downarrow$ & Rend & Sketch \\
\midrule
 Ti/16 RAm=0.1 & 63.58 & 4.80 & 79.23 & 23.66 & 12.54 \\
 +PixelAT & 64.66 & 4.39 & 74.54 & \textbf{32.52} & \textbf{17.65} \\
 +PyramidAT & \textbf{65.49} & \textbf{5.16} & \textbf{74.30} & 29.18 & 16.55 \\
\midrule
 Ti/16 RAm=0.4 & 64.27 & 4.69 & 78.10 & 24.99 & 13.47 \\
 +PixelAT & 62.78 & 4.05 & 77.67 & \textbf{29.75} & \textbf{16.35} \\
 +PyramidAT & \textbf{65.61} & \textbf{4.80} & \textbf{74.72} & 28.89 & 16.14 \\
\end{tabular}
\vspace{-1mm}
\caption{
Results on Ti/16 with lower random augmentation. RAm is the RandAugment\cite{randaugment:2020} magnitude  -- larger means stronger augmentation; both have RandAugment number of  transforms $=1$.  The strength of the random augmentation affects whether PixelAT improves clean accuracy; in contrast, PyramidAT provides consistent gains over the baseline.
}
\label{tab:tiny}
\end{table}

\begin{figure}[t!]
    \centering
	\newcommand{\Figwidth}{\linewidth}
    \includegraphics[width=\Figwidth]{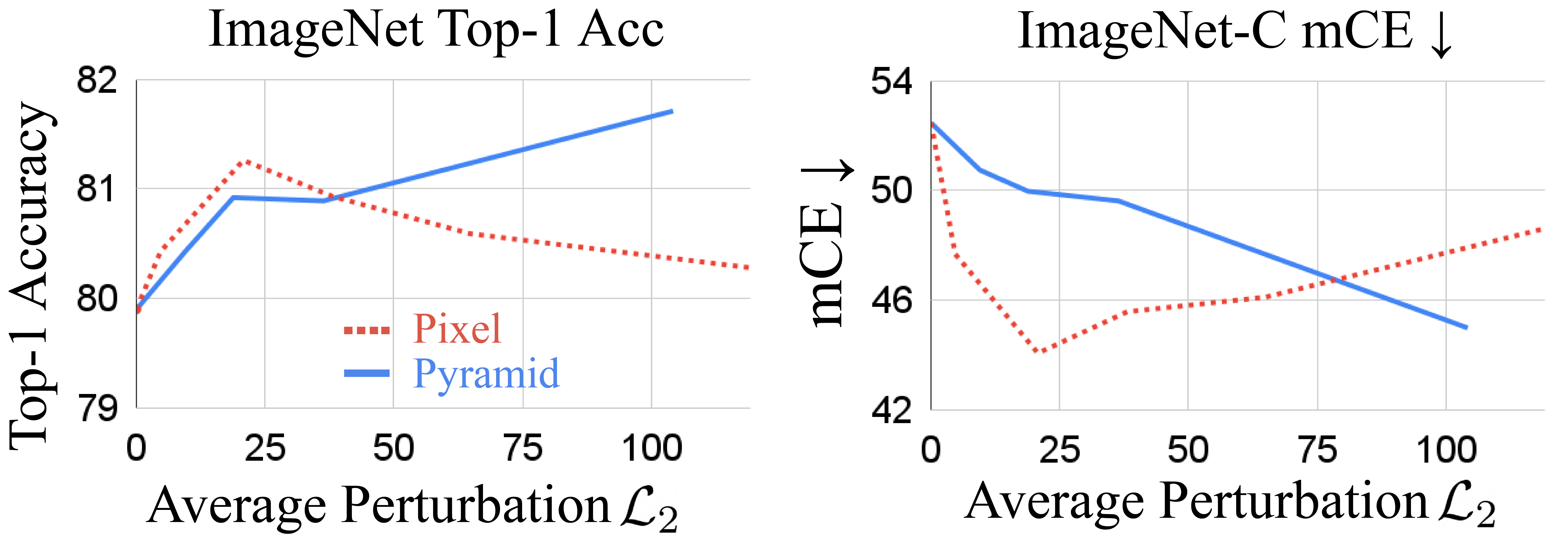}
    \caption{Performance on clean and robust data as a function of perturbation size. Pyramid performance increases as perturbation size is increased, while pixel performance with large perturbation size is poor}
    \label{fig:strength_visualization}
\end{figure}

\vspace{-3mm}
\paragraph{Attack strength} Pixel attacks are much smaller in $L_2$ norm than pyramid attacks. 
We check that simply scaling up the PixelAT cannot achieve the same performance as PyramidAT in Figure~\ref{fig:strength_visualization}. For both ImageNet and ImageNet-C, we show the effect of raising the pixel and pyramid attack strength.  While the best PyramidAT performance is achieved at high $L_2$ perturbation norm, the PixelAT  performance degrades beyond a certain norm. 

\subsection{Analysis and Discussions}

\vspace{-1mm}
\paragraph{Qualitative results}

\begin{figure}
\setlength{\tabcolsep}{.166em}
\begin{tabular}{cccc}
\includegraphics[width=0.24\columnwidth]{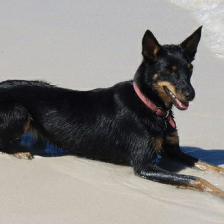} &
\includegraphics[width=0.24\columnwidth]{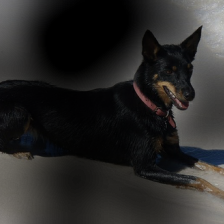} &
\includegraphics[width=0.24\columnwidth]{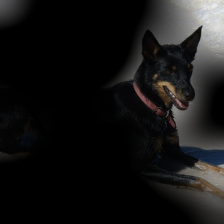} &
\includegraphics[width=0.24\columnwidth]{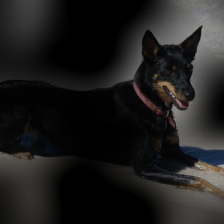} \\
\includegraphics[width=0.24\columnwidth]{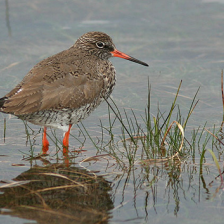} &
\includegraphics[width=0.24\columnwidth]{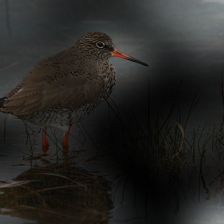} &
\includegraphics[width=0.24\columnwidth]{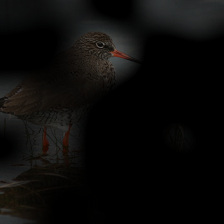} &
\includegraphics[width=0.24\columnwidth]{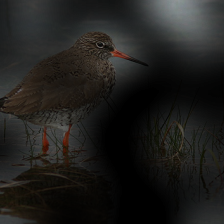} \\
\includegraphics[width=0.24\columnwidth]{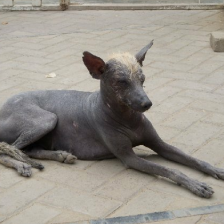} &\includegraphics[width=0.24\columnwidth]{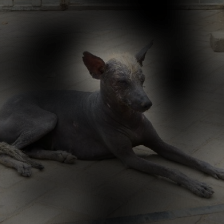} &
\includegraphics[width=0.24\columnwidth]{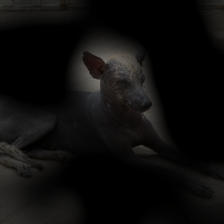} &
\includegraphics[width=0.24\columnwidth]{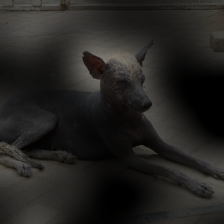} \\
Original & Baseline & PixelAT & PyramidAT \\
\end{tabular}
\vspace{-2mm}
\caption{Visualizations of the attention for different models. PixelAT focuses aggressively on the perceived object. However, if the object is not identified correctly, this focus can be detrimental, as shown above where large parts of the object are discarded. PyramidAT uses a more global perspective and considers context.}
\label{fig:attn_viz_examples}
\end{figure}

\begin{figure}
\setlength{\tabcolsep}{.166em}
\begin{tabular}{ccc}
\includegraphics[width=0.33\columnwidth]{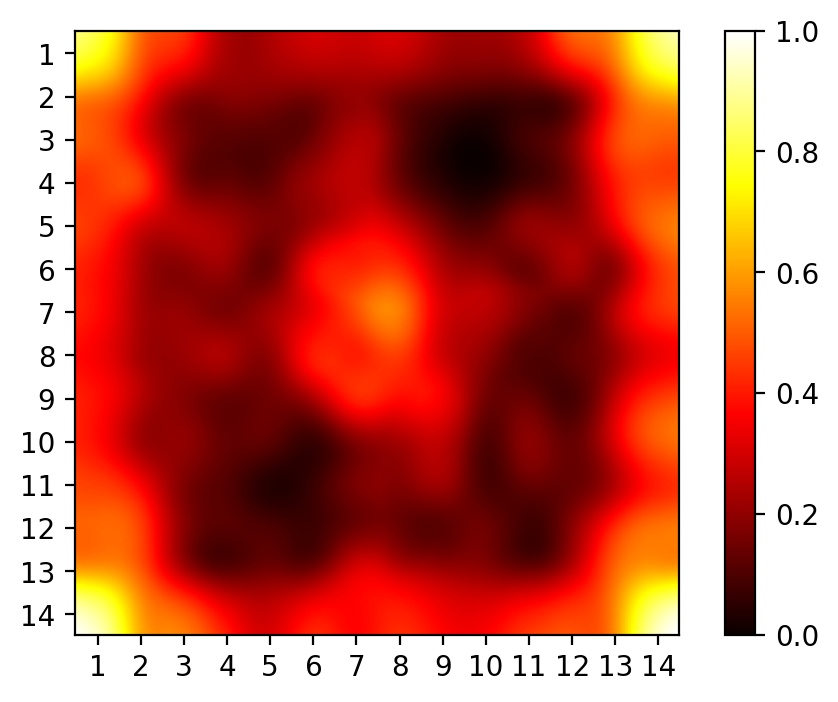} &
\includegraphics[width=0.33\columnwidth]{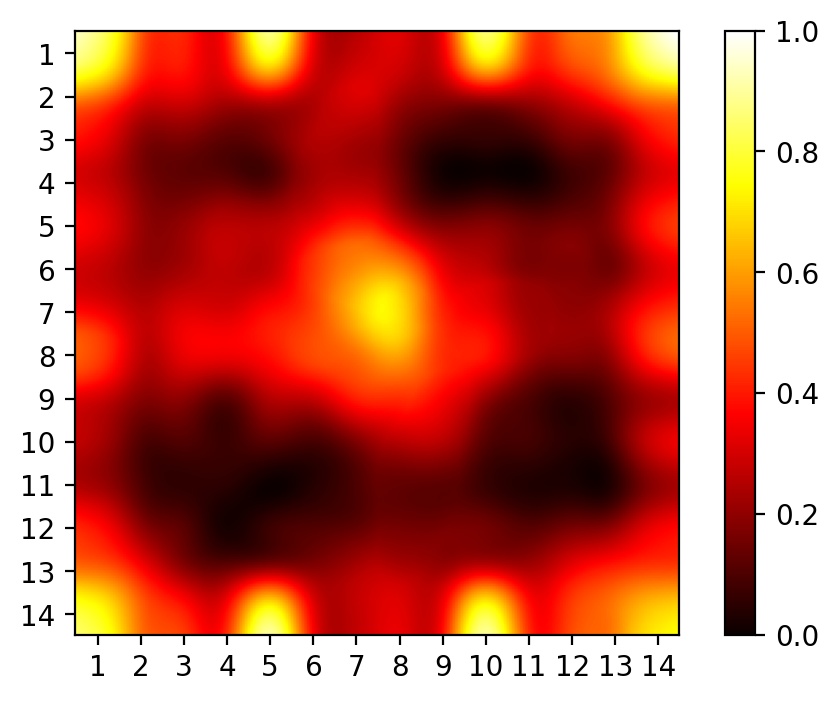} &
\includegraphics[width=0.33\columnwidth]{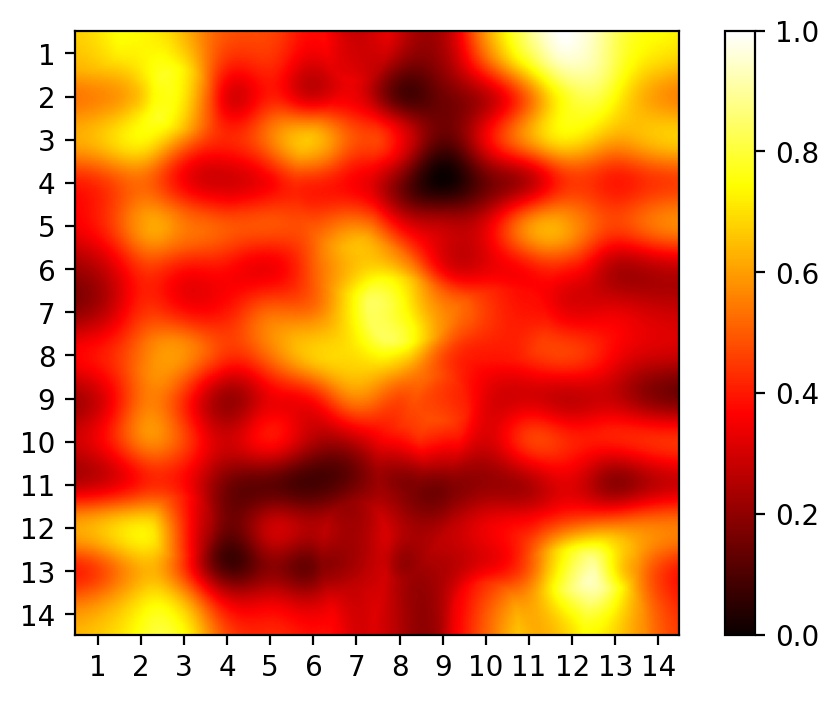} \\
 Baseline & PixelAT & PyramidAT \\
\end{tabular}
\vspace{-2mm}
\caption{Averaged attentions on ImageNet-A: PyramidAT models attend to more of the image than the baseline or PixelAT.}
\label{fig:attn_viz_averages}
\end{figure}

Following~\cite{vit:2020}, we  visualize the learned pixel embeddings (filters) of models trained normally, with pixel adversaries, and with pyramid adversaries in Fig.~\ref{fig:attn_viz_examples}. We observe that the PixelAT model tends to tightly ``snap'' its attention to the perceived object, disregarding the majority of the background. While this may appear to be a desirable behavior, this kind of focusing can be suboptimal for the in-distribution datasets (where the background can provide valuable context) and prone to errors for out-of-distribution datasets. Specifically, the PixelAT model may under-estimate the size or shape of the object and focus on a part of the object and not the whole. This can be problematic for fine-grained classification when the difference between two classes comes down to something as small as the stripes or subtle shape cues (tiger shark vs great white); or texture and context (green mamba vs vine snake). Figure~\ref{fig:attn_viz_averages} shows the heat maps for the average attention on images in the evaluation set of ImageNet-A. We observe that PyramidAT tends to more evenly spread its attention across the entire image than both the baseline and PixelAT.

\begin{figure}
\setlength{\tabcolsep}{.166em}
\begin{tabular}{cccc}
\includegraphics[width=0.24\columnwidth]{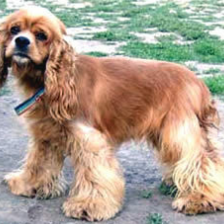} &
\includegraphics[width=0.24\columnwidth]{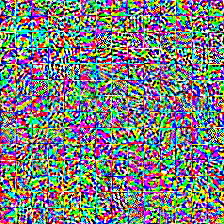} &
\includegraphics[width=0.24\columnwidth]{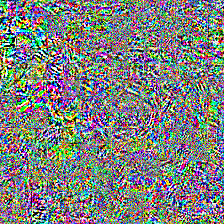} &
\includegraphics[width=0.24\columnwidth]{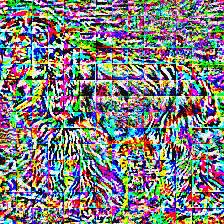} \\
Image & Baseline  & PixelAT & PyramidAT  \\
\end{tabular}
\vspace{-2mm}
\caption{Visualizations of attacks: a pixel attack on a baseline ViT; a pixel attack on a PixelAT ViT; and the pixel level of a pyramid attack on a PyramidAT ViT. 
The pixel attack on the baseline exhibits low amounts of structure and can perturb the label with small changes. The pixel level on the PyramidAT model makes larger changes to the structure; this suggests that the representation is robust to semi-random noise and focuses primarily on structures.}
\label{fig:pixel_attacks_viz}
\end{figure}

Figure~\ref{fig:pixel_attacks_viz} demonstrates the difference in representation between the baseline, PixelAT, and PyramidAT models. The pixel attacks on the baseline and PixelAT have a small amount of structure but appear to consist of mostly texture-level noise. In contrast, the pixel level of the PyramidAT  shows structures from the original image: the legs and back of the dog. This suggests that the representation for the PyramidAT model focuses on shape and is less sensitive to texture than the baseline model.

\begin{figure}
    \centering
    \includegraphics[width=\linewidth]{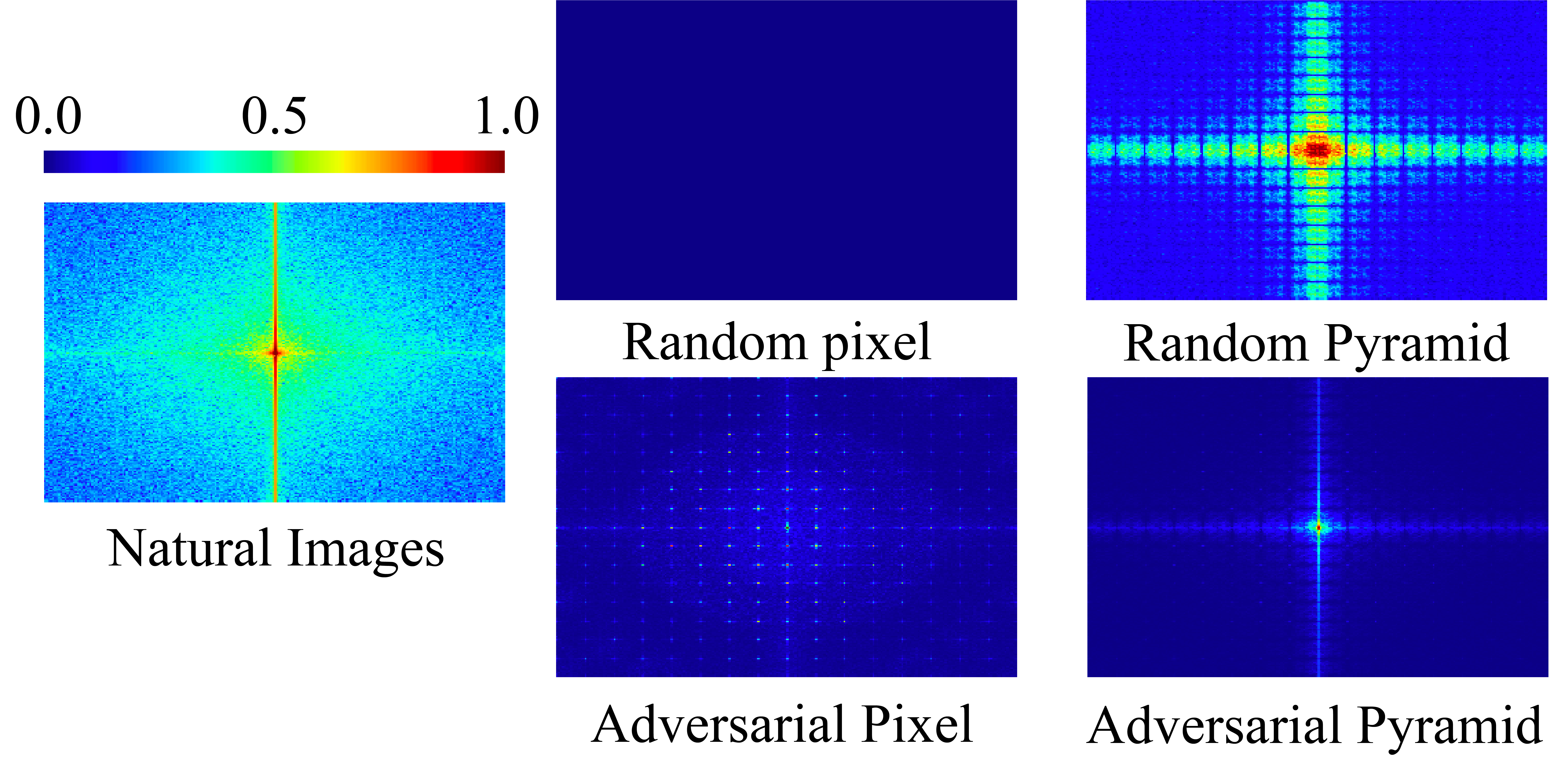}
    \caption{Heatmaps of fourier spectrum for various perturbations.}
    \label{fig:fourier_viz}
\end{figure}

\begin{figure}[t!]
    \centering
	\newcommand{\Figwidth}{\linewidth}
    \includegraphics[width=\Figwidth]{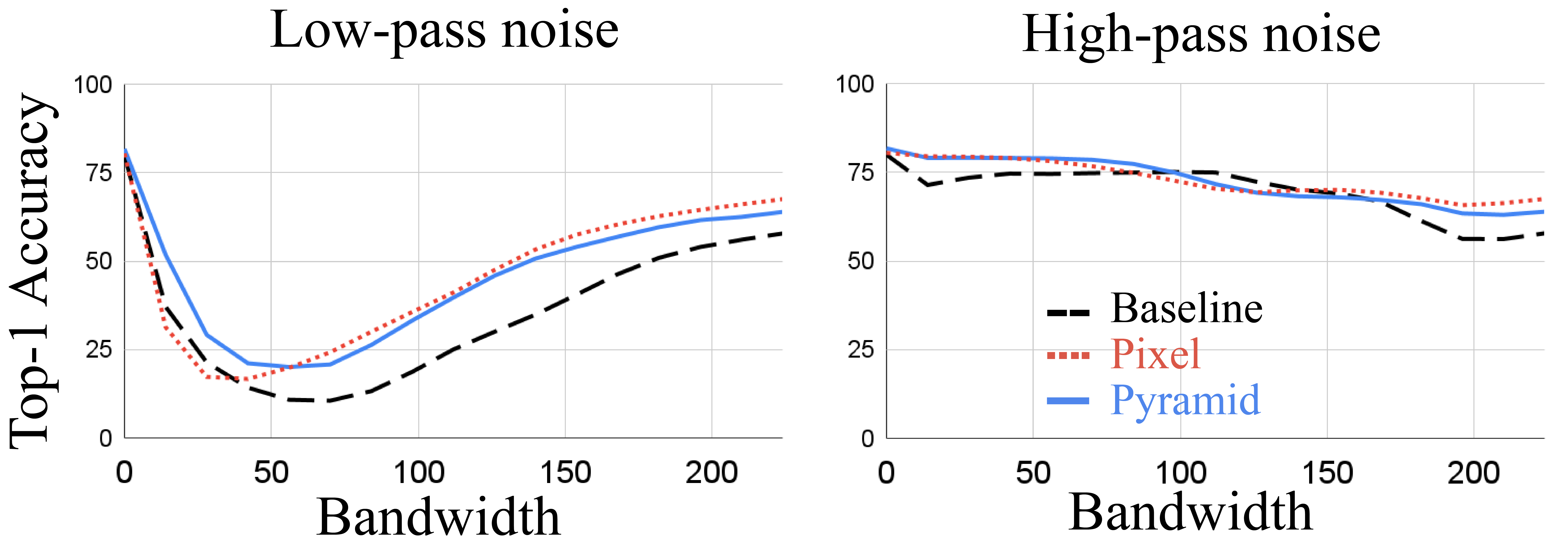}
    \caption{Model performance when inputs are corrupted with low-pass/high-pass filtered noise. The $L_2$ norm of the filtered noise is held constant as the bandwidth is increased. 
    }
    \label{fig:fourier_noise}
\end{figure}

\vspace{-3mm}
\paragraph{Analysis of attacks}
Inspired by~\cite{fourier_perspective:2019}, we analyze the \ourmethod~from a frequency perspective. For this analysis, all visualizations and graphs are averaged over  the entire ImageNet validation set. Figure~\ref{fig:fourier_viz} shows a Fourier heat map of random and adversarial versions of the pixel and pyramid attacks. While random pixel noise is evenly concentrated over all frequencies, adversarial pixel attack tends to concentrate in the lower frequencies. Random pyramid shows a bias towards low frequency as well, a trend which is amplified in the adversarial pyramid. To further explore this, we replicate an analysis from~\cite{fourier_perspective:2019}, where low-pass- and high-pass-filtered random noise is added to test data to perturb a classifier. Figure~\ref{fig:fourier_noise} gives the result for our baseline, pixel, and pyramid adversarially trained models. While pixel and pyramid models are generally more robust than the baseline, the pyramid model is more robust than the pixel  model to low-frequency perturbations.

\vspace{-3mm}
\paragraph{Limitations} 
The cost of our technique is increased training time. A $k$-step PGD attack requires $k$ forward and backward passes for each step of training. Note that this limitation holds for any adversarial training and the inference time is the same. Without adversarial training, more training time does not improve the baseline ViT-B/16.

\section{Conclusion}
\label{sec:conclusion}

We have introduced \ourmethod{}, a simple and effective data augmentation technique that 
substantially improves the performance of ViT and MLP-Mixer architectures on in-distribution and a number of out-of-distribution ImageNet datasets.

{
\small
\bibliographystyle{ieee_fullname}
\bibliography{ref}
}

\appendix
\onecolumn

Thanks for viewing the supplementary material, in which we provide a detailed explanation of the pyramid structure in Section~\ref{sup:pyramid}, detailed experiments for different backbones in  Section~\ref{sup:backbones},  more ablation study in Section~\ref{sup:ablations}, additional analysis in Section~\ref{sup:analysis}, visualizations in Section~\ref{sup:viz}, and finally a discussion on the effect of optimizers in Section~\ref{sup:optimizers}.

\section{Pyramid Attack Details}
\label{sup:pyramid}
In this section, we provide a conceptual description and pseudocode for the pyramid attack.

\noindent \paragraph{Description} Scale $s$ determines the size of the patch that an individual perturbation parameter will be applied to; e.g. for $s\!=\!16$, we learn and add a single adv parameter to each non-overlapping patch of size 16x16. The application is equivalent to a nearest neighbor resize on the 14x14 adv tensor to the image size of 224x224 and then addition. The scales $s$ and multipliers $m_s$ used by PyramidAT are hyperparameters.

\noindent\paragraph{Code} We provide a minimal implementation of our technique in Fig.~\ref{fig:pseudocode}.

\begin{figure}[!htbp]
\begin{lstlisting}{python}
import jax
import jax.numpy as jnp

H=224,lr=1/255, M=[20,10,1], S=[32,16,1], BOUNDS=[0, 1], n_steps=5
def get_attacked_image(model, loss_fn, image):
  def get_perturbed_image(delta):
    return image + sum(
      M[i]*jax.image.resize(delta[i], (H,H,3), 'nearest')
      for i in delta)
  def get_perturbed_loss(delta):
    return loss_fn(model(get_perturbed_image(delta)))
  delta = {i: jnp.zeros((H/s, H/s, 3)) for (i,s) in enumerate(S)}
  for _ in range(n_steps):
    delta_grad = jax.grad(get_perturbed_loss)(delta)
    delta = {i: delta[i]+lr*jnp.sign(delta_grad[i]) for i in delta}
  perturbed_image = get_perturbed_image(delta)
  return jnp.clip(perturbed_image, BOUNDS[0], BOUNDS[1])
\end{lstlisting}
\caption{Implementation of our technique in JAX.}
\label{fig:pseudocode}
\end{figure}

\section{Discussing Backbones}
\label{sup:backbones}
\subsection{ResNets}
Table \ref{tab:backbones_resnet} shows results for multiple variants of ResNet\cite{resnet:2015}: ResNet-50, ResNet-101, and ResNet-200. As the capacity of the network increases, we observe larger gains from both PixelAT and PyramidAT. PyramidAT performs the best on all evaluation sets.

For these runs, we follow the training protocol and network details set up in \cite{advprop:2020}. We use the proposed split BN and standard ResNet training: 90 epochs, cosine learning rate at 0.1 with linear warmup for 7 epochs, minimal augmentations  (left right flip and Inception crop). For network optimization, we use SGD with momentum and for the adversarial steps, we use SGD.

\begin{table*}[!htbp]\centering\small
\begin{tabular}{l|cc|cccc|ccc}
\toprule
&\multicolumn{2}{c|}{}&\multicolumn{7}{c}{Out of Distribution Robustness Test} \\
 Method & ImageNet & Real & A & C$\downarrow$ & ObjectNet & V2 & Rendition & Sketch & Stylized \\
\midrule
 ResNet-50 & 76.70 & 83.11 & 4.49 & 74.90 & 26.47 & 64.31 & 36.24 & 23.44 & 6.41 \\
 +PixelAT & 77.37 & 84.11 & 6.03 & 66.88 & 27.80 & 65.59 & 41.75 & 27.04 & 8.13 \\
 +PyramidAT & \textbf{77.48} & \textbf{84.22} & \textbf{6.24} & \textbf{66.77} & \textbf{27.91} & \textbf{65.96} & \textbf{43.32} & \textbf{28.55} & \textbf{8.83} \\
\midrule
 ResNet-101 & 78.44 & 84.29 & 6.16 & 69.75 & 29.31 & 66.54 & 38.74 & 26.15 & 7.19 \\
 +PixelAT & 79.57 & 85.73 & 9.55 & 59.92 & 31.00 & 67.95 & 44.63 & 29.83 & 10.23 \\
 +PyramidAT & \textbf{79.69} & \textbf{85.82} & \textbf{9.96} & \textbf{59.15} & \textbf{31.57} & \textbf{68.12} & \textbf{46.50} & \textbf{31.74} & \textbf{10.94} \\
\midrule
 ResNet-200 & 79.47 & 85.25 & 8.65 & 65.58 & 31.86 & 67.36 & 40.55 & 28.21 & 7.81 \\
 +PixelAT & 80.51 & 86.34 & 12.93 & 56.99 & 33.84 & 69.73 & 46.32 & 32.74 & 9.14 \\
 +PyramidAT & \textbf{80.92} & \textbf{86.59} & \textbf{14.17} & \textbf{55.72} & \textbf{34.04} & \textbf{70.08} & \textbf{48.05} & \textbf{34.54} & \textbf{11.48} \\
\end{tabular}
\caption{
For all variants of ResNet, PyramidAT leads to improvements. Note that this is with the standard training of ResNet.
}
\label{tab:backbones_resnet}
\end{table*}

\subsection{ViT Tiny/16}

ViT Ti/16 has the same overall structure and design as ViT B/16 (the primary model used in our main paper) but is significantly smaller, at 5.8 million parameters (as opposed to the 86 million parameters of B/16). More specifically, Ti/16 has a width of 192 (instead of 768), MLP size of 768 (instead of 3072), and 3 heads (instead of 12). In total, this decrease in parameters and model size leads to a substantial decrease in capacity. We experiment with ViT Ti/16 primarily in order to understand the impact of this decreased capacity on our adversarial training methods.

We start with an exploration of the impact of the random augmentation's strength on the overall performance of the model. Table \ref{tab:tiny_baselines} shows the performance of Ti/16 models with different RandAugment parameters (the two parameters are in order the number of transforms applied and the magnitude of the transforms); this table suggests that the network's lower capacity benefits from weaker random augmentation. Specifically, the best RandAugment parameters for the majority of the evaluation datasets is (1,0.8), which is considerably lower than the RandAugment parameters tuned for B/16 (2, 15). 

\begin{table*}[!htbp]\centering\small
\begin{tabular}{l|cc|cccc|ccc}
\toprule
&\multicolumn{2}{c|}{}&\multicolumn{7}{c}{Out of Distribution Robustness Test} \\
 Method & ImageNet & Real & A & C$\downarrow$ & ObjectNet & V2 & Rendition & Sketch & Stylized \\
\midrule
 RA=(2,10) & 61.07 & 68.77 & 3.95 & 85.84 & 12.88 & 48.50 & 21.95 & 11.21 & 4.84 \\
 RA=(2,5) & 64.62 & 72.57 & 4.59 & 80.79 & 15.44 & 52.37 & 25.68 & \textbf{14.48} & 8.36 \\
 RA=(1,10) & 63.64 & 71.26 & 4.64 & 83.04 & 14.53 & 51.37 & 23.67 & 13.14 & 7.27 \\
 RA=(1,5) & 64.96 & 72.54 & \textbf{4.80} & 81.32 & 14.94 & 52.05 & 25.03 & 13.69 & \textbf{8.98} \\
 RA=(1,3) & 64.88 & 72.66 & 4.80 & 79.04 & 15.61 & 52.59 & 25.43 & 13.54 & 8.13 \\
 RA=(1,0.8) & \textbf{65.33} & \textbf{73.19} & 4.79 & \textbf{77.08} & \textbf{16.16} & \textbf{53.03} & \textbf{25.98} & 14.15 & 8.98 \\
 RA=(1,0.4) & 64.27 & 72.17 & 4.69 & 78.10 & 15.46 & 52.18 & 24.99 & 13.47 & 8.59 \\
 RA=(1,0.1) & 63.58 & 71.41 & 4.80 & 79.23 & 15.39 & 51.43 & 23.66 & 12.54 & 8.36 \\
\end{tabular}
\caption{
ViT Ti/16 baseline training with different random augmentations. In this table, RA denotes the two RandAugment parameters: number of applied transforms, and mangitude of transforms. Note that weaker augmentation tends to perform better.
}
\label{tab:tiny_baselines}
\end{table*}

In Table \ref{tab:tiny_steps1} and \ref{tab:tiny_steps3}, we pick several of the better performing RandAugment parameters and then show  results from adversarial training with steps of 1 and 3, respectively.

Table \ref{tab:tiny_steps1} shows that the performance of adversarial training depends heavily on both the random augmentation and the type of attack. Note, RAm refers to the RandAugment mangitude parameter. As shown by RAm=0.1, pixel attacks can improve performance for in-distribution evaluation datasets when the random augmentation strength is low. However, at higher random augmentation, RAm=0.4 and RAm=0.8, PixelAT leads to the commonly observed trade-off between clean performance and adversarial robustness. In contrast, pyramid tends to improve performance across the board regardless of the starting augmentation (for all RAm of 0.1, 0.4, and 0.8). Interestingly, PixelAT exhibits better robustness properties (out-of-distribution performance) than PyramidAT for Ti/16. We hypothesize that the limited capacity can be ``spent'' on either in-distribution or out-of-distribution representations and that pyramid tends to bias the network towards in-distribution as opposed to pixel which has a bias towards out-of-distribution.

\begin{table*}[!htbp]\centering\small
\begin{tabular}{l|cc|cccc|ccc}
\toprule
&\multicolumn{2}{c|}{}&\multicolumn{7}{c}{Out of Distribution Robustness Test} \\
 Method & ImageNet & Real & A & C$\downarrow$ & ObjectNet & V2 & Rendition & Sketch & Stylized \\
\midrule
 Ti/16 RAm=0.1 & 63.58 & 71.41 & 4.80 & 79.23 & 15.39 & 51.43 & 23.66 & 12.54 & 8.36 \\
 +PixelAT steps=1 & 64.66 & 72.75 & 4.39 & 74.54 & 14.61 & 52.05 & \textbf{32.52} & \textbf{17.65} & \textbf{14.77} \\
 +PyramidAT steps=1 & \textbf{65.49} & \textbf{73.53} & \textbf{5.16} & \textbf{74.30} & \textbf{16.15} & \textbf{53.08} & 29.18 & 16.55 & 12.58 \\
\midrule
 Ti/16 RAm=0.4 & 64.27 & 72.17 & 4.69 & 78.10 & 15.46 & 52.18 & 24.99 & 13.47 & 8.59 \\
 +PixelAT steps=1 & 62.78 & 70.53 & 4.05 & 77.67 & 13.89 & 50.37 & \textbf{29.75} & \textbf{16.35} & 11.80 \\
 +PyramidAT steps=1 & \textbf{65.61} & \textbf{73.68} & \textbf{4.80} & \textbf{74.72} & \textbf{15.97} & \textbf{52.88} & 28.89 & 16.14 & \textbf{11.95} \\
\midrule
 Ti/16 RAm=0.8 & 65.33 & 73.19 & 4.79 & 77.08 & \textbf{16.16} & 53.03 & 25.98 & 14.15 & 8.98 \\
 +PixelAT steps=1 & 63.49 & 71.64 & 3.80 & 75.68 & 13.79 & 51.13 & \textbf{31.80} & \textbf{17.65} & \textbf{13.91} \\
 +PyramidAT steps=1 & \textbf{65.67} & \textbf{73.73} & \textbf{4.84} & \textbf{75.25} & 15.71 & \textbf{53.43} & 29.08 & 16.41 & 12.97 \\
\end{tabular}
\caption{
ViT Ti/16 adversarial training experiments with steps=1. RAm gives the RandAugment magnitude parameter; all experiments have RandAugment number of transforms equal to 1.
Pixel's performance on 0.4 and 0.8 is consistent with earlier work that suggests that adversarial training causes a trade-off
between in distribution and out of distribution datasets. However, we show that a low random augmentation starting point can break this trade-off and lead to gains. Pyramid tends
to outperform pixel on in-distribution performance for all random augmentations. However, pixel performs well for out-of-distribution datasets.
}
\label{tab:tiny_steps1}
\end{table*}

Table \ref{tab:tiny_steps3} shows that the strength of the adversarial attack also matters substantially to the overall performance of the model. Both attacks, pixel and pyramid, with 3 steps tend to degrade the model's performance on in-distribution evaluation datasets. Adversarial training still provides some benefits for out-of-distribution, with pyramid performing the best in terms of robustness. We hypothesize that pyramid outperforms pixel in the steps=3 runs because pyramid is a weaker out-of-distribution augmentation and pixel at 3 steps has over-regularized the network, leading to decreased performance. Note that pyramid at steps=3 produces the best out-of-distribution performance out of any Ti/16 runs including strong random augmentation and PixelAT at steps=1.

\begin{table*}[!htbp]\centering\small
\begin{tabular}{l|cc|cccc|ccc}
\toprule
&\multicolumn{2}{c|}{}&\multicolumn{7}{c}{Out of Distribution Robustness Test} \\
 Method & ImageNet & Real & A & C$\downarrow$ & ObjectNet & V2 & Rendition & Sketch & Stylized \\
\midrule
 Ti/16 RAm=0.1 & \textbf{63.58} & \textbf{71.41} & \textbf{4.80} & 79.23 & \textbf{15.39} & \textbf{51.43} & 23.66 & 12.54 & 8.36 \\
 +PixelAT steps=3 & 59.91 & 67.77 & 3.39 & 81.44 & 12.04 & 47.28 & 28.52 & 14.78 & 11.02 \\
 +PyramidAT steps=3 & 62.71 & 71.08 & 4.08 & \textbf{74.49} & 14.72 & 50.25 & \textbf{35.05} & \textbf{20.67} & \textbf{18.67} \\
\midrule
 Ti/16 RAm=0.8 & \textbf{65.33} & \textbf{73.19} & \textbf{4.79} & 77.08 & \textbf{16.16} & \textbf{53.03} & 25.98 & 14.15 & 8.98 \\
 +PixelAT steps=3 & 60.10 & 67.88 & 3.36 & 80.48 & 11.90 & 47.78 & 29.83 & 15.21 & 13.44 \\
 +PyramidAT steps=3 & 62.92 & 71.11 & 4.05 & \textbf{74.70} & 14.76 & 50.77 & \textbf{34.74} & \textbf{20.35} & \textbf{18.13} \\
\end{tabular}
\caption{
ViT Ti/16 adversarial training experiments with steps=3. RAm gives the RandAugment magnitude parameter; all experiments have RandAugment number of transforms equal to 1.
All techniques degrade from the baseline suggesting that 3 adversarial steps produces augmentation that is too strong for Ti's capacity.
}
\label{tab:tiny_steps3}
\end{table*}

\subsection{MLP-Mixer}

As shown in the main paper, we observe gains across the board for MLP-Mixer with PyramidAT. Here, we show that the gain is robust to a change in the LR schedule and that the gain is, again, affected by the starting augmentation.

Table \ref{tab:mlp1} shows baseline and adversarially trained models for two different training schedules of MLP-Mixer, one with the default LR schedule of 10k warm-up steps and then linear decay to an end learning rate (LR) of $1e-5$ and another with a more aggressive end learning rate of $1e-7$. We show that this change in LR schedule does not affect the gains from  adversarial training.

\begin{table*}[!htbp]\centering\small
\begin{tabular}{l|cc|cccc|ccc}
\toprule
&\multicolumn{2}{c|}{}&\multicolumn{7}{c}{Out of Distribution Robustness Test} \\
 Method & ImageNet & Real & A & C$\downarrow$ & ObjectNet & V2 & Rendition & Sketch & Stylized \\
\midrule
 MLP-Mixer~\cite{tolstikhin2021mlp} end LR=1e-5 (default) & 78.27 & 83.64 & 10.84 & 58.50 & 25.90 & 64.97 & 38.51 & 29.00 & 10.08 \\
 +PixelAT & 77.17 & 82.99 & 9.93 & 57.68 & 24.75 & 64.03 & 44.43 & 33.68 & \textbf{15.31} \\
 +PyramidAT & \textbf{79.29} & \textbf{84.78} & \textbf{12.97} & \textbf{52.88} & \textbf{28.60} & \textbf{66.56} & \textbf{45.34} & \textbf{34.79} & 14.77 \\
\midrule
 MLP-Mixer~\cite{tolstikhin2021mlp} end LR=1e-7 & 75.92 & 81.28 & 9.45 & 64.29 & 22.13 & 62.17 & 33.70 & 25.15 & 7.27 \\
 +PixelAT & 74.96 & 80.81 & 7.81 & 61.85 & 20.87 & 60.94 & 39.82 & 28.59 & 12.27 \\
 +PyramidAT & \textbf{77.98} & \textbf{83.60} & \textbf{11.17} & \textbf{56.19} & \textbf{25.59} & \textbf{64.92} & \textbf{41.65} & \textbf{31.99} & \textbf{12.66} \\
\end{tabular}
\caption{
MLP-Mixer ablations with different training. The pyramid gains are preserved even with different training schedules.
}
\label{tab:mlp1}
\end{table*}

Table \ref{tab:mlp2} shows that, similar to ViT Ti/16, the gains are improved when the random augmentation is weakened. However, in this case, the gain is not enough to overcome the drop in performance from using the weaker augmentation.

\begin{table*}[!htbp]\centering\small
\begin{tabular}{l|cc|cccc|ccc}
\toprule
&\multicolumn{2}{c|}{}&\multicolumn{7}{c}{Out of Distribution Robustness Test} \\
 Method & ImageNet & Real & A & C$\downarrow$ & ObjectNet & V2 & Rendition & Sketch & Stylized \\
\midrule
 MLP-Mixer~\cite{tolstikhin2021mlp} end LR=1e-7, RA=(2,15) & 75.92 & 81.28 & 9.45 & 64.29 & 22.13 & 62.17 & 33.70 & 25.15 & 7.27 \\
 +PixelAT & 74.96 & 80.81 & 7.81 & 61.85 & 20.87 & 60.94 & 39.82 & 28.59 & 12.27 \\
 +PyramidAT & \textbf{77.98} & \textbf{83.60} & \textbf{11.17} & \textbf{56.19} & \textbf{25.59} & \textbf{64.92} & \textbf{41.65} & \textbf{31.99} & \textbf{12.66} \\
\midrule
 MLP-Mixer~\cite{tolstikhin2021mlp} end LR=1e-7, RA=(2,9) & 73.56 & 79.01 & 7.79 & 67.18 & 18.09 & 60.30 & 31.47 & 22.80 & 7.58 \\
 +PixelAT & 72.44 & 78.27 & 7.80 & 63.38 & 16.96 & 58.40 & 37.18 & 25.78 & 12.27 \\
 +PyramidAT & \textbf{76.19} & \textbf{81.47} & \textbf{10.61} & \textbf{57.71} & \textbf{21.86} & \textbf{63.03} & \textbf{39.52} & \textbf{29.40} & \textbf{14.37} \\
\end{tabular}
\caption{
MLP-Mixer ablations with random augmentation magnitude. Weaker random augmentations lead to larger gains from the adversarial training but with these parameters do not lead to better overall performance than the
strong random augmentation plus adversarial training. Note that for both the starting point and weaker random augmentation, the gain from pyramid is substantial compared to the gain from pixel.
}
\label{tab:mlp2}
\end{table*}

\section{Additional Ablations}
\label{sup:ablations}
\subsection{Dropout}

One of the key findings of this paper is the importance of ``matched'' Dropout\cite{dropout:2014} and stochastic depth\cite{stochastic_depth:2016}. Here we describe numerous ablations on these Dropout terms and list several detailed findings including:

\begin{itemize}
    \item Matching the Dropout and stochastic depth matters significantly for balanced clean performance and robustness.
    \item Running without Dropout in the adversarial training branch can improve robustness even more.
    \item Dropout matters more than Stochastic Depth
\end{itemize}

Note that in the tables below, we use the term ``dropparams'' to refer to a tuple of the Dropout probability and stochastic depth probability. Clean dropparams (abbreviated as c\_dp) refer to the dropparams used for the clean training branch; adversarial dropparams (abbreviated as a\_dp) refer to the dropparams used for the adversarial training branch; and matched dropparams (abbreviated as m\_dp) refer to dropparams used for both clean and adversarial branches. So $\textrm{c\_dp}=(10, 0)$ means that the clean training branch had a 10\% probability of Dropout but a 0\% probability of stochastic depth. 

Table \ref{tab:adv_aux_sweep} explores different possible values for adversarial dropparams. In general, lower values of Dropout and stochastic depth in the adversarial branch improve out-of-distribution performance while hurting in-distribution performance; however, the opposite is not true: higher levels of Dropout and stochastic depth in the adversarial branch do not improve in-distribution performance. In-distribution performance seems to peak when the params for the adversarial and clean branches match.

\begin{table*}[!htbp]\centering\small
\begin{tabular}{l|cc|cccc|ccc}
\toprule
&\multicolumn{2}{c|}{}&\multicolumn{7}{c}{Out of Distribution Robustness Test} \\
 Method & ImageNet & Real & A & C$\downarrow$ & ObjectNet & V2 & Rendition & Sketch & Stylized \\
\midrule
 Baseline $\textrm{c\_dp}=(10,10)$ & 79.92 & 85.14 & 17.48 & 52.46 & 29.30 & 67.49 & 38.24 & 29.08 & 11.02 \\
 PixelAT $\textrm{a\_dp}=(0,0)$ & 79.13 & 84.75 & 14.37 & 52.73 & 29.01 & 66.94 & \textbf{49.94} & \textbf{37.03} & \textbf{22.34} \\
 PixelAT $\textrm{a\_dp}=(5,5)$ & 78.41 & 84.39 & 14.00 & 54.50 & 28.04 & 66.30 & 49.29 & 36.84 & 21.80 \\
 PixelAT $\textrm{a\_dp}=(10,10)$ & 80.42 & 85.78 & 19.15 & 47.68 & 30.11 & 68.78 & 45.39 & 34.40 & 18.28 \\
 PixelAT $\textrm{a\_dp}=(20,20)$ & \textbf{81.00} & \textbf{86.16} & \textbf{21.43} & \textbf{46.07} & \textbf{30.85} & \textbf{69.60} & 46.00 & 34.73 & 18.36 \\
 PixelAT $\textrm{a\_dp}=(30,30)$ & 76.53 & 82.56 & 16.20 & 66.09 & 24.22 & 64.40 & 42.93 & 34.89 & 15.00 \\
\midrule
 Baseline $\textrm{c\_dp}=(10,10)$ & 79.92 & 85.14 & 17.48 & 52.46 & 29.30 & 67.49 & 38.24 & 29.08 & 11.02 \\
 PyramidAT $\textrm{a\_dp}=(0,0)$ & 79.31 & 85.13 & 13.95 & 55.09 & 29.98 & 67.43 & \textbf{53.69} & \textbf{41.07} & 23.44 \\
 PyramidAT $\textrm{a\_dp}=(5,5)$ & 79.13 & 84.90 & 13.43 & 54.08 & 28.99 & 67.50 & 52.39 & 40.23 & \textbf{24.84} \\
 PyramidAT $\textrm{a\_dp}=(10,10)$ & \textbf{81.71} & \textbf{86.82} & 22.99 & \textbf{44.99} & \textbf{32.92} & \textbf{70.82} & 47.66 & 36.77 & 19.14 \\
 PyramidAT $\textrm{a\_dp}=(20,20)$ & 81.67 & 86.76 & \textbf{23.33} & 45.19 & 32.74 & 70.58 & 48.15 & 38.01 & 17.50 \\
 PyramidAT $\textrm{a\_dp}=(30,30)$ & 74.26 & 80.57 & 11.55 & 69.85 & 22.87 & 61.46 & 41.59 & 29.89 & 17.66 \\
\end{tabular}
\caption{
Ablation on the values of Dropout and stochastic depth for adversarial training branch. All $\textrm{c\_dp}$ are kept constant at $(10, 10)$ in the
adversarial section.
}
\label{tab:adv_aux_sweep}
\end{table*}

Table \ref{tab:matched_aux_sweep} explores different possible values for matched dropparams. In general, the dropparams determined by RegViT\cite{how_to_train_your_vit:2021} seem to be roughly optimal for both the baselines and the adversarially trained models, with some variation for some datasets.

\begin{table*}[!htbp]\centering\small
\begin{tabular}{l|cc|cccc|ccc}
\toprule
&\multicolumn{2}{c|}{}&\multicolumn{7}{c}{Out of Distribution Robustness Test} \\
 Method & ImageNet & Real & A & C$\downarrow$ & ObjectNet & V2 & Rendition & Sketch & Stylized \\
\midrule
 Baseline $\textrm{c\_dp}=(0,0)$ & 75.69 & 80.92 & 12.68 & 61.78 & 23.65 & 62.49 & 34.23 & 24.38 & 7.73 \\
 Baseline $\textrm{c\_dp}=(5,5)$ & 79.05 & 84.19 & 16.52 & 54.32 & 28.27 & 66.40 & 37.38 & 28.08 & 9.53 \\
 Baseline $\textrm{c\_dp}=(10,10)$ & \textbf{79.92} & \textbf{85.14} & 17.48 & \textbf{52.46} & \textbf{29.30} & \textbf{67.49} & 38.24 & \textbf{29.08} & 11.02 \\
 Baseline $\textrm{c\_dp}=(15,15)$ & 79.35 & 84.74 & \textbf{17.71} & 52.96 & 29.11 & 67.47 & \textbf{39.13} & 28.67 & \textbf{11.41} \\
 Baseline $\textrm{c\_dp}=(20,20)$ & 78.40 & 84.24 & 16.16 & 54.93 & 27.80 & 66.07 & 37.85 & 27.77 & 9.84 \\
\midrule
 PixelAT $\textrm{m\_dp}=(0,0)$ & 79.13 & 84.20 & 19.71 & 49.68 & 29.77 & 67.02 & 43.34 & 32.14 & 15.23 \\
 PixelAT $\textrm{m\_dp}=(5,5)$ & \textbf{80.82} & 85.59 & \textbf{22.12} & \textbf{46.15} & \textbf{31.27} & \textbf{68.99} & 44.86 & 33.31 & 16.95 \\
 PixelAT $\textrm{m\_dp}=(10,10)$ & 80.42 & \textbf{85.78} & 19.15 & 47.68 & 30.11 & 68.78 & 45.39 & \textbf{34.40} & 18.28 \\
 PixelAT $\textrm{m\_dp}=(15,15)$ & 79.03 & 84.88 & 15.16 & 50.67 & 27.72 & 66.53 & \textbf{46.43} & 34.21 & \textbf{21.95} \\
 PixelAT $\textrm{m\_dp}=(20,20)$ & 77.89 & 84.12 & 13.99 & 52.20 & 26.28 & 66.06 & 45.21 & 32.75 & 21.95 \\
\midrule
 PyramidAT $\textrm{m\_dp}=(0,0)$ & 79.54 & 84.66 & 18.91 & 50.02 & 30.10 & 67.44 & 44.43 & 33.46 & 13.67 \\
 PyramidAT $\textrm{m\_dp}=(5,5)$ & \textbf{81.80} & 86.59 & \textbf{23.47} & 45.52 & 32.76 & 70.36 & 46.62 & 36.50 & 17.27 \\
 PyramidAT $\textrm{m\_dp}=(10,10)$ & 81.71 & \textbf{86.82} & 22.99 & \textbf{44.99} & \textbf{32.92} & \textbf{70.82} & \textbf{47.66} & 36.77 & \textbf{19.14} \\
 PyramidAT $\textrm{m\_dp}=(15,15)$ & 80.95 & 86.49 & 21.33 & 46.18 & 31.62 & 69.70 & 47.21 & \textbf{36.83} & 18.91 \\
 PyramidAT $\textrm{m\_dp}=(20,20)$ & 79.56 & 85.77 & 18.35 & 49.14 & 29.73 & 68.13 & 45.72 & 35.41 & 18.36 \\
\end{tabular}
\caption{
Ablation on the values of Dropout and stochatic depth for matched attacks.
}
\label{tab:matched_aux_sweep}
\end{table*}

Table \ref{tab:dropout_importance} explores if one of these parameters is more important than the others. To do so, we set clean dropparams to $(10, 10)$ for the entire table (besides the included baselines) and only vary the adversarial dropparams. For both PixelAT and PyramidAT, the Dropout parameter seems to be more important for clean, in-distribution performance. Without Dropout, the top-1 of ImageNet drops $0.41$ for PixelAT and $0.92$ for PyramidAT. However, no Dropout does give a substantial boost to out-of-distribution performance, with Rendition gains of $11.59$ for PixelAT and $15.68$ for PyramidAT and Sketch gains of $7.49$ for PixelAT and $11.96$ for PyramidAT. Without stochastic depth, the adversarially trained models seem to perform roughly as well as with stochastic depth, exhibitly marginally more clean accuracy for PyramidAT than the model with both Dropout and stochastic depth. Our main takeaway is that Dropout seems to be the primary determinant in whether the gains are balanced between in-distribution and out-of-distribution or primarily focused on out-of-distribution. In fact, no Dropout PyramidAT performs so well on out-of-distribution that it sets new state-of-the-art numbers for Rendition and Sketch.

\begin{table*}[!htbp]\centering\small
\begin{tabular}{l|cc|cccc|ccc}
\toprule
&\multicolumn{2}{c|}{}&\multicolumn{7}{c}{Out of Distribution Robustness Test} \\
 Method & ImageNet & Real & A & C$\downarrow$ & ObjectNet & V2 & Rendition & Sketch & Stylized \\
\midrule
 Baseline $\textrm{c\_dp}=(0,0)$ & 75.69 & 80.92 & 12.68 & 61.78 & 23.65 & 62.49 & 34.23 & 24.38 & 7.73 \\
 Baseline $\textrm{c\_dp}=(10,10)$ & \textbf{79.92} & \textbf{85.14} & \textbf{17.48} & \textbf{52.46} & \textbf{29.30} & \textbf{67.49} & \textbf{38.24} & \textbf{29.08} & \textbf{11.02} \\
\midrule
 PixelAT $\textrm{a\_dp}=(0,10)$ & 79.40 & 84.98 & 14.32 & 52.17 & 28.87 & 67.24 & \textbf{49.83} & \textbf{36.57} & \textbf{21.88} \\
 PixelAT $\textrm{a\_dp}=(10,0)$ & 79.51 & 85.12 & 15.65 & 49.38 & 29.45 & 67.50 & 47.32 & 34.64 & 21.17 \\
 PixelAT $\textrm{a\_dp}=(10,10)$ & \textbf{80.42} & \textbf{85.78} & \textbf{19.15} & \textbf{47.68} & \textbf{30.11} & \textbf{68.78} & 45.39 & 34.40 & 18.28 \\
\midrule
 PyramidAT $\textrm{a\_dp}=(0,10)$ & 79.00 & 85.02 & 13.69 & 55.58 & 29.38 & 66.96 & \textbf{53.92} & \textbf{41.04} & \textbf{24.22} \\
 PyramidAT $\textrm{a\_dp}=(10,0)$ & \textbf{81.80} & 86.67 & \textbf{23.51} & 45.00 & \textbf{33.37} & 70.52 & 47.82 & 37.09 & 19.38 \\
 PyramidAT $\textrm{a\_dp}=(10,10)$ & 81.71 & \textbf{86.82} & 22.99 & \textbf{44.99} & 32.92 & \textbf{70.82} & 47.66 & 36.77 & 19.14 \\
\end{tabular}
\caption{
Ablation on the values of Dropout and stochastic depth for unmatched attacks. For the adversarial techniques, clean dropparams will be the same as RegVit at (10, 10).
For the adversarial training rows, either Dropout or stochastic depth will be 0 and the other will be the base value. This table explores whether one of these parameters
is more important than the other. For both PixelAT and PyramidAT, Dropout appears to be more important in determining the balance between in-distribution and out-of-distribution
performance. PyramidAT no Dropout is SOTA for Rendition and Sketch.
}
\label{tab:dropout_importance}
\end{table*}

Table \ref{tab:adv_uneven} explores parameter settings where the Dropout and stochastic depth are not equal. In general, there does not seem to be a consistent trend or recognizable pattern for the overall performance, though some patterns exist for specific attacks and evaluation datasets. For example, increasing stochastic depth probability for pixel attacks tend to improve Real, ImageNet-C, and ObjectNet performance.

\begin{table*}[!htbp]\centering\small
\begin{tabular}{l|cc|cccc|ccc}
\toprule
&\multicolumn{2}{c|}{}&\multicolumn{7}{c}{Out of Distribution Robustness Test} \\
 Method & ImageNet & Real & A & C$\downarrow$ & ObjectNet & V2 & Rendition & Sketch & Stylized \\
\midrule
 Baseline $\textrm{c\_dp}=(10,10)$ & \textbf{79.92} & \textbf{85.14} & \textbf{17.48} & \textbf{52.46} & \textbf{29.30} & \textbf{67.49} & \textbf{38.24} & \textbf{29.08} & \textbf{11.02} \\
\midrule
 PixelAT $\textrm{a\_dp}=(0,10)$ & 79.40 & 84.98 & 14.32 & 52.17 & 28.87 & 67.24 & \textbf{49.83} & \textbf{36.57} & \textbf{21.88} \\
 PixelAT $\textrm{a\_dp}=(10,10)$ & \textbf{80.42} & \textbf{85.78} & 19.15 & \textbf{47.68} & \textbf{30.11} & \textbf{68.78} & 45.39 & 34.40 & 18.28 \\
 PixelAT $\textrm{a\_dp}=(20,10)$ & 74.58 & 80.74 & 13.07 & 69.14 & 23.93 & 61.82 & 42.64 & 33.54 & 17.42 \\
 PixelAT $\textrm{a\_dp}=(30,10)$ & 80.14 & 85.59 & \textbf{20.91} & 48.99 & 29.74 & 68.50 & 44.62 & 35.06 & 16.02 \\
\midrule
 PixelAT $\textrm{a\_dp}=(10,0)$ & 79.51 & 85.12 & 15.65 & 49.38 & 29.45 & 67.50 & \textbf{47.32} & 34.64 & \textbf{21.17} \\
 PixelAT $\textrm{a\_dp}=(10,10)$ & 80.42 & 85.78 & 19.15 & 47.68 & 30.11 & 68.78 & 45.39 & 34.40 & 18.28 \\
 PixelAT $\textrm{a\_dp}=(10,20)$ & \textbf{81.14} & 86.22 & \textbf{21.37} & 45.91 & 31.53 & \textbf{69.75} & 46.24 & \textbf{35.14} & 19.45 \\
 PixelAT $\textrm{a\_dp}=(10,30)$ & 80.94 & \textbf{86.26} & 21.37 & \textbf{45.19} & \textbf{31.63} & 69.69 & 45.76 & 34.69 & 19.61 \\
\midrule
 PyramidAT $\textrm{a\_dp}=(0,10)$ & 79.00 & 85.02 & 13.69 & 55.58 & 29.38 & 66.96 & \textbf{53.92} & \textbf{41.04} & \textbf{24.22} \\
 PyramidAT $\textrm{a\_dp}=(10,10)$ & \textbf{81.71} & \textbf{86.82} & \textbf{22.99} & \textbf{44.99} & \textbf{32.92} & \textbf{70.82} & 47.66 & 36.77 & 19.14 \\
 PyramidAT $\textrm{a\_dp}=(20,10)$ & 75.36 & 81.40 & 14.03 & 70.76 & 23.65 & 63.14 & 39.36 & 26.92 & 15.00 \\
 PyramidAT $\textrm{a\_dp}=(30,10)$ & 79.18 & 84.91 & 17.43 & 56.44 & 28.95 & 67.81 & 43.64 & 29.91 & 19.45 \\
\midrule
 PyramidAT $\textrm{a\_dp}=(10,0)$ & \textbf{81.80} & 86.67 & \textbf{23.51} & 45.00 & \textbf{33.37} & 70.52 & 47.82 & 37.09 & \textbf{19.38} \\
 PyramidAT $\textrm{a\_dp}=(10,10)$ & 81.71 & \textbf{86.82} & 22.99 & \textbf{44.99} & 32.92 & \textbf{70.82} & 47.66 & 36.77 & 19.14 \\
 PyramidAT $\textrm{a\_dp}=(10,20)$ & 81.50 & 86.58 & 23.13 & 45.72 & 32.45 & 70.48 & \textbf{48.08} & 36.91 & 17.66 \\
 PyramidAT $\textrm{a\_dp}=(10,30)$ & 81.53 & 86.75 & 21.87 & 45.59 & 32.80 & 70.34 & 47.73 & \textbf{37.12} & 17.89 \\
\end{tabular}
\caption{
Ablation on settings where the Dropout parameter and stochastic depth parameter are not equal for adversarial training branch. All $\textrm{c\_dp}$ are kept constant at $(10, 10)$ in the
adversarial section.
}
\label{tab:adv_uneven}
\end{table*}

Table \ref{tab:clean_at_zero} explores the effects of adversarial training with different dropparams without Dropout or stochastic depth in the main branch. In general, the lack of Dropout and stochastic depth in the clean branch has a substantial negative effect on the performance of the model and all of the resulting models under-perform their counterparts with non-zero clean dropparams. In this setting, adversarial training does provide substantial improvements for both in-distribution ($+5.18$ for clean using PyramidAT) and out-of-distribution performances ($+12.29$ for Rendition using PyramidAT and $+11.68$ for Sketch using PyramidAT), but not enough to offset the poor starting performance of the baseline model.

\begin{table*}[!htbp]\centering\small
\begin{tabular}{l|cc|cccc|ccc}
\toprule
&\multicolumn{2}{c|}{}&\multicolumn{7}{c}{Out of Distribution Robustness Test} \\
 Method & ImageNet & Real & A & C$\downarrow$ & ObjectNet & V2 & Rendition & Sketch & Stylized \\
\midrule
 Baseline $\textrm{c\_dp}=(0,0)$ & 75.69 & 80.92 & 12.68 & 61.78 & 23.65 & 62.49 & 34.23 & 24.38 & 7.73 \\
 PixelAT $\textrm{c\_dp}=(0,0)$ $\textrm{a\_dp}=(5,5)$ & 79.86 & 84.78 & 21.23 & 48.66 & 30.19 & 67.92 & \textbf{43.94} & 32.69 & \textbf{15.00} \\
 PixelAT $\textrm{c\_dp}=(0,0)$ $\textrm{a\_dp}=(10,10)$ & 79.82 & 84.81 & 21.32 & \textbf{47.82} & 30.84 & 67.99 & 43.25 & 32.24 & 14.92 \\
 PixelAT $\textrm{c\_dp}=(0,0)$ $\textrm{a\_dp}=(20,20)$ & \textbf{80.03} & \textbf{84.96} & \textbf{21.69} & 47.86 & \textbf{31.34} & \textbf{68.16} & 43.56 & \textbf{32.78} & 14.77 \\
\midrule
 Baseline $\textrm{c\_dp}=(0,0)$ & 75.69 & 80.92 & 12.68 & 61.78 & 23.65 & 62.49 & 34.23 & 24.38 & 7.73 \\
 PyramidAT $\textrm{c\_dp}=(0,0)$ $\textrm{a\_dp}=(5,5)$ & 80.79 & 85.66 & \textbf{23.65} & 47.39 & 32.39 & 69.12 & 46.30 & \textbf{36.40} & 15.70 \\
 PyramidAT $\textrm{c\_dp}=(0,0)$ $\textrm{a\_dp}=(10,10)$ & 80.69 & 85.55 & 23.25 & 47.14 & \textbf{32.74} & 69.21 & 46.49 & 36.09 & 15.78 \\
 PyramidAT $\textrm{c\_dp}=(0,0)$ $\textrm{a\_dp}=(15,15)$ & \textbf{80.87} & \textbf{85.74} & 23.63 & \textbf{46.68} & 32.67 & \textbf{69.27} & \textbf{46.52} & 36.06 & \textbf{17.19} \\
\end{tabular}
\caption{
Ablation on the values of Dropout and stochastic depth for adversarial training branch for a clean branch with no Dropout or stochastic depth; specifically, all $\textrm{c\_dp}$ are kept constant at $(0, 0)$. The loss of Dropout and stochastic depth causes poor performance across the board.
}
\label{tab:clean_at_zero}
\end{table*}

\subsection{Pyramid Structure}

In the main paper, Table 8 presented an abridged version (with only a subset of the evaluation datasets) of an ablation on the structure of the pyramid used in the pyramid adversarial training. We present the full version (complete with all the evaluation datasets) of this ablation in Table \ref{tab:pyramid_structure}. This table remains consistent with the description and explanation in the main table: adding more layers to the pyramid tends to improve performance. In fact, Table \ref{tab:pyramid_structure} shows the full extent of the trade-off between the 3rd and 4th levels of the pyramid. Specifically, the 4th level seems to lead to a slight improvement in out-of-distribution performance and a slight decline in in-distribution performance. Note that 2-level Pyramid is simply the combination of Pixel and Patch.

\begin{table*}[!htbp]\centering\small
\begin{tabular}{l|cc|cccc|ccc}
\toprule
&\multicolumn{2}{c|}{}&\multicolumn{7}{c}{Out of Distribution Robustness Test} \\
 Method & ImageNet & Real & A & C$\downarrow$ & ObjectNet & V2 & Rendition & Sketch & Stylized \\
\midrule
 Pixel & 80.42 & 85.78 & 19.15 & 47.68 & 30.11 & 68.78 & 45.39 & 34.40 & 18.28 \\
 Patch & 81.20 & 86.10 & 21.33 & 50.30 & 31.87 & 68.98 & 42.87 & 33.75 & 15.00 \\
 2-level Pyramid & 81.65 & 86.69 & 22.79 & 45.27 & 32.46 & 69.86 & 47.00 & 36.71 & 19.06 \\
 3-level Pyramid & \textbf{81.71} & \textbf{86.82} & 22.99 & \textbf{44.99} & \textbf{32.92} & \textbf{70.82} & 47.66 & 36.77 & 19.14 \\
 4-level Pyramid & 81.66 & 86.68 & \textbf{23.21} & 45.29 & 32.85 & 70.56 & \textbf{47.68} & \textbf{37.41} & \textbf{20.47} \\
\end{tabular}
\caption{
Pyramid structure ablations. This shows the effect of the number of layers of the pyramid.
Adding coarser layers with larger magnitudes generally improves performance. 
}
\label{tab:pyramid_structure}
\end{table*}

Using the the scale notation established in 3.2 Pyramid Adversarial Training, the details of these layers are as follows in Table \ref{tab:pyramid_details}.

\begin{table*}[!htbp]\centering
\begin{tabular}{l|lcc}
Layer & Name & s & $m_s$ \\ \toprule
Layer 1 & Pixel & 1 & 1 \\
Layer 2 & Patch & 16 & 10 \\
Layer 3 & $2\times 2$ patches & 32 & 20 \\
Layer 4 & Global & 224 & 25 \\
\end{tabular}
\caption{Pyramid details.}
\label{tab:pyramid_details}
\end{table*}

In Table \ref{tab:pyramid_patch_magnitudes}, we explore different magnitudes for the patch level. We note that some of the gains from 2-level are from the higher magnitude for the coarse level. 

\begin{table*}[!htbp]\centering\small
\begin{tabular}{l|cc|cccc|ccc}
\toprule
&\multicolumn{2}{c|}{}&\multicolumn{7}{c}{Out of Distribution Robustness Test} \\
 Method & ImageNet & Real & A & C$\downarrow$ & ObjectNet & V2 & Rendition & Sketch & Stylized \\
\midrule
 Pixel $m=1$ & 80.42 & 85.78 & 19.15 & 47.68 & 30.11 & 68.78 & 45.39 & 34.40 & 18.28 \\
 Patch $m=1$ & 80.09 & 85.09 & 18.40 & 52.17 & 29.78 & 68.09 & 40.46 & 30.46 & 13.83 \\
 Patch $m=10$ & 80.95 & 85.77 & 20.01 & 50.62 & 31.37 & 68.86 & 42.46 & 32.65 & 11.64 \\
 Patch $m=20$ & 81.20 & 86.10 & 21.33 & 50.30 & 31.87 & 68.98 & 42.87 & 33.75 & 15.00 \\
 2-level Pyramid $m=[10,1]$ & \textbf{81.65} & \textbf{86.69} & \textbf{22.79} & \textbf{45.27} & \textbf{32.46} & \textbf{69.86} & \textbf{47.00} & \textbf{36.71} & \textbf{19.06} \\
\end{tabular}
\caption{
Pyramid structure ablations where $m$ is the multiplicative term of the perturbation. Shows that the combination of patch and pixel is better than only patch, even when
patch is tested at different magnitudes.
}
\label{tab:pyramid_patch_magnitudes}
\end{table*}

We additionally include Table \ref{tab:pyramid_random} which shows a random subset of pyramid structures tested. The best pyramids tend to be structured based on the patches of the ViT.

\begin{table*}[!htbp]
  \centering
  \begin{tabular}{lc|ccccccc}
    \toprule
    Scale factor & Strengths &
    ImageNet \cite{imagenet:2009} & 
    Real \cite{imagenet:2009} & 
    A \cite{imagenet-a:2021} & 
    C \cite{imagenet-c:2019}$\downarrow$ & 
    Rendition \cite{imagenet-r:2021} & 
    Sketch \cite{stylized-imagenet:2019} &
    Stylized \cite{imagenet-sketch:2019} \\ 
    \midrule
    $[224, 16, 1]$ & $[20, 10, 1]$ &  81.37	& 86.50	& 21.65	& 45.84	& 46.72& 	36.33& 	17.97 \\
    $[32, 16, 1]$ & $[20, 10, 1]$ &  \textbf{81.71} &	\textbf{86.82}&	\textbf{22.99}&	\textbf{44.99}&	\textbf{47.66} &	36.77&	\textbf{19.14} \\
    $[224, 32, 16, 1]$ & $[20, 10, 5, 1]$ &  81.49& 	86.66& 	21.93& 	45.89& 	47.08& 	37.22& 18.98 \\
    $[16, 1]$ & $[10, 1]$ & 81.65	& 86.69	& 22.79	& 45.27	& 47.00	& 36.71& 	19.06 \\
    $[16, 4, 1]$ & $[20, 10, 1]$ &  81.43&	86.59&	21.83&	49.15&	47.49&	\textbf{37.85}&	17.19 \\
    \bottomrule
  \end{tabular}
  \caption{Random set of pyramid configurations.}
  \label{tab:pyramid_random}
\end{table*}

\subsection{More epochs for baseline}

We tested the effect of additional epochs for the baseline training. We found that going from $300$ epochs to $500$ (with the learning rate being adjusted accordingly) did not provide any benefits to the network's performance. In fact, Table \ref{tab:long_run} shows that the longer run performs worse in most evaluation datasets than the shorter run.

\begin{table*}[!htbp]\centering\small
\begin{tabular}{l|cc|cccc|ccc}
\toprule
&\multicolumn{2}{c|}{}&\multicolumn{7}{c}{Out of Distribution Robustness Test} \\
 Method & ImageNet & Real & A & C$\downarrow$ & ObjectNet & V2 & Rendition & Sketch & Stylized \\
\midrule
 Baseline 300 epoch.& \textbf{79.92} & \textbf{85.14} & 17.48 & \textbf{52.46} & \textbf{29.30} & \textbf{67.49} & \textbf{38.24} & \textbf{29.08} & \textbf{11.02} \\
 Baseline 500 epoch. & 79.34 & 78.83 & \textbf{18.39} & 54.28 & 28.43 & 66.69 & 37.64 & 27.60 & 10.31 \\
\end{tabular}
\caption{Exploration of the number of steps for the baseline.}
\label{tab:long_run}
\end{table*}

\subsection{Number of Attack Steps}

We perform an ablation on the number of steps in the adversarial attack. AdvProp\cite{advprop:2020} uses 5 for their main paper; we also adopt this parameter as a reasonable balance between performance and train time (each additional step in the attack requires a forward and backward pass of the model and increases the train time accordingly). Table ~\ref{tab:num_steps} shows that higher number of steps tends to lead to better performance for both pixel and pyramid.

\begin{table*}[!htbp]\centering\small
\begin{tabular}{l|cc|cccc|ccc}
\toprule
&\multicolumn{2}{c|}{}&\multicolumn{7}{c}{Out of Distribution Robustness Test} \\
 Method & ImageNet & Real & A & C$\downarrow$ & ObjectNet & V2 & Rendition & Sketch & Stylized \\
\midrule
 PixelAT steps=1 & 80.46 & 85.48 & 17.96 & 49.59 & 30.12 & 68.66 & 43.31 & 32.12 & 13.59 \\
 PixelAT steps=3 & 80.39 & 85.62 & 17.71 & 48.50 & 29.67 & 68.58 & \textbf{46.56} & 33.84 & 17.50 \\
 PixelAT steps=5 & 80.42 & 85.78 & 19.15 & 47.68 & 30.11 & 68.78 & 45.39 & 34.40 & 18.28 \\
 PixelAT steps=7 & \textbf{80.77} & \textbf{86.03} & \textbf{20.44} & \textbf{46.31} & \textbf{31.12} & \textbf{69.25} & 45.95 & \textbf{34.54} & \textbf{18.59} \\
\midrule
 PyramidAT steps=1 & 79.93 & 84.89 & 18.17 & 50.92 & 28.91 & 68.13 & 40.50 & 30.48 & 12.81 \\
 PyramidAT steps=3 & 81.47 & 86.46 & 22.39 & 46.21 & 32.33 & 70.11 & 45.07 & 34.89 & 18.20 \\
 PyramidAT steps=5 & \textbf{81.71} & \textbf{86.82} & 22.99 & \textbf{44.99} & \textbf{32.92} & \textbf{70.82} & \textbf{47.66} & \textbf{36.77} & \textbf{19.14} \\
 PyramidAT steps=7 & 81.65 & 86.69 & \textbf{23.63} & 45.33 & 32.53 & 70.47 & 47.29 & 36.52 & 16.95 \\
\end{tabular}
\caption{
Ablation on the number of steps in the adversarial attack. For both PixelAT and PyramidAT, larger number of steps tend to give higher performance. Note that increasing the number of steps also increases the train time.
We chose 5 for both PixelAT and PyramidAT as a reasonable tradeoff between performance and train time.
}
\label{tab:num_steps}
\end{table*}

\subsection{Magnitude}

We also perform ablations on the magnitude of perturbations (specifically L2 of the difference between adversarial image and the original image) and show that there exists an inverted U curve for both PixelAT and PyramidAT where one perturbation setting tends to produce the best model for most evaluation datasets.

For PixelAT, we change the perturbation magnitude by editing the learning rate ($\textrm{lr}$) and the epsilon parameter ($\epsilon$) which is used for the clipping function. Since we use the SGD optimizer, a larger learning rate and epsilon will naturally lead to larger perturbations. Table \ref{tab:mag_pixel} shows the results of these experiments, which suggests that pixel attacks can very quickly become too large to help the overall network performance.

\begin{table*}[!htbp]\centering\small
\begin{tabular}{l|cc|cccc|ccc}
\toprule
&\multicolumn{2}{c|}{}&\multicolumn{7}{c}{Out of Distribution Robustness Test} \\
 Method & ImageNet & Real & A & C$\downarrow$ & ObjectNet & V2 & Rendition & Sketch & Stylized \\
\midrule
 Baseline & 79.92 & 85.14 & 17.48 & 52.46 & 29.30 & 67.49 & 38.24 & 29.08 & 11.02 \\
 PixelAT $\textrm{lr}=5/255, \epsilon=20/255$ & \textbf{81.25} & \textbf{86.61} & \textbf{21.59} & \textbf{44.07} & 32.06 & \textbf{69.75} & \textbf{47.59} & \textbf{37.29} & \textbf{20.39} \\
 PixelAT $\textrm{lr}=10/255, \epsilon=40/255$ & 80.93 & 86.35 & 21.43 & 45.58 & \textbf{32.88} & 69.57 & 45.40 & 35.01 & 18.44 \\
 PixelAT $\textrm{lr}=20/255, \epsilon=80/255$ & 80.59 & 85.90 & 20.31 & 46.11 & 30.83 & 69.05 & 45.08 & 33.39 & 18.28 \\
 PixelAT $\textrm{lr}=40/255, \epsilon=160/255$ & 80.27 & 85.59 & 18.95 & 48.61 & 30.34 & 68.40 & 42.12 & 32.97 & 15.16 \\
\end{tabular}
\caption{
Ablation on the magnitude of PixelAT. PixelAT tends to degrade with higher $\textrm{lr}$ and $\epsilon$.
}
\label{tab:mag_pixel}
\end{table*}

For PyramidAT, we adjust the perturbation size by editing the magnitude of the multiplicative terms. In Table \ref{tab:mag_pyramid}, we perform an exhaustive sweep of these terms starting with an initial list of $[20, 10, 1]$ and multiplying the list by a constant. This table shows that there also exists an inverted U curve where the performance will degrade if the perturbation magnitude is either too small or too big.

\begin{table*}[!htbp]\centering\small
\begin{tabular}{l|cc|cccc|ccc}
\toprule
&\multicolumn{2}{c|}{}&\multicolumn{7}{c}{Out of Distribution Robustness Test} \\
 Method & ImageNet & Real & A & C$\downarrow$ & ObjectNet & V2 & Rendition & Sketch & Stylized \\
\midrule
 Baseline & 79.92 & 85.14 & 17.48 & 52.46 & 29.30 & 67.49 & 38.24 & 29.08 & 11.02 \\
 PyramidAT $m = [1, 0.5, 0.05]$ & 80.94 & 85.71 & 19.29 & 49.97 & 30.57 & 68.98 & 43.38 & 32.99 & 13.52 \\
 PyramidAT $m = [2, 1, 0.1]$ & 80.94 & 85.71 & 19.29 & 49.97 & 30.57 & 68.98 & 43.38 & 32.99 & 13.52 \\
 PyramidAT $m = [4, 2, 0.2]$ & 80.94 & 85.71 & 19.29 & 49.97 & 30.57 & 68.98 & 43.38 & 32.99 & 13.52 \\
 PyramidAT $m = [8, 4, 0.4]$ & 80.73 & 86.02 & 20.20 & 47.20 & 31.36 & 69.08 & 44.10 & 33.48 & 18.83 \\
 PyramidAT $m = [10, 5, 0.5]$ & 80.94 & 85.71 & 19.29 & 49.97 & 30.57 & 68.98 & 43.38 & 32.99 & 13.52 \\
 PyramidAT $m = [12, 6, 0.6]$ & 80.47 & 85.83 & 19.32 & 47.48 & 30.86 & 68.97 & 44.10 & 33.46 & 17.03 \\
 PyramidAT $m = [20, 10, 1]$ & \textbf{81.71} & \textbf{86.82} & 22.99 & 44.99 & 32.92 & \textbf{70.82} & 47.66 & 36.77 & \textbf{19.14} \\
 PyramidAT $m = [22, 11, 1.1]$ & 81.71 & 86.70 & \textbf{23.55} & \textbf{44.84} & \textbf{32.98} & 70.57 & \textbf{47.81} & \textbf{37.93} & 18.59 \\
 PyramidAT $m = [24, 12, 1.2]$ & 81.19 & 86.40 & 20.63 & 46.78 & 31.59 & 69.27 & 45.48 & 35.25 & 16.72 \\
 PyramidAT $m = [28, 14, 1.4]$ & 81.14 & 86.31 & 20.68 & 46.71 & 30.99 & 69.61 & 46.23 & 35.86 & 17.50 \\
\end{tabular}
\caption{
Ablation on the magnitude of the pyramid adversarial training.
}
\label{tab:mag_pyramid}
\end{table*}

\section{Additional Analysis}
\label{sup:analysis}
\subsection{Positional embedding}

In Table \ref{tab:positional_embedding}, we explore training on a ViT model without the positional embedding in order to understand the effects of the PixelAT and PyramidAT. We observe that without the positional embedding, PixelAT and PyramidAT tend to perform similarly; in fact, the gap between PixelAT and PyramidAT for clean ImageNet decreases from $1.29$ with the positional embedding to $0.17$ without the positional embedding. This suggests that much of the improvements for in-distribution performance come from improved training of the positional embedding. However, even without the positional embedding, we observe improvements in the out-of-distribution datasets; e.g. going from pixel to pyramid results in a gain of $2.27$ on Rendition and $2.37$ on Sketch with the positional embedding and slightly smaller gains of $1.27$ and $1.43$ without the positional embedding. This suggests that PyramidAT is still improving the learned features used for out-of-distribution performance.

\begin{table*}[!htbp]\centering\small
\begin{tabular}{l|cc|cccc|ccc}
\toprule
&\multicolumn{2}{c|}{}&\multicolumn{7}{c}{Out of Distribution Robustness Test} \\
 Method & ImageNet & Real & A & C$\downarrow$ & ObjectNet & V2 & Rendition & Sketch & Stylized \\
\midrule
 Baseline with positional embedding & 79.92 & 85.14 & 17.48 & 52.46 & 29.30 & 67.49 & 38.24 & 29.08 & 11.02 \\
 Pixel with positional embedding & 80.42 & 85.78 & 19.15 & 47.68 & 30.11 & 68.78 & 45.39 & 34.40 & 18.28 \\
 Pyramid with positional embedding & \textbf{81.71} & \textbf{86.82} & \textbf{22.99} & \textbf{44.99} & \textbf{32.92} & \textbf{70.82} & \textbf{47.66} & \textbf{36.77} & \textbf{19.14} \\
\midrule
 Baseline without positional embedding & 76.58 & 82.04 & 12.76 & 63.56 & 24.15 & 63.26 & 27.99 & 15.79 & 6.25 \\
 Pixel without positional embedding & 78.30 & 83.85 & 14.52 & \textbf{56.86} & 26.19 & 65.50 & 33.09 & 19.93 & \textbf{9.92} \\
 Pyramid without positional embedding & \textbf{78.47} & \textbf{84.28} & \textbf{14.79} & 57.11 & \textbf{27.28} & \textbf{66.21} & \textbf{34.46} & \textbf{21.36} & 9.38 \\
\end{tabular}
\caption{
Analysis of the effect of adversarial training on a ViT without positional embedding. We observe that without the positional embedding, PixelAT and PyramidAT tend to perform similarly for many of the evaluation
datasets.
}
\label{tab:positional_embedding}
\end{table*}

\subsection{Optimizing each level individually}

In the pyramid attack, the different multiplicative magnitudes for each level mean that each level's parameter takes different sized steps; for example, with the default settings, a change of $1$ in the patch parameter leads to a change of $10$ on the final image, whereas a change of $1$ in the pixel parameter leads to a change of $1$. Here, we attempt to understand whether the gradients for the different levels of the pyramid can be informative in the presence of each other; specifically, if the patch level makes a step of $10$ in one direction, will this invalidate the gradient in the pixel level which only makes a step of $1$. To do this, we experiment with running each level of the pyramid separately, going from coarse to fine: for a given $k$, we run $k$ steps of only the coarsest level, $k$ steps of only the next coarsest level, etc. In this experiment, we try to keep the amount of training time roughly equal and select $k$ so that the sum of $k$ on each level is roughly equal to the steps taken in the pyramid method in the main paper (5). Table \ref{tab:coarse_then_fine} shows the results from this experiment and suggests that the gradients from each individual level are still useful when combined and that separating this optimization does not in fact lead to performance improvements; note that $k=2$ leads to more overall optimization steps (6 total steps) than the main technique (5 total steps).

\begin{table*}[!htbp]\centering\small
\begin{tabular}{l|cc|cccc|ccc}
\toprule
&\multicolumn{2}{c|}{}&\multicolumn{7}{c}{Out of Distribution Robustness Test} \\
 Method & ImageNet & Real & A & C$\downarrow$ & ObjectNet & V2 & Rendition & Sketch & Stylized \\
\midrule
 Pyramid Separate $k=1$ (3 total) & 81.17 & 86.06 & 19.41 & 49.10 & 30.35 & 69.27 & 42.94 & 33.41 & 15.55 \\
 Pyramid Separate $k=2$ (6 total) & 81.36 & 86.36 & 22.31 & 47.73 & 32.12 & 69.92 & 46.13 & 35.66 & 15.31 \\
 Pyramid (5 steps total) & \textbf{81.71} & \textbf{86.82} & \textbf{22.99} & \textbf{44.99} & \textbf{32.92} & \textbf{70.82} & \textbf{47.66} & \textbf{36.77} & \textbf{19.14} \\
\end{tabular}
\caption{
Ablation on CoarseThenFine. The separated gradients do not seem strictly better than simply running all levels at once.
}
\label{tab:coarse_then_fine}
\end{table*}

\subsection{Evaluation of white-box attacks}

We evaluate the performance of the B/16 baseline, PixelAT, and PyramidAT models against pixel and pyramid PGD attacks. The results are given in table \ref{tab:white_box_attacks}. Both adversarially trained models give the best performance when attacked in the setting in which they were trained. PyramidAT provides comparably more protection against pixel attacks (48.8\% Top-1) than PixelAT against pyramid attacks (43.0\% Top-1).

\begin{table*}[!htbp]\centering\small
\begin{tabular}{l|cc}
\toprule
 Method & Pixel PGD & Pyramid PGD \\
\midrule
 Baseline & 13.7 & 26.1 \\
 +PixelAT & 53.6 & 43.0 \\
 +PyramidAT & 48.8 & 68.7 \\
\end{tabular}
\caption{
Top-1 accuracy of 3 models (baseline, PixelAT, PyramidAT) against 2 types of white-box adversarial attacks (Pixel PGD, Pyramid PGD, both at 5 steps).
}
\label{tab:white_box_attacks}
\end{table*}

\section{Additional Visualizations}
\label{sup:viz}
\subsection{Pixel attacks}

\begin{figure}\centering
\setlength{\tabcolsep}{.12em}
\begin{tabular}{cccc}
\includegraphics[width=0.24\columnwidth]{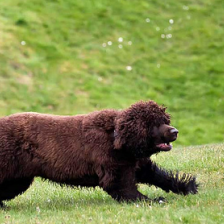} &
\includegraphics[width=0.24\columnwidth]{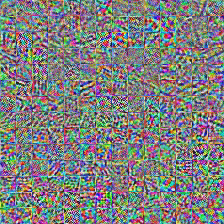} &
\includegraphics[width=0.24\columnwidth]{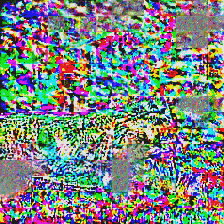} &
\includegraphics[width=0.24\columnwidth]{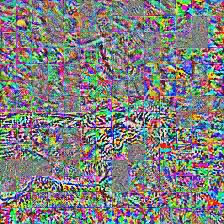} \\
\includegraphics[width=0.24\columnwidth]{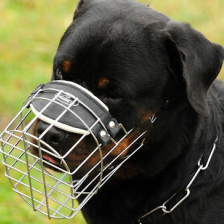} &
\includegraphics[width=0.24\columnwidth]{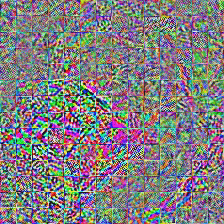} &
\includegraphics[width=0.24\columnwidth]{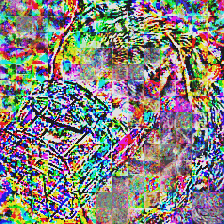} &
\includegraphics[width=0.24\columnwidth]{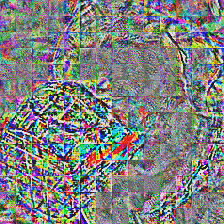} \\
\includegraphics[width=0.24\columnwidth]{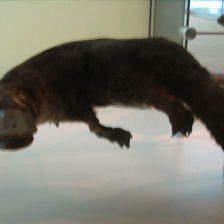} &
\includegraphics[width=0.24\columnwidth]{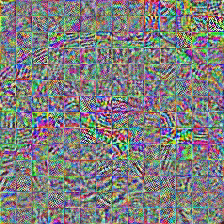} &
\includegraphics[width=0.24\columnwidth]{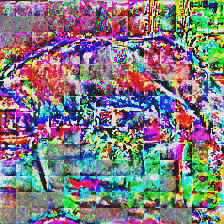} &
\includegraphics[width=0.24\columnwidth]{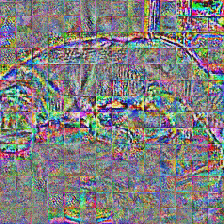} \\
\includegraphics[width=0.24\columnwidth]{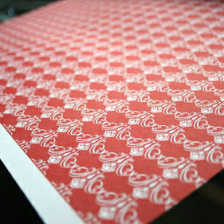} &
\includegraphics[width=0.24\columnwidth]{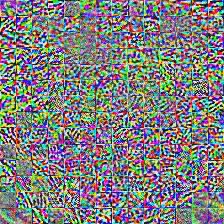} &
\includegraphics[width=0.24\columnwidth]{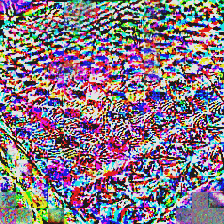} &
\includegraphics[width=0.24\columnwidth]{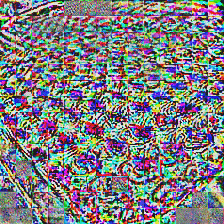} \\
Original & Baseline & PixelAT & PyramidAT \\
\end{tabular}
\caption{Visualizations of pixel attacks on different pre-trainings: baseline, PixelAT, and PyramidAT. Note that PixelAT models should have better defense against
pixel attacks than PyramidAT models since the attack is in-distribution to the train data.}
\label{fig:pixel_attacks}
\end{figure}

In Figure \ref{fig:pixel_attacks}, we include 4 additional visualizations of pixel attacks against the baseline and PixelAT models. Some structure is visible in the PixelAT model. Note that for pixel attacks, we would expect more structure to appear in the PixelAT model than the PyramidAT model since the attack is in-distribution for the PixelAT model but out-of-distribution for the PyramidAT.

\subsection{Pyramid attacks}
\begin{figure}\centering
\setlength{\tabcolsep}{.12em}
\begin{tabular}{ccccc}
\includegraphics[width=0.2\columnwidth]{SupplementalFigures/attack_viz/original_0.png} &
\includegraphics[width=0.2\columnwidth]{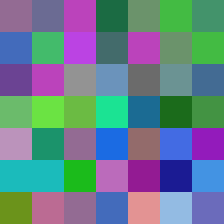} &
\includegraphics[width=0.2\columnwidth]{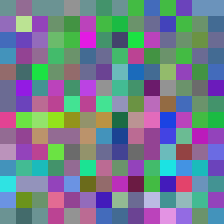} &
\includegraphics[width=0.2\columnwidth]{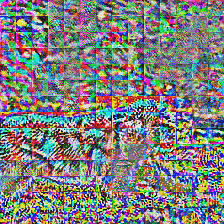} &
\includegraphics[width=0.2\columnwidth]{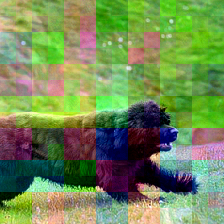}  \\
\includegraphics[width=0.2\columnwidth]{SupplementalFigures/attack_viz/original_1.png} &
\includegraphics[width=0.2\columnwidth]{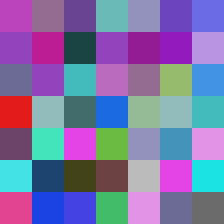} &
\includegraphics[width=0.2\columnwidth]{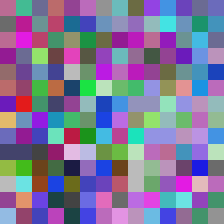} &
\includegraphics[width=0.2\columnwidth]{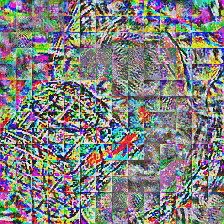} &
\includegraphics[width=0.2\columnwidth]{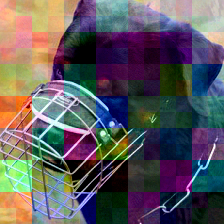}  \\
\includegraphics[width=0.2\columnwidth]{SupplementalFigures/attack_viz/original_2.png} &
\includegraphics[width=0.2\columnwidth]{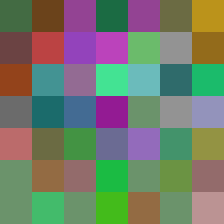} &
\includegraphics[width=0.2\columnwidth]{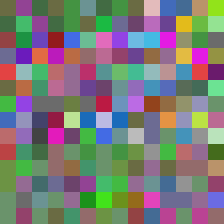} &
\includegraphics[width=0.2\columnwidth]{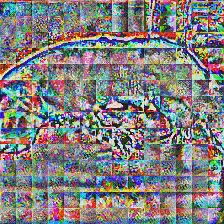} &
\includegraphics[width=0.2\columnwidth]{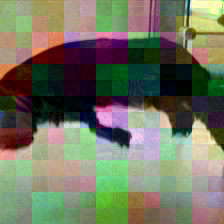}  \\
\includegraphics[width=0.2\columnwidth]{SupplementalFigures/attack_viz/original_3.png} &
\includegraphics[width=0.2\columnwidth]{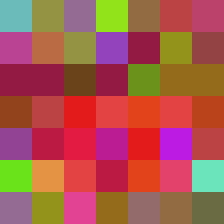} &
\includegraphics[width=0.2\columnwidth]{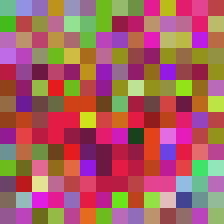} &
\includegraphics[width=0.2\columnwidth]{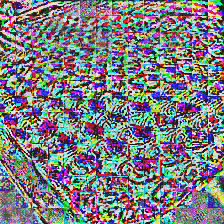} &
\includegraphics[width=0.2\columnwidth]{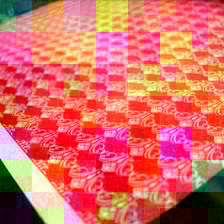}  \\
Original & Top layer & Middle layer & Pixel layer & Perturbed Image \\
\end{tabular}
\caption{Visualizations of pyramid attacks on pyramid-trained model.}
\label{fig:pyramid_attacks}
\end{figure}

In Figure \ref{fig:pyramid_attacks}, we include 4 additional visualizations of pyramid attacks against the PyramidAT models. Note that in the finest level, more structure is visible.

\subsection{Attention}

\begin{figure}\centering
\setlength{\tabcolsep}{.12em}
\begin{tabular}{ccc}

\includegraphics[width=0.25\columnwidth]{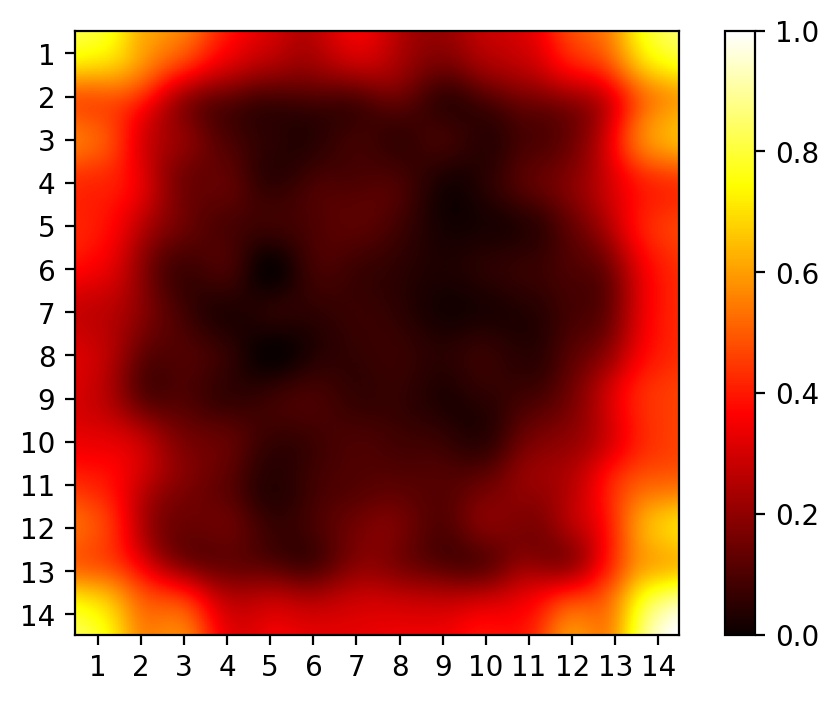} &
\includegraphics[width=0.25\columnwidth]{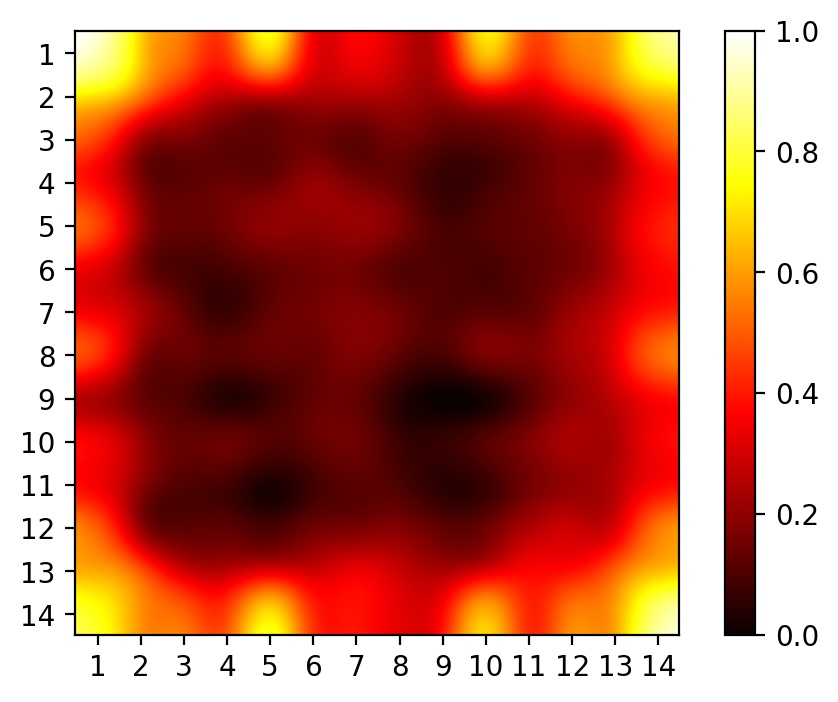} &
\includegraphics[width=0.25\columnwidth]{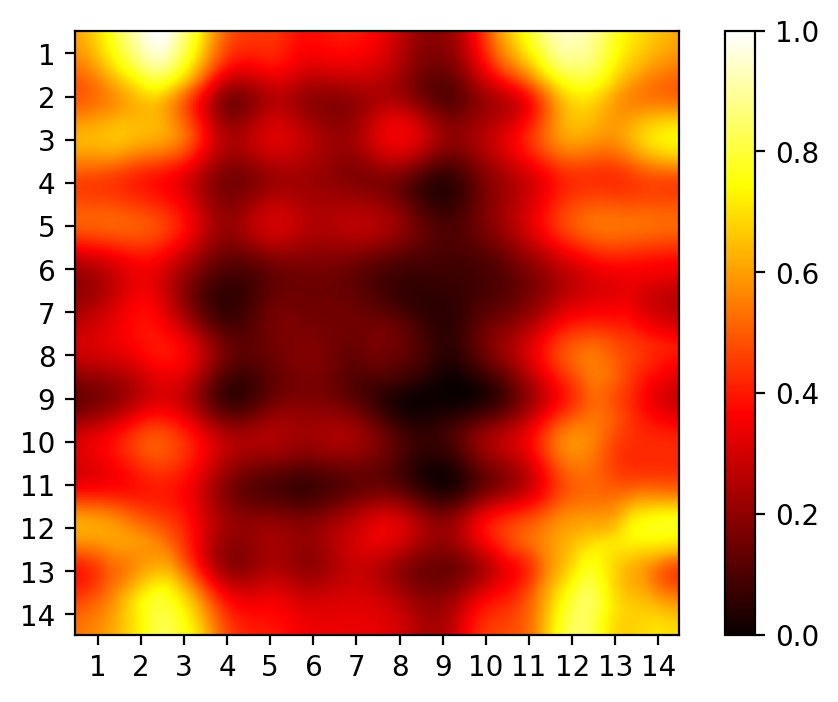} \\

Baseline & Pixel & Pyramid \\
\end{tabular}
\caption{Visualizations of the average attention for different pre-trainings. Examples on dataset ImageNet.}
\label{fig:avg_attn_viz_examples_dataset_imagenet}
\end{figure}

\begin{figure}\centering
\setlength{\tabcolsep}{.12em}
\begin{tabular}{ccc}

\includegraphics[width=0.25\columnwidth]{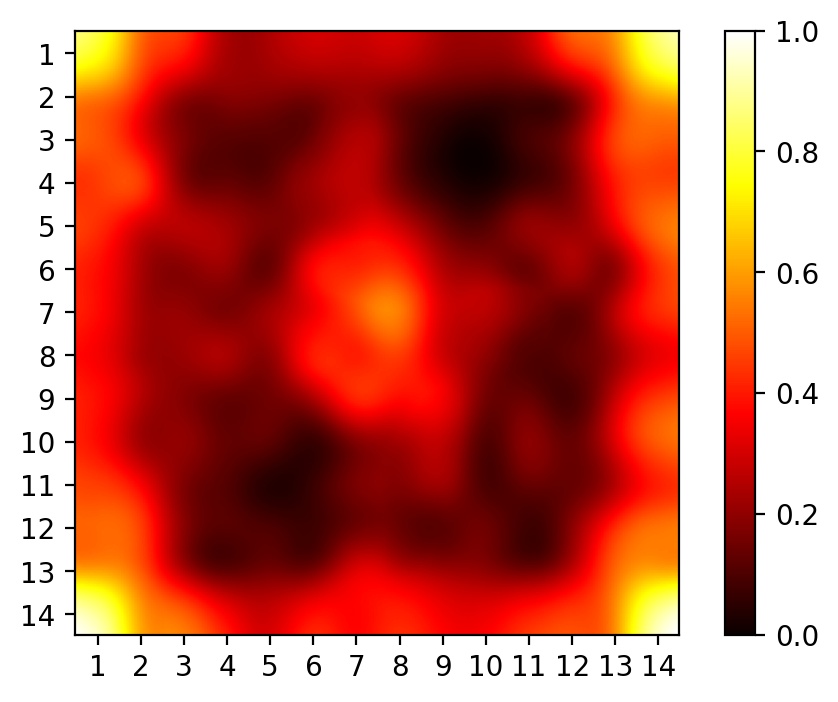} &
\includegraphics[width=0.25\columnwidth]{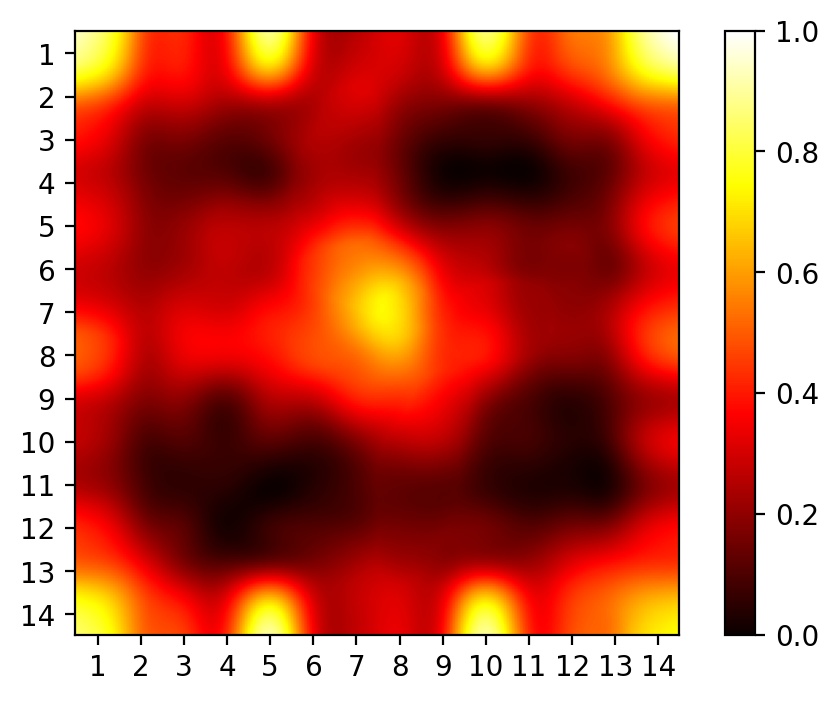} &
\includegraphics[width=0.25\columnwidth]{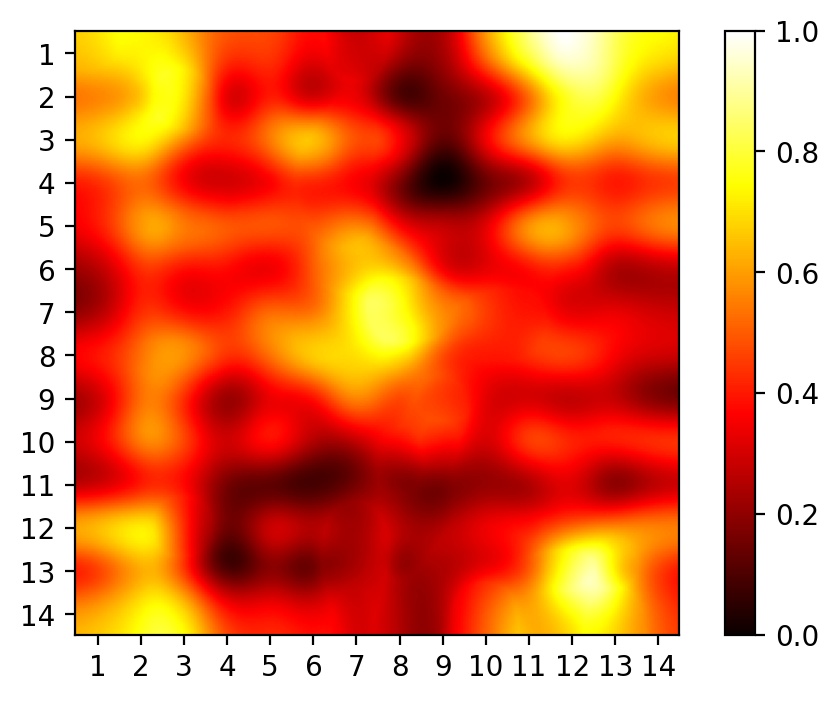} \\

Baseline & PixelAT & PyramidAT \\
\end{tabular}
\caption{Visualizations of the average attention for different pre-trainings. Examples on dataset ImageNet-A.}
\label{fig:avg_attn_viz_examples_dataset_imagenet_a}
\end{figure}

\begin{figure}\centering
\setlength{\tabcolsep}{.12em}
\begin{tabular}{ccc}

\includegraphics[width=0.25\columnwidth]{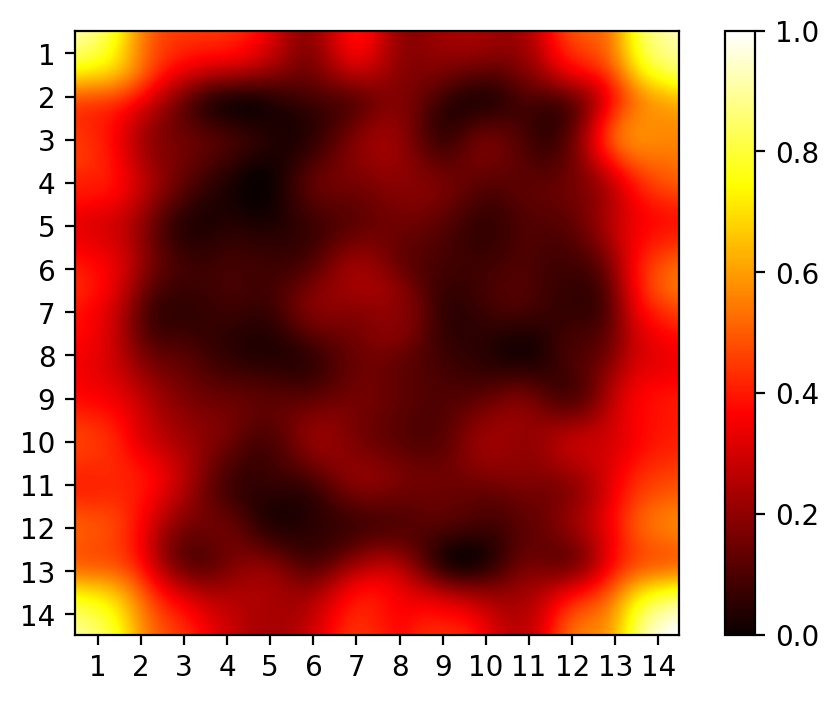} &
\includegraphics[width=0.25\columnwidth]{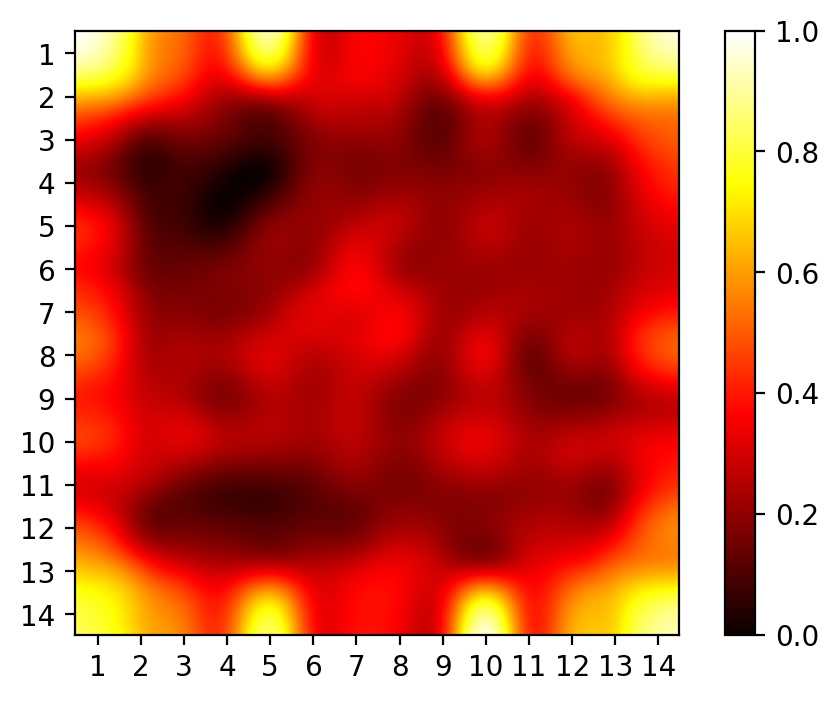} &
\includegraphics[width=0.25\columnwidth]{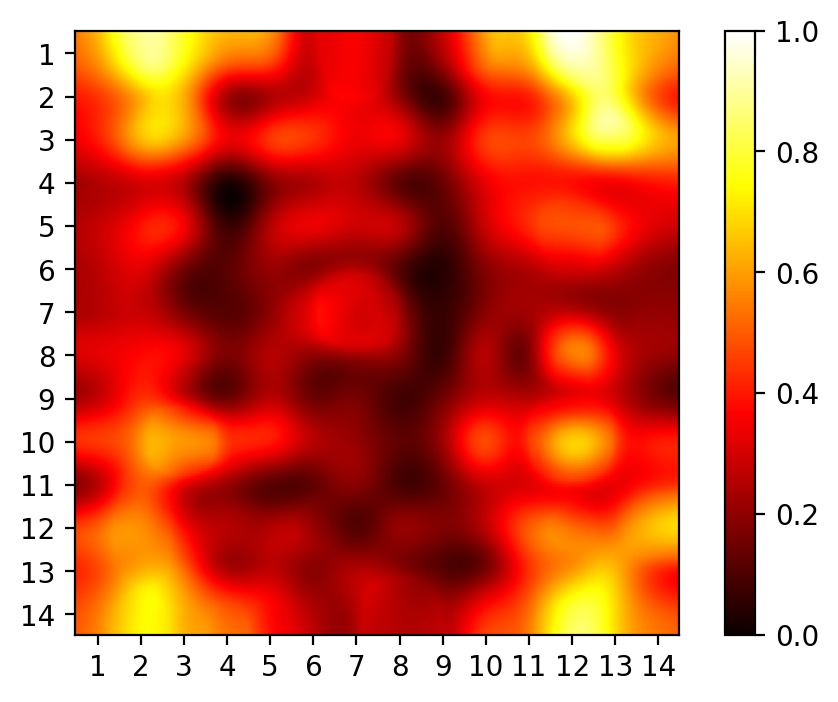} \\

Baseline & PixelAT & PyramidAT \\
\end{tabular}
\caption{Visualizations of the average attention for different pre-trainings. Examples on dataset ImageNet2012-ReaL.}
\label{fig:avg_attn_viz_examples_dataset_imagenet2012_real}
\end{figure}

\begin{figure}\centering
\setlength{\tabcolsep}{.12em}
\begin{tabular}{ccc}

\includegraphics[width=0.25\columnwidth]{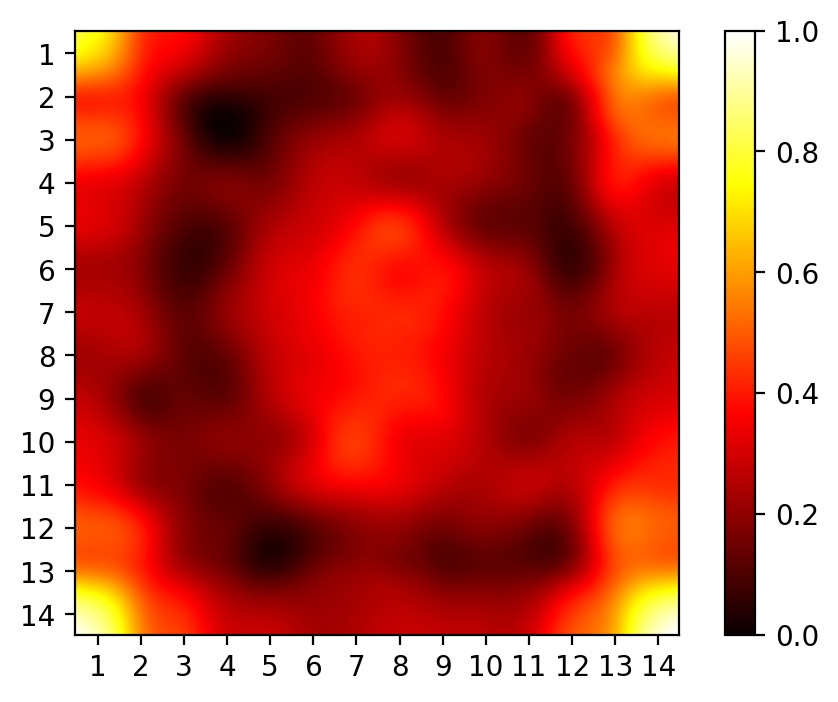} &
\includegraphics[width=0.25\columnwidth]{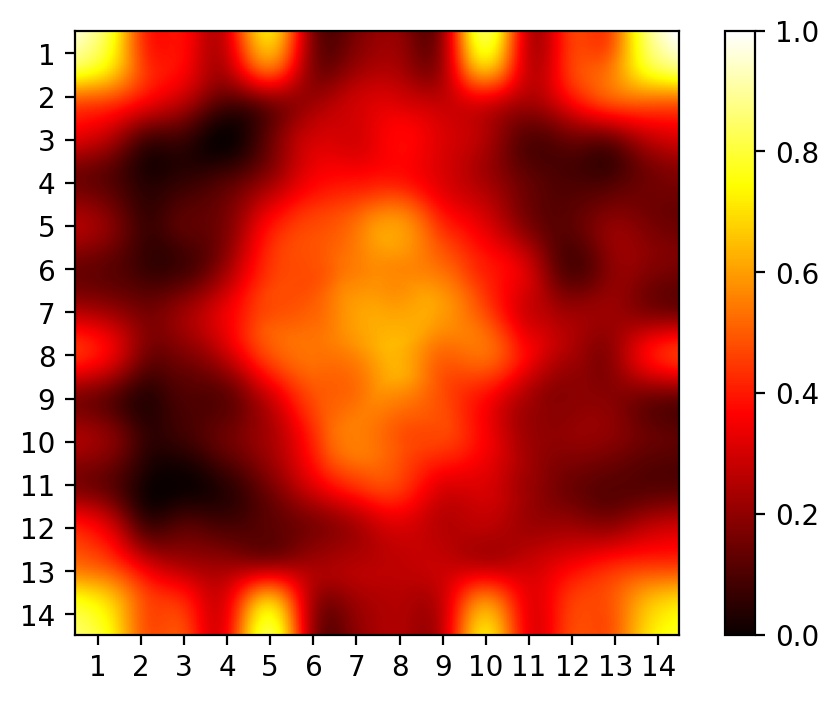} &
\includegraphics[width=0.25\columnwidth]{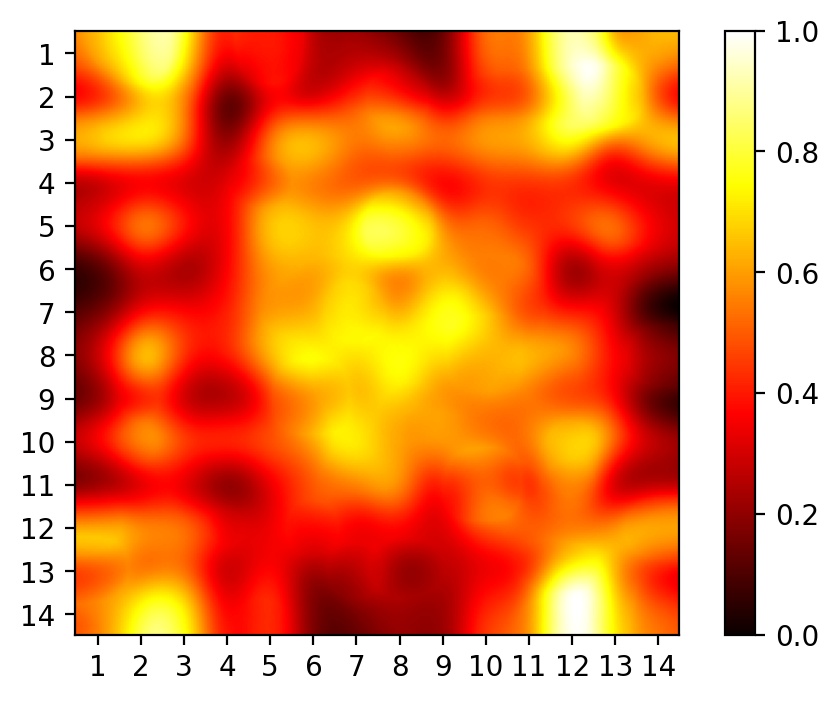} \\

Baseline & PixelAT & PyramidAT \\
\end{tabular}
\caption{Visualizations of the average attention for different pre-trainings. Examples on dataset ImageNet-Rendition.}
\label{fig:avg_attn_viz_examples_dataset_imagenet_r}
\end{figure}

\begin{figure}\centering
\setlength{\tabcolsep}{.12em}
\begin{tabular}{ccc}

\includegraphics[width=0.25\columnwidth]{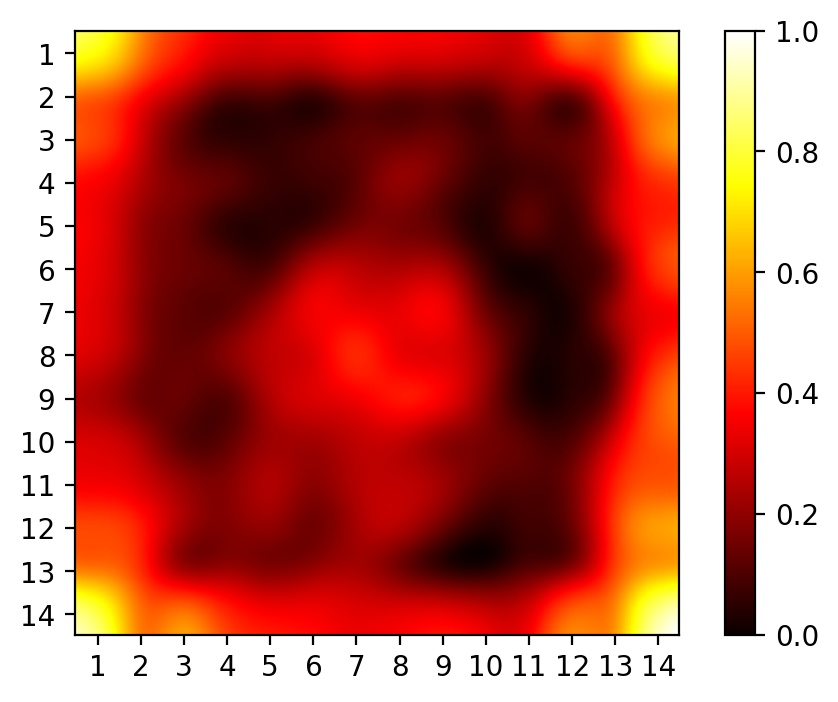} &
\includegraphics[width=0.25\columnwidth]{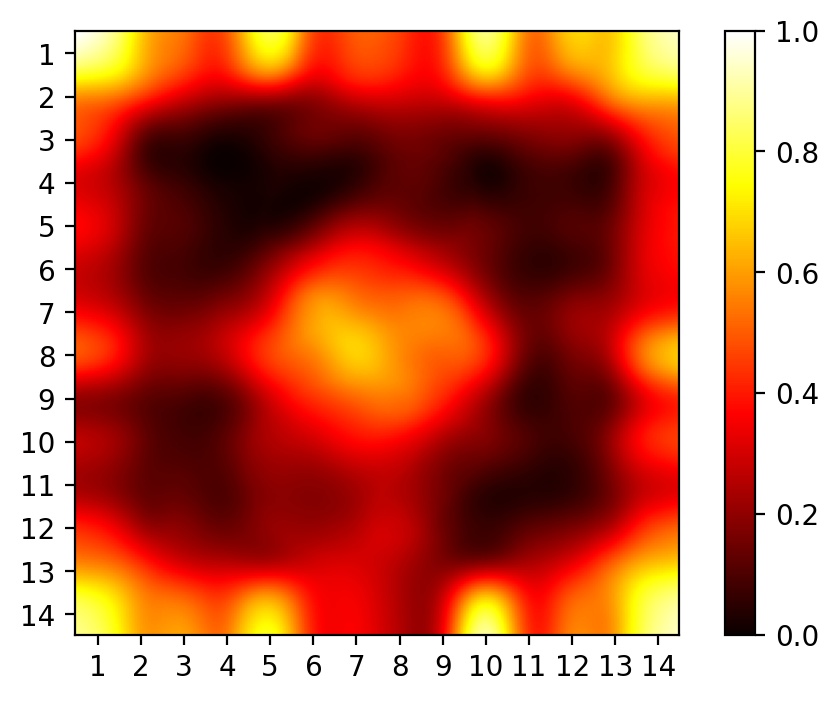} &
\includegraphics[width=0.25\columnwidth]{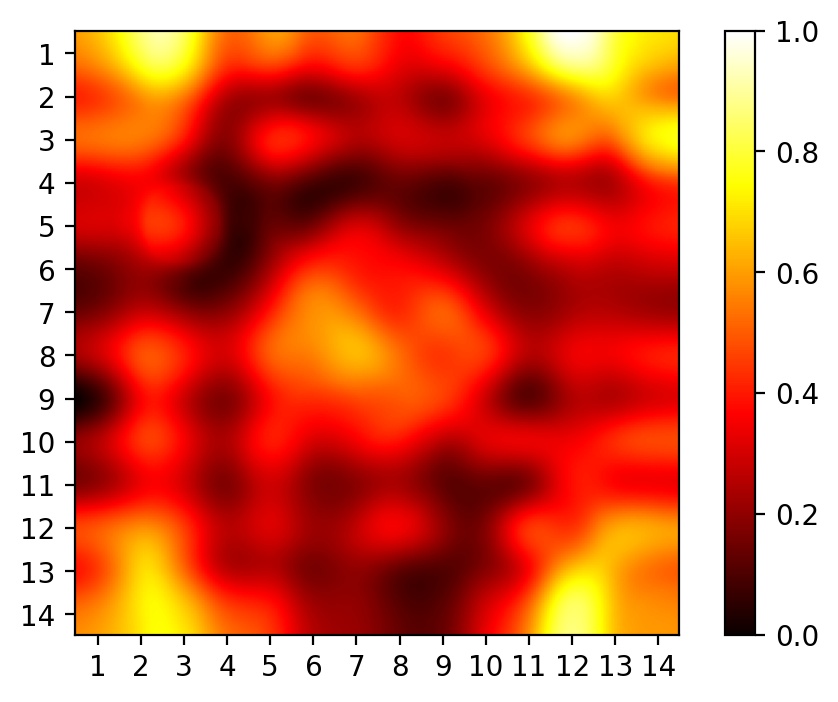} \\

Baseline & PixelAT & PyramidAT \\
\end{tabular}
\caption{Visualizations of the average attention for different pre-trainings. Examples on dataset ObjectNet.}
\label{fig:avg_attn_viz_examples_dataset_objectnet}
\end{figure}

\begin{figure}\centering
\setlength{\tabcolsep}{.12em}
\begin{tabular}{ccc}

\includegraphics[width=0.25\columnwidth]{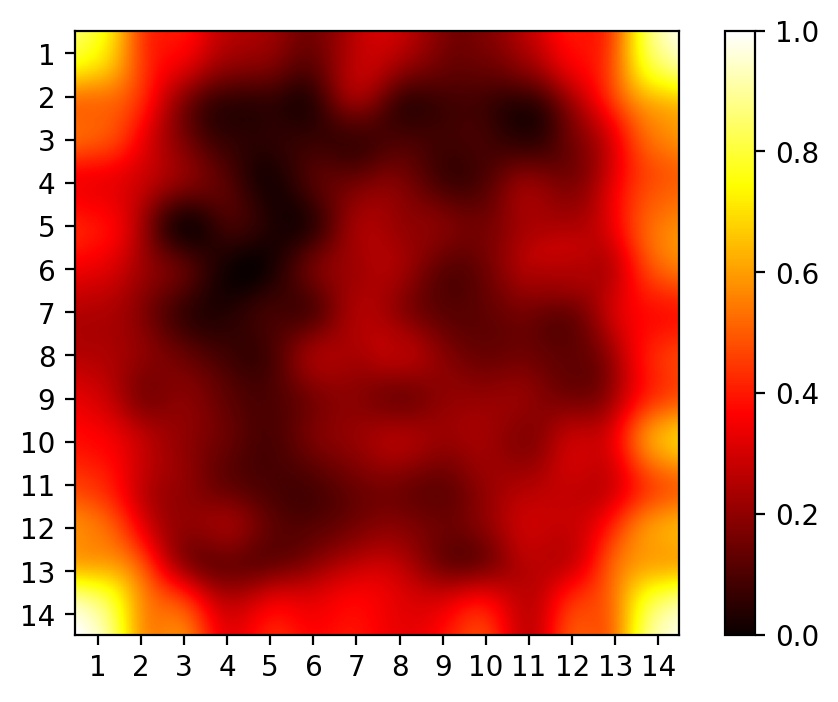} &
\includegraphics[width=0.25\columnwidth]{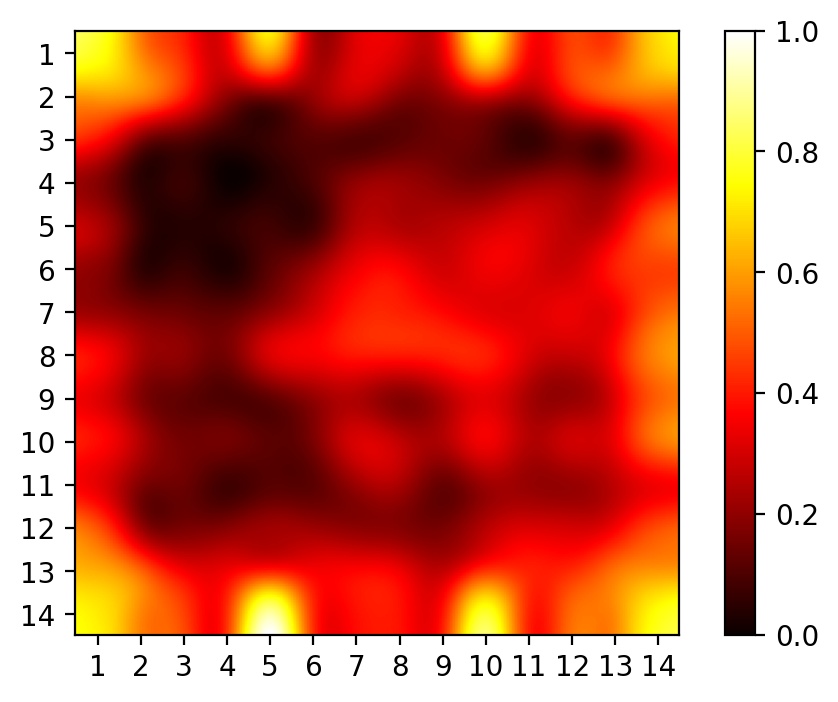} &
\includegraphics[width=0.25\columnwidth]{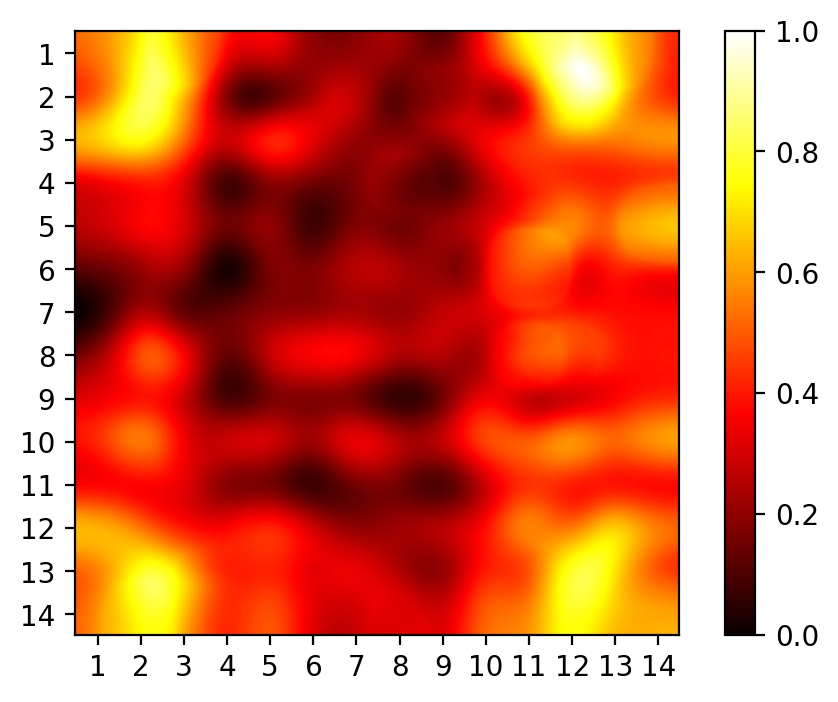} \\

Baseline & Pixel & Pyramid \\
\end{tabular}
\caption{Visualizations of the average attention for different pre-trainings. Examples on dataset StylizedImageNet.}
\label{fig:avg_attn_viz_examples_dataset_stylized_imagenet}
\end{figure}

We include the average attentions of baseline, PixelAT, and PyramidAT on the following datasets: ImageNet (Figure \ref{fig:avg_attn_viz_examples_dataset_imagenet}), ImageNet-A (Figure \ref{fig:avg_attn_viz_examples_dataset_imagenet_a}), ImageNet-ReaL (Figure \ref{fig:avg_attn_viz_examples_dataset_imagenet2012_real}), ImageNet-Rendition (Figure \ref{fig:avg_attn_viz_examples_dataset_imagenet_r}), ObjectNet (Figure \ref{fig:avg_attn_viz_examples_dataset_objectnet}), and Stylized ImageNet (Figure \ref{fig:avg_attn_viz_examples_dataset_stylized_imagenet}). The trend, as stated in the main paper, (PixelAT tightly focusing on the center and PyramidAT taking a more global perspective) stays consistent across the various evaluation datasets.

\begin{figure}
\setlength{\tabcolsep}{.12em}
\begin{tabular}{cccccccc}

\includegraphics[width=0.125\columnwidth]{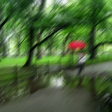} &
\includegraphics[width=0.125\columnwidth]{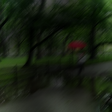} &
\includegraphics[width=0.125\columnwidth]{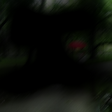} &
\includegraphics[width=0.125\columnwidth]{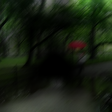} &
\includegraphics[width=0.125\columnwidth]{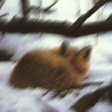} &
\includegraphics[width=0.125\columnwidth]{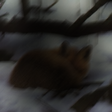} &
\includegraphics[width=0.125\columnwidth]{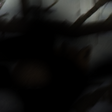} &
\includegraphics[width=0.125\columnwidth]{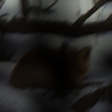} \\

\includegraphics[width=0.125\columnwidth]{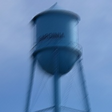} &
\includegraphics[width=0.125\columnwidth]{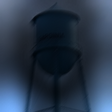} &
\includegraphics[width=0.125\columnwidth]{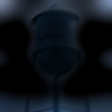} &
\includegraphics[width=0.125\columnwidth]{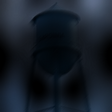} &
\includegraphics[width=0.125\columnwidth]{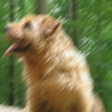} &
\includegraphics[width=0.125\columnwidth]{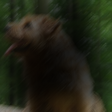} &
\includegraphics[width=0.125\columnwidth]{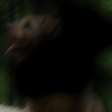} &
\includegraphics[width=0.125\columnwidth]{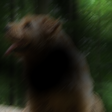} \\

\includegraphics[width=0.125\columnwidth]{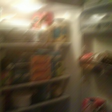} &
\includegraphics[width=0.125\columnwidth]{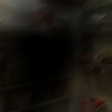} &
\includegraphics[width=0.125\columnwidth]{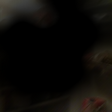} &
\includegraphics[width=0.125\columnwidth]{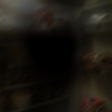} &
\includegraphics[width=0.125\columnwidth]{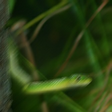} &
\includegraphics[width=0.125\columnwidth]{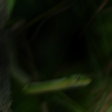} &
\includegraphics[width=0.125\columnwidth]{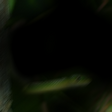} &
\includegraphics[width=0.125\columnwidth]{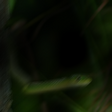} \\

\includegraphics[width=0.125\columnwidth]{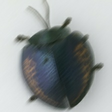} &
\includegraphics[width=0.125\columnwidth]{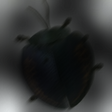} &
\includegraphics[width=0.125\columnwidth]{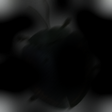} &
\includegraphics[width=0.125\columnwidth]{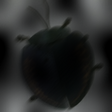} &
\includegraphics[width=0.125\columnwidth]{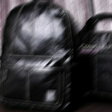} &
\includegraphics[width=0.125\columnwidth]{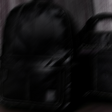} &
\includegraphics[width=0.125\columnwidth]{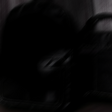} &
\includegraphics[width=0.125\columnwidth]{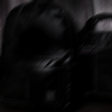} \\

\includegraphics[width=0.125\columnwidth]{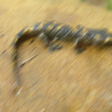} &
\includegraphics[width=0.125\columnwidth]{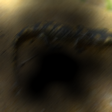} &
\includegraphics[width=0.125\columnwidth]{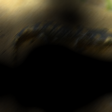} &
\includegraphics[width=0.125\columnwidth]{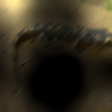} &
\includegraphics[width=0.125\columnwidth]{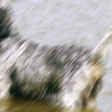} &
\includegraphics[width=0.125\columnwidth]{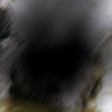} &
\includegraphics[width=0.125\columnwidth]{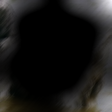} &
\includegraphics[width=0.125\columnwidth]{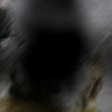} \\

\includegraphics[width=0.125\columnwidth]{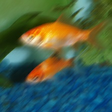} &
\includegraphics[width=0.125\columnwidth]{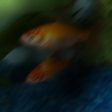} &
\includegraphics[width=0.125\columnwidth]{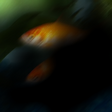} &
\includegraphics[width=0.125\columnwidth]{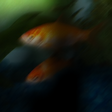} &
\includegraphics[width=0.125\columnwidth]{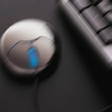} &
\includegraphics[width=0.125\columnwidth]{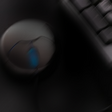} &
\includegraphics[width=0.125\columnwidth]{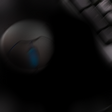} &
\includegraphics[width=0.125\columnwidth]{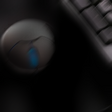} \\

\includegraphics[width=0.125\columnwidth]{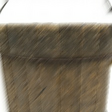} &
\includegraphics[width=0.125\columnwidth]{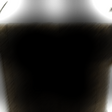} &
\includegraphics[width=0.125\columnwidth]{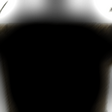} &
\includegraphics[width=0.125\columnwidth]{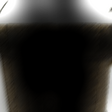} &
\includegraphics[width=0.125\columnwidth]{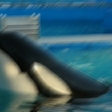} &
\includegraphics[width=0.125\columnwidth]{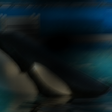} &
\includegraphics[width=0.125\columnwidth]{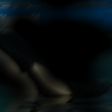} &
\includegraphics[width=0.125\columnwidth]{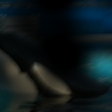} \\

\includegraphics[width=0.125\columnwidth]{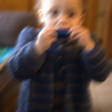} &
\includegraphics[width=0.125\columnwidth]{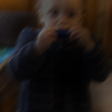} &
\includegraphics[width=0.125\columnwidth]{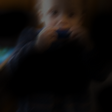} &
\includegraphics[width=0.125\columnwidth]{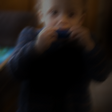} &
\includegraphics[width=0.125\columnwidth]{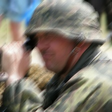} &
\includegraphics[width=0.125\columnwidth]{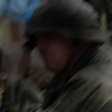} &
\includegraphics[width=0.125\columnwidth]{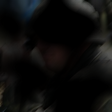} &
\includegraphics[width=0.125\columnwidth]{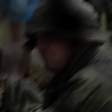} \\

Original & Baseline & PixelAT & PyramidAT & Original & Baseline & PixelAT & PyramidAT \\
\end{tabular}
\caption{Visualizations of the attention for different pre-trainings. Examples on dataset ImageNet.}
\label{fig:attn_viz_examples_dataset_imagenet}
\end{figure}

\begin{figure}
\setlength{\tabcolsep}{.12em}
\begin{tabular}{cccccccc}

\includegraphics[width=0.125\columnwidth]{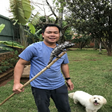} &
\includegraphics[width=0.125\columnwidth]{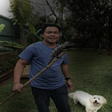} &
\includegraphics[width=0.125\columnwidth]{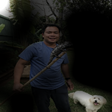} &
\includegraphics[width=0.125\columnwidth]{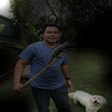} &
\includegraphics[width=0.125\columnwidth]{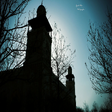} &
\includegraphics[width=0.125\columnwidth]{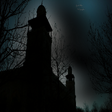} &
\includegraphics[width=0.125\columnwidth]{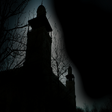} &
\includegraphics[width=0.125\columnwidth]{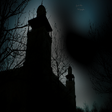} \\

\includegraphics[width=0.125\columnwidth]{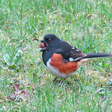} &
\includegraphics[width=0.125\columnwidth]{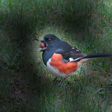} &
\includegraphics[width=0.125\columnwidth]{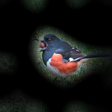} &
\includegraphics[width=0.125\columnwidth]{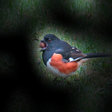} &
\includegraphics[width=0.125\columnwidth]{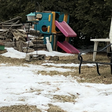} &
\includegraphics[width=0.125\columnwidth]{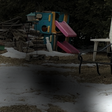} &
\includegraphics[width=0.125\columnwidth]{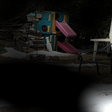} &
\includegraphics[width=0.125\columnwidth]{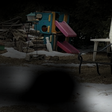} \\

\includegraphics[width=0.125\columnwidth]{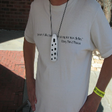} &
\includegraphics[width=0.125\columnwidth]{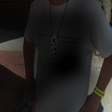} &
\includegraphics[width=0.125\columnwidth]{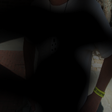} &
\includegraphics[width=0.125\columnwidth]{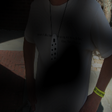} &
\includegraphics[width=0.125\columnwidth]{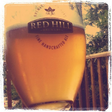} &
\includegraphics[width=0.125\columnwidth]{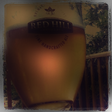} &
\includegraphics[width=0.125\columnwidth]{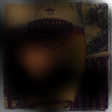} &
\includegraphics[width=0.125\columnwidth]{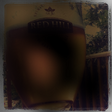} \\

\includegraphics[width=0.125\columnwidth]{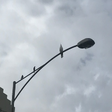} &
\includegraphics[width=0.125\columnwidth]{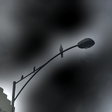} &
\includegraphics[width=0.125\columnwidth]{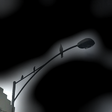} &
\includegraphics[width=0.125\columnwidth]{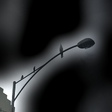} &
\includegraphics[width=0.125\columnwidth]{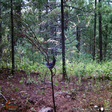} &
\includegraphics[width=0.125\columnwidth]{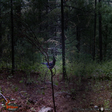} &
\includegraphics[width=0.125\columnwidth]{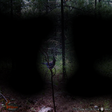} &
\includegraphics[width=0.125\columnwidth]{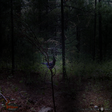} \\

\includegraphics[width=0.125\columnwidth]{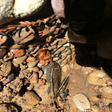} &
\includegraphics[width=0.125\columnwidth]{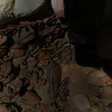} &
\includegraphics[width=0.125\columnwidth]{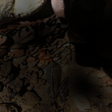} &
\includegraphics[width=0.125\columnwidth]{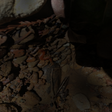} &
\includegraphics[width=0.125\columnwidth]{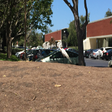} &
\includegraphics[width=0.125\columnwidth]{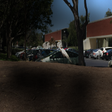} &
\includegraphics[width=0.125\columnwidth]{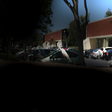} &
\includegraphics[width=0.125\columnwidth]{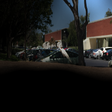} \\

\includegraphics[width=0.125\columnwidth]{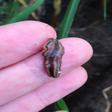} &
\includegraphics[width=0.125\columnwidth]{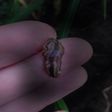} &
\includegraphics[width=0.125\columnwidth]{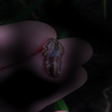} &
\includegraphics[width=0.125\columnwidth]{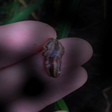} &
\includegraphics[width=0.125\columnwidth]{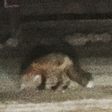} &
\includegraphics[width=0.125\columnwidth]{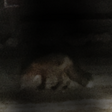} &
\includegraphics[width=0.125\columnwidth]{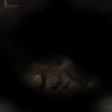} &
\includegraphics[width=0.125\columnwidth]{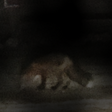} \\

\includegraphics[width=0.125\columnwidth]{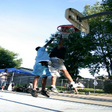} &
\includegraphics[width=0.125\columnwidth]{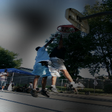} &
\includegraphics[width=0.125\columnwidth]{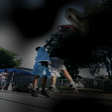} &
\includegraphics[width=0.125\columnwidth]{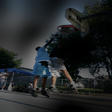} &
\includegraphics[width=0.125\columnwidth]{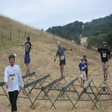} &
\includegraphics[width=0.125\columnwidth]{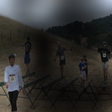} &
\includegraphics[width=0.125\columnwidth]{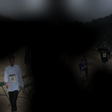} &
\includegraphics[width=0.125\columnwidth]{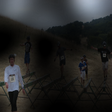} \\

\includegraphics[width=0.125\columnwidth]{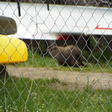} &
\includegraphics[width=0.125\columnwidth]{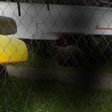} &
\includegraphics[width=0.125\columnwidth]{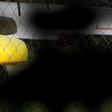} &
\includegraphics[width=0.125\columnwidth]{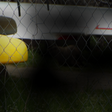} &
\includegraphics[width=0.125\columnwidth]{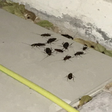} &
\includegraphics[width=0.125\columnwidth]{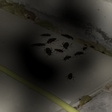} &
\includegraphics[width=0.125\columnwidth]{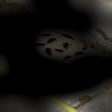} &
\includegraphics[width=0.125\columnwidth]{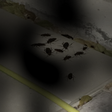} \\

Original & Baseline & PixelAT & PyramidAT & Original & Baseline & PixelAT & PyramidAT \\
\end{tabular}
\caption{Visualizations of the attention for different pre-trainings. Examples on dataset ImageNet-A.}
\label{fig:attn_viz_examples_dataset_imagenet_a}
\end{figure}

\begin{figure}
\setlength{\tabcolsep}{.12em}
\begin{tabular}{cccccccc}

\includegraphics[width=0.125\columnwidth]{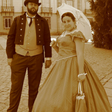} &
\includegraphics[width=0.125\columnwidth]{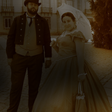} &
\includegraphics[width=0.125\columnwidth]{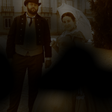} &
\includegraphics[width=0.125\columnwidth]{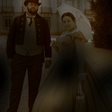} &
\includegraphics[width=0.125\columnwidth]{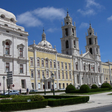} &
\includegraphics[width=0.125\columnwidth]{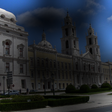} &
\includegraphics[width=0.125\columnwidth]{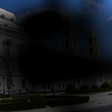} &
\includegraphics[width=0.125\columnwidth]{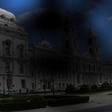} \\

\includegraphics[width=0.125\columnwidth]{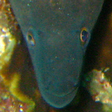} &
\includegraphics[width=0.125\columnwidth]{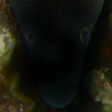} &
\includegraphics[width=0.125\columnwidth]{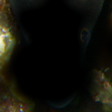} &
\includegraphics[width=0.125\columnwidth]{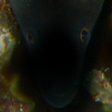} &
\includegraphics[width=0.125\columnwidth]{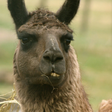} &
\includegraphics[width=0.125\columnwidth]{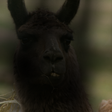} &
\includegraphics[width=0.125\columnwidth]{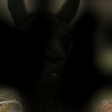} &
\includegraphics[width=0.125\columnwidth]{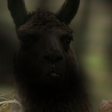} \\

\includegraphics[width=0.125\columnwidth]{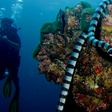} &
\includegraphics[width=0.125\columnwidth]{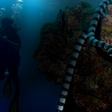} &
\includegraphics[width=0.125\columnwidth]{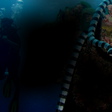} &
\includegraphics[width=0.125\columnwidth]{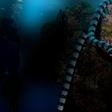} &
\includegraphics[width=0.125\columnwidth]{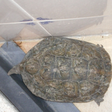} &
\includegraphics[width=0.125\columnwidth]{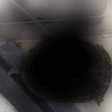} &
\includegraphics[width=0.125\columnwidth]{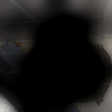} &
\includegraphics[width=0.125\columnwidth]{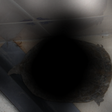} \\

\includegraphics[width=0.125\columnwidth]{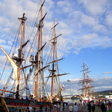} &
\includegraphics[width=0.125\columnwidth]{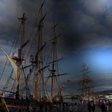} &
\includegraphics[width=0.125\columnwidth]{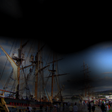} &
\includegraphics[width=0.125\columnwidth]{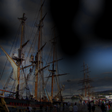} &
\includegraphics[width=0.125\columnwidth]{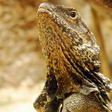} &
\includegraphics[width=0.125\columnwidth]{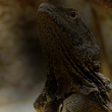} &
\includegraphics[width=0.125\columnwidth]{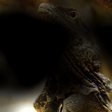} &
\includegraphics[width=0.125\columnwidth]{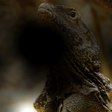} \\

\includegraphics[width=0.125\columnwidth]{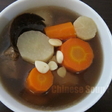} &
\includegraphics[width=0.125\columnwidth]{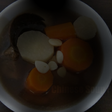} &
\includegraphics[width=0.125\columnwidth]{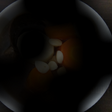} &
\includegraphics[width=0.125\columnwidth]{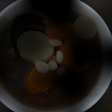} &
\includegraphics[width=0.125\columnwidth]{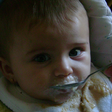} &
\includegraphics[width=0.125\columnwidth]{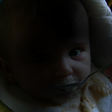} &
\includegraphics[width=0.125\columnwidth]{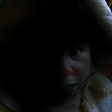} &
\includegraphics[width=0.125\columnwidth]{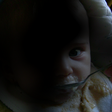} \\

\includegraphics[width=0.125\columnwidth]{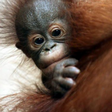} &
\includegraphics[width=0.125\columnwidth]{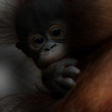} &
\includegraphics[width=0.125\columnwidth]{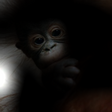} &
\includegraphics[width=0.125\columnwidth]{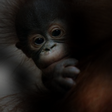} &
\includegraphics[width=0.125\columnwidth]{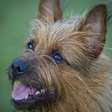} &
\includegraphics[width=0.125\columnwidth]{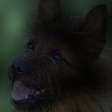} &
\includegraphics[width=0.125\columnwidth]{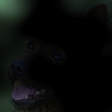} &
\includegraphics[width=0.125\columnwidth]{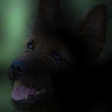} \\

\includegraphics[width=0.125\columnwidth]{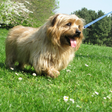} &
\includegraphics[width=0.125\columnwidth]{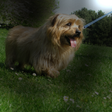} &
\includegraphics[width=0.125\columnwidth]{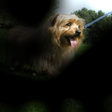} &
\includegraphics[width=0.125\columnwidth]{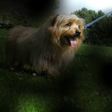} &
\includegraphics[width=0.125\columnwidth]{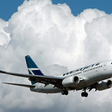} &
\includegraphics[width=0.125\columnwidth]{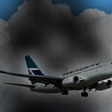} &
\includegraphics[width=0.125\columnwidth]{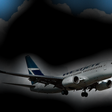} &
\includegraphics[width=0.125\columnwidth]{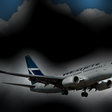} \\

\includegraphics[width=0.125\columnwidth]{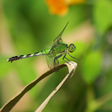} &
\includegraphics[width=0.125\columnwidth]{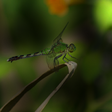} &
\includegraphics[width=0.125\columnwidth]{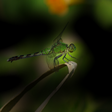} &
\includegraphics[width=0.125\columnwidth]{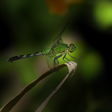} &
\includegraphics[width=0.125\columnwidth]{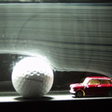} &
\includegraphics[width=0.125\columnwidth]{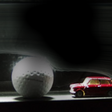} &
\includegraphics[width=0.125\columnwidth]{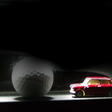} &
\includegraphics[width=0.125\columnwidth]{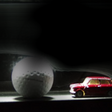} \\

Original & Baseline & PixelAT & PyramidAT & Original & Baseline & PixelAT & PyramidAT \\
\end{tabular}
\caption{Visualizations of the attention for different pre-trainings. Examples on dataset ImageNet2012-ReaL.}
\label{fig:attn_viz_examples_dataset_imagenet2012_real}
\end{figure}

\begin{figure}
\setlength{\tabcolsep}{.12em}
\begin{tabular}{cccccccc}

\includegraphics[width=0.125\columnwidth]{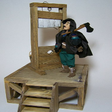} &
\includegraphics[width=0.125\columnwidth]{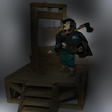} &
\includegraphics[width=0.125\columnwidth]{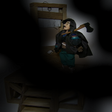} &
\includegraphics[width=0.125\columnwidth]{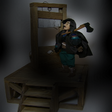} &
\includegraphics[width=0.125\columnwidth]{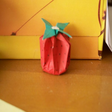} &
\includegraphics[width=0.125\columnwidth]{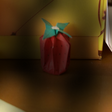} &
\includegraphics[width=0.125\columnwidth]{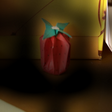} &
\includegraphics[width=0.125\columnwidth]{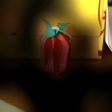} \\

\includegraphics[width=0.125\columnwidth]{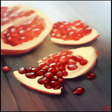} &
\includegraphics[width=0.125\columnwidth]{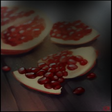} &
\includegraphics[width=0.125\columnwidth]{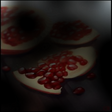} &
\includegraphics[width=0.125\columnwidth]{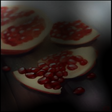} &
\includegraphics[width=0.125\columnwidth]{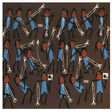} &
\includegraphics[width=0.125\columnwidth]{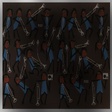} &
\includegraphics[width=0.125\columnwidth]{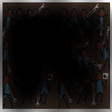} &
\includegraphics[width=0.125\columnwidth]{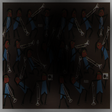} \\

\includegraphics[width=0.125\columnwidth]{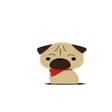} &
\includegraphics[width=0.125\columnwidth]{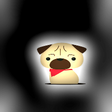} &
\includegraphics[width=0.125\columnwidth]{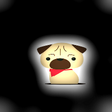} &
\includegraphics[width=0.125\columnwidth]{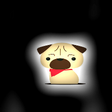} &
\includegraphics[width=0.125\columnwidth]{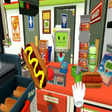} &
\includegraphics[width=0.125\columnwidth]{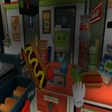} &
\includegraphics[width=0.125\columnwidth]{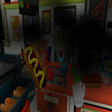} &
\includegraphics[width=0.125\columnwidth]{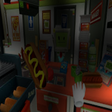} \\

\includegraphics[width=0.125\columnwidth]{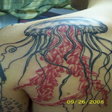} &
\includegraphics[width=0.125\columnwidth]{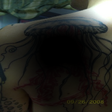} &
\includegraphics[width=0.125\columnwidth]{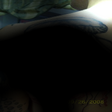} &
\includegraphics[width=0.125\columnwidth]{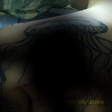} &
\includegraphics[width=0.125\columnwidth]{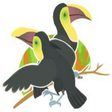} &
\includegraphics[width=0.125\columnwidth]{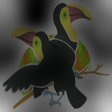} &
\includegraphics[width=0.125\columnwidth]{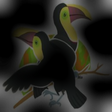} &
\includegraphics[width=0.125\columnwidth]{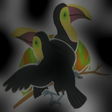} \\

\includegraphics[width=0.125\columnwidth]{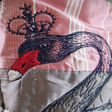} &
\includegraphics[width=0.125\columnwidth]{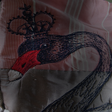} &
\includegraphics[width=0.125\columnwidth]{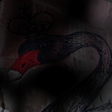} &
\includegraphics[width=0.125\columnwidth]{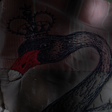} &
\includegraphics[width=0.125\columnwidth]{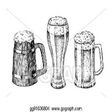} &
\includegraphics[width=0.125\columnwidth]{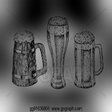} &
\includegraphics[width=0.125\columnwidth]{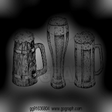} &
\includegraphics[width=0.125\columnwidth]{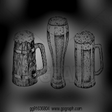} \\

\includegraphics[width=0.125\columnwidth]{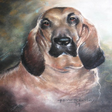} &
\includegraphics[width=0.125\columnwidth]{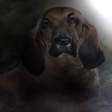} &
\includegraphics[width=0.125\columnwidth]{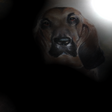} &
\includegraphics[width=0.125\columnwidth]{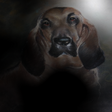} &
\includegraphics[width=0.125\columnwidth]{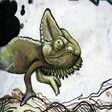} &
\includegraphics[width=0.125\columnwidth]{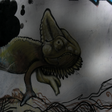} &
\includegraphics[width=0.125\columnwidth]{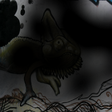} &
\includegraphics[width=0.125\columnwidth]{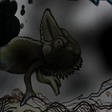} \\

\includegraphics[width=0.125\columnwidth]{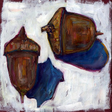} &
\includegraphics[width=0.125\columnwidth]{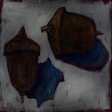} &
\includegraphics[width=0.125\columnwidth]{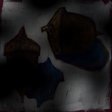} &
\includegraphics[width=0.125\columnwidth]{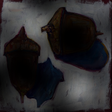} &
\includegraphics[width=0.125\columnwidth]{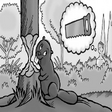} &
\includegraphics[width=0.125\columnwidth]{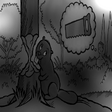} &
\includegraphics[width=0.125\columnwidth]{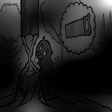} &
\includegraphics[width=0.125\columnwidth]{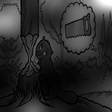} \\

\includegraphics[width=0.125\columnwidth]{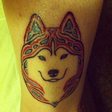} &
\includegraphics[width=0.125\columnwidth]{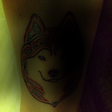} &
\includegraphics[width=0.125\columnwidth]{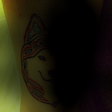} &
\includegraphics[width=0.125\columnwidth]{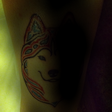} &
\includegraphics[width=0.125\columnwidth]{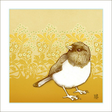} &
\includegraphics[width=0.125\columnwidth]{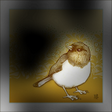} &
\includegraphics[width=0.125\columnwidth]{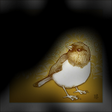} &
\includegraphics[width=0.125\columnwidth]{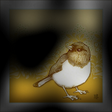} \\

Original & Baseline & PixelAT & PyramidAT & Original & Baseline & PixelAT & PyramidAT \\
\end{tabular}
\caption{Visualizations of the attention for different pre-trainings. Examples on dataset ImageNet-Rendition.}
\label{fig:attn_viz_examples_dataset_imagenet_r}
\end{figure}

\begin{figure}
\setlength{\tabcolsep}{.12em}
\begin{tabular}{cccccccc}

\includegraphics[width=0.125\columnwidth]{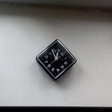} &
\includegraphics[width=0.125\columnwidth]{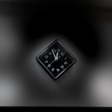} &
\includegraphics[width=0.125\columnwidth]{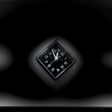} &
\includegraphics[width=0.125\columnwidth]{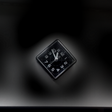} &
\includegraphics[width=0.125\columnwidth]{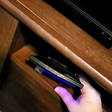} &
\includegraphics[width=0.125\columnwidth]{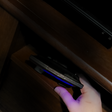} &
\includegraphics[width=0.125\columnwidth]{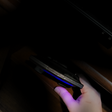} &
\includegraphics[width=0.125\columnwidth]{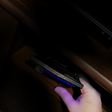} \\

\includegraphics[width=0.125\columnwidth]{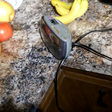} &
\includegraphics[width=0.125\columnwidth]{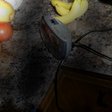} &
\includegraphics[width=0.125\columnwidth]{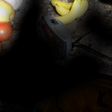} &
\includegraphics[width=0.125\columnwidth]{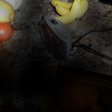} &
\includegraphics[width=0.125\columnwidth]{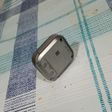} &
\includegraphics[width=0.125\columnwidth]{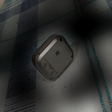} &
\includegraphics[width=0.125\columnwidth]{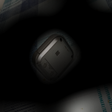} &
\includegraphics[width=0.125\columnwidth]{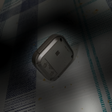} \\

\includegraphics[width=0.125\columnwidth]{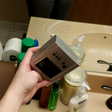} &
\includegraphics[width=0.125\columnwidth]{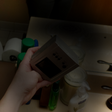} &
\includegraphics[width=0.125\columnwidth]{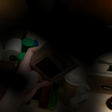} &
\includegraphics[width=0.125\columnwidth]{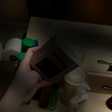} &
\includegraphics[width=0.125\columnwidth]{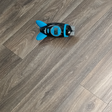} &
\includegraphics[width=0.125\columnwidth]{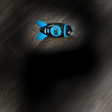} &
\includegraphics[width=0.125\columnwidth]{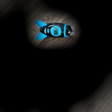} &
\includegraphics[width=0.125\columnwidth]{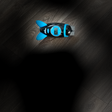} \\

\includegraphics[width=0.125\columnwidth]{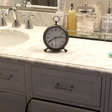} &
\includegraphics[width=0.125\columnwidth]{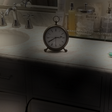} &
\includegraphics[width=0.125\columnwidth]{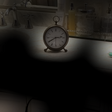} &
\includegraphics[width=0.125\columnwidth]{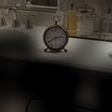} &
\includegraphics[width=0.125\columnwidth]{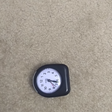} &
\includegraphics[width=0.125\columnwidth]{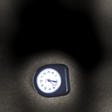} &
\includegraphics[width=0.125\columnwidth]{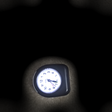} &
\includegraphics[width=0.125\columnwidth]{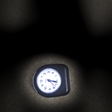} \\

\includegraphics[width=0.125\columnwidth]{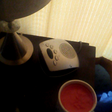} &
\includegraphics[width=0.125\columnwidth]{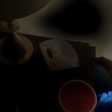} &
\includegraphics[width=0.125\columnwidth]{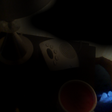} &
\includegraphics[width=0.125\columnwidth]{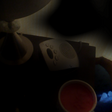} &
\includegraphics[width=0.125\columnwidth]{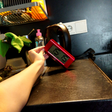} &
\includegraphics[width=0.125\columnwidth]{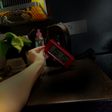} &
\includegraphics[width=0.125\columnwidth]{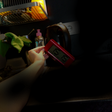} &
\includegraphics[width=0.125\columnwidth]{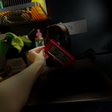} \\

\includegraphics[width=0.125\columnwidth]{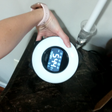} &
\includegraphics[width=0.125\columnwidth]{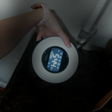} &
\includegraphics[width=0.125\columnwidth]{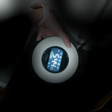} &
\includegraphics[width=0.125\columnwidth]{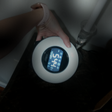} &
\includegraphics[width=0.125\columnwidth]{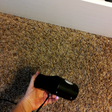} &
\includegraphics[width=0.125\columnwidth]{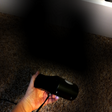} &
\includegraphics[width=0.125\columnwidth]{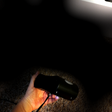} &
\includegraphics[width=0.125\columnwidth]{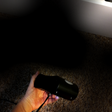} \\

\includegraphics[width=0.125\columnwidth]{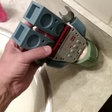} &
\includegraphics[width=0.125\columnwidth]{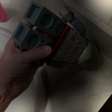} &
\includegraphics[width=0.125\columnwidth]{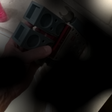} &
\includegraphics[width=0.125\columnwidth]{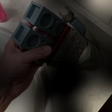} &
\includegraphics[width=0.125\columnwidth]{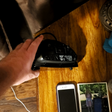} &
\includegraphics[width=0.125\columnwidth]{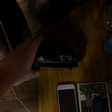} &
\includegraphics[width=0.125\columnwidth]{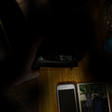} &
\includegraphics[width=0.125\columnwidth]{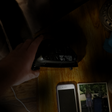} \\

\includegraphics[width=0.125\columnwidth]{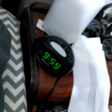} &
\includegraphics[width=0.125\columnwidth]{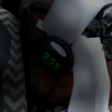} &
\includegraphics[width=0.125\columnwidth]{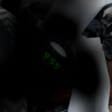} &
\includegraphics[width=0.125\columnwidth]{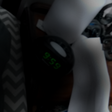} &
\includegraphics[width=0.125\columnwidth]{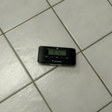} &
\includegraphics[width=0.125\columnwidth]{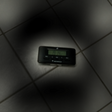} &
\includegraphics[width=0.125\columnwidth]{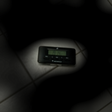} &
\includegraphics[width=0.125\columnwidth]{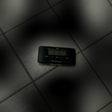} \\

Original & Baseline & PixelAT & PyramidAT & Original & Baseline & PixelAT & PyramidAT \\
\end{tabular}
\caption{Visualizations of the attention for different pre-trainings. Examples on dataset ObjectNet.}
\label{fig:attn_viz_examples_dataset_objectnet}
\end{figure}

\begin{figure}
\setlength{\tabcolsep}{.12em}
\begin{tabular}{cccccccc}

\includegraphics[width=0.125\columnwidth]{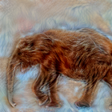} &
\includegraphics[width=0.125\columnwidth]{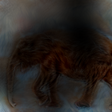} &
\includegraphics[width=0.125\columnwidth]{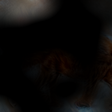} &
\includegraphics[width=0.125\columnwidth]{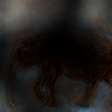} &
\includegraphics[width=0.125\columnwidth]{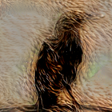} &
\includegraphics[width=0.125\columnwidth]{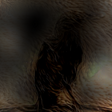} &
\includegraphics[width=0.125\columnwidth]{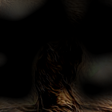} &
\includegraphics[width=0.125\columnwidth]{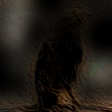} \\

\includegraphics[width=0.125\columnwidth]{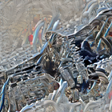} &
\includegraphics[width=0.125\columnwidth]{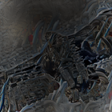} &
\includegraphics[width=0.125\columnwidth]{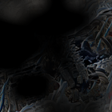} &
\includegraphics[width=0.125\columnwidth]{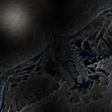} &
\includegraphics[width=0.125\columnwidth]{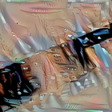} &
\includegraphics[width=0.125\columnwidth]{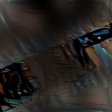} &
\includegraphics[width=0.125\columnwidth]{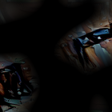} &
\includegraphics[width=0.125\columnwidth]{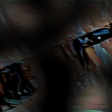} \\

\includegraphics[width=0.125\columnwidth]{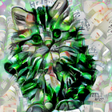} &
\includegraphics[width=0.125\columnwidth]{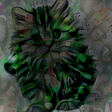} &
\includegraphics[width=0.125\columnwidth]{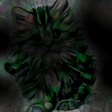} &
\includegraphics[width=0.125\columnwidth]{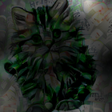} &
\includegraphics[width=0.125\columnwidth]{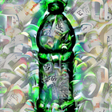} &
\includegraphics[width=0.125\columnwidth]{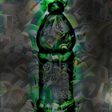} &
\includegraphics[width=0.125\columnwidth]{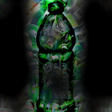} &
\includegraphics[width=0.125\columnwidth]{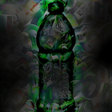} \\

\includegraphics[width=0.125\columnwidth]{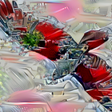} &
\includegraphics[width=0.125\columnwidth]{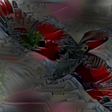} &
\includegraphics[width=0.125\columnwidth]{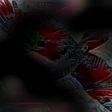} &
\includegraphics[width=0.125\columnwidth]{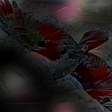} &
\includegraphics[width=0.125\columnwidth]{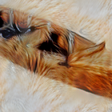} &
\includegraphics[width=0.125\columnwidth]{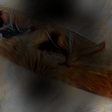} &
\includegraphics[width=0.125\columnwidth]{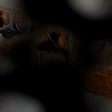} &
\includegraphics[width=0.125\columnwidth]{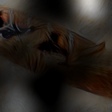} \\

\includegraphics[width=0.125\columnwidth]{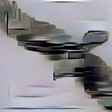} &
\includegraphics[width=0.125\columnwidth]{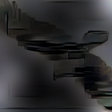} &
\includegraphics[width=0.125\columnwidth]{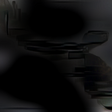} &
\includegraphics[width=0.125\columnwidth]{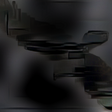} &
\includegraphics[width=0.125\columnwidth]{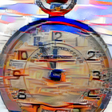} &
\includegraphics[width=0.125\columnwidth]{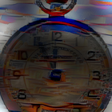} &
\includegraphics[width=0.125\columnwidth]{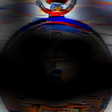} &
\includegraphics[width=0.125\columnwidth]{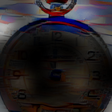} \\

\includegraphics[width=0.125\columnwidth]{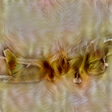} &
\includegraphics[width=0.125\columnwidth]{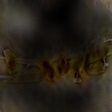} &
\includegraphics[width=0.125\columnwidth]{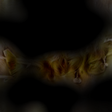} &
\includegraphics[width=0.125\columnwidth]{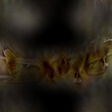} &
\includegraphics[width=0.125\columnwidth]{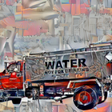} &
\includegraphics[width=0.125\columnwidth]{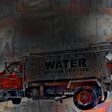} &
\includegraphics[width=0.125\columnwidth]{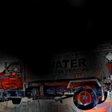} &
\includegraphics[width=0.125\columnwidth]{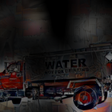} \\

\includegraphics[width=0.125\columnwidth]{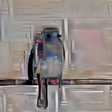} &
\includegraphics[width=0.125\columnwidth]{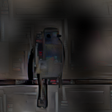} &
\includegraphics[width=0.125\columnwidth]{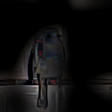} &
\includegraphics[width=0.125\columnwidth]{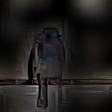} &
\includegraphics[width=0.125\columnwidth]{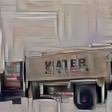} &
\includegraphics[width=0.125\columnwidth]{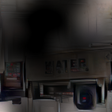} &
\includegraphics[width=0.125\columnwidth]{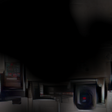} &
\includegraphics[width=0.125\columnwidth]{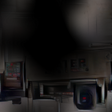} \\

\includegraphics[width=0.125\columnwidth]{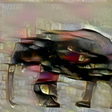} &
\includegraphics[width=0.125\columnwidth]{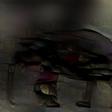} &
\includegraphics[width=0.125\columnwidth]{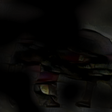} &
\includegraphics[width=0.125\columnwidth]{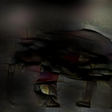} &
\includegraphics[width=0.125\columnwidth]{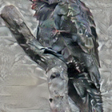} &
\includegraphics[width=0.125\columnwidth]{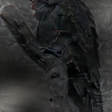} &
\includegraphics[width=0.125\columnwidth]{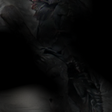} &
\includegraphics[width=0.125\columnwidth]{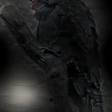} \\

Original & Baseline & PixelAT & PyramidAT & Original & Baseline & PixelAT & PyramidAT \\
\end{tabular}
\caption{Visualizations of the attention for different pre-trainings. Examples on dataset StylizedImageNet.}
\label{fig:attn_viz_examples_dataset_stylized_imagenet}
\end{figure}

We also include 32 examples of the attention for individual images sampled from the following datasets: ImageNet (Figure \ref{fig:attn_viz_examples_dataset_imagenet}), ImageNet-A (Figure \ref{fig:attn_viz_examples_dataset_imagenet_a}), ImageNet-ReaL (Figure \ref{fig:attn_viz_examples_dataset_imagenet2012_real}), ImageNet-Rendition  (Figure \ref{fig:attn_viz_examples_dataset_imagenet_r}), ObjectNet (Figure \ref{fig:attn_viz_examples_dataset_objectnet}), and StylizedImageNet (Figure \ref{fig:attn_viz_examples_dataset_stylized_imagenet}). The trend, as stated in the main paper, remains consistent through most of the examples. Baseline tends to be random and highlight both the object and background (particularly corners); PixelAT tries to aggressively crop to the object in the image, often cutting off parts of the object; and PyramidAT crops more closely than baseline but less aggressively than PixelAT. PyramidAT tends to take a more global perspective on the image and attends to both the object but also potentially relevant pieces of the background.

\section{Optimizers}
\label{sup:optimizers}
We observe different behavior from adversarial training depending on the optimizer used in generating the attacks; note, discussion of optimizers was omitted from the main paper due to concerns regarding space and complexity. Throughout the main paper, we use SGD, the standard optimizer in the adversarial attack and training community. However after testing multiple optimizers (Adam\cite{adam:2014}, AdaBelief\cite{zhuang2020adabelief}), we observe significantly different behavior from AdaBelief. Specifically, as shown in Table \ref{tab:adabelief}, AdaBelief provides a significant improvement to PixelAT ($0.71$ to ImageNet, $1.72$ in ImageNet-R) and a marginal improvement to PyramidAT ($0.08$ to ImageNet, $0.98$ in ImageNet-R).

\begin{table*}[!htbp]\centering\small
\begin{tabular}{l|cc|cccc|ccc}
\toprule
&\multicolumn{2}{c|}{}&\multicolumn{7}{c}{Out of Distribution Robustness Test} \\
 Method & ImageNet & Real & A & C$\downarrow$ & ObjectNet & V2 & Rendition & Sketch & Stylized \\
\midrule
 PixelAT SGD & 80.42 & 85.78 & 19.15 & 47.68 & 30.11 & 68.78 & 45.39 & 34.40 & 18.28 \\
 PixelAT AdaBelief & \textbf{81.13} & \textbf{86.40} & \textbf{21.45} & \textbf{45.25} & \textbf{32.03} & \textbf{69.97} & \textbf{47.11} & \textbf{36.18} & \textbf{20.55} \\
\midrule
 PyramidAT SGD & 81.71 & \textbf{86.82} & 22.99 & 44.99 & \textbf{32.92} & 70.82 & 47.66 & 36.77 & 19.14 \\
 PyramidAT AdaBelief & \textbf{81.79} & 86.79 & \textbf{23.24} & \textbf{44.79} & 32.81 & \textbf{70.87} & \textbf{48.64} & \textbf{38.38} & \textbf{20.78} \\
\end{tabular}
\caption{SGD vs AdaBelief}
\label{tab:adabelief}
\end{table*}

\begin{figure}\centering
\setlength{\tabcolsep}{.12em}
\begin{tabular}{ccccc}
\includegraphics[width=0.19\columnwidth]{SupplementalFigures/attack_viz/original_0.png} &
\includegraphics[width=0.19\columnwidth]{SupplementalFigures/attack_viz/pixel_attack_against_regvit_viz_baseline_0.png} &
\includegraphics[width=0.19\columnwidth]{SupplementalFigures/attack_viz/pixel_attack_against_regvit_viz_advprop_0.png} &
\includegraphics[width=0.19\columnwidth]{SupplementalFigures/attack_viz/pixel_attack_against_regvit_viz_pyramid_0.png} &
\includegraphics[width=0.19\columnwidth]{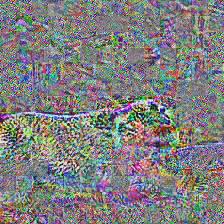}  \\
\includegraphics[width=0.19\columnwidth]{SupplementalFigures/attack_viz/original_1.png} &
\includegraphics[width=0.19\columnwidth]{SupplementalFigures/attack_viz/pixel_attack_against_regvit_viz_baseline_1.png} &
\includegraphics[width=0.19\columnwidth]{SupplementalFigures/attack_viz/pixel_attack_against_regvit_viz_advprop_1.png} &
\includegraphics[width=0.19\columnwidth]{SupplementalFigures/attack_viz/pixel_attack_against_regvit_viz_pyramid_1.png} &
\includegraphics[width=0.19\columnwidth]{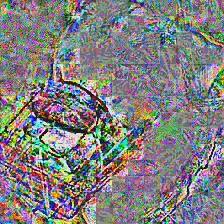}  \\
\includegraphics[width=0.19\columnwidth]{SupplementalFigures/attack_viz/original_2.png} &
\includegraphics[width=0.19\columnwidth]{SupplementalFigures/attack_viz/pixel_attack_against_regvit_viz_baseline_2.png} &
\includegraphics[width=0.19\columnwidth]{SupplementalFigures/attack_viz/pixel_attack_against_regvit_viz_advprop_2.png} &
\includegraphics[width=0.19\columnwidth]{SupplementalFigures/attack_viz/pixel_attack_against_regvit_viz_pyramid_2.png} &
\includegraphics[width=0.19\columnwidth]{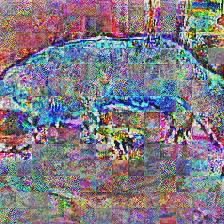}  \\
\includegraphics[width=0.19\columnwidth]{SupplementalFigures/attack_viz/original_3.png} &
\includegraphics[width=0.19\columnwidth]{SupplementalFigures/attack_viz/pixel_attack_against_regvit_viz_baseline_3.png} &
\includegraphics[width=0.19\columnwidth]{SupplementalFigures/attack_viz/pixel_attack_against_regvit_viz_advprop_3.png} &
\includegraphics[width=0.19\columnwidth]{SupplementalFigures/attack_viz/pixel_attack_against_regvit_viz_pyramid_3.png} &
\includegraphics[width=0.19\columnwidth]{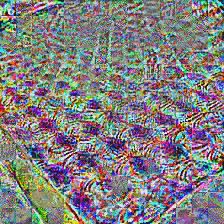}  \\
Original & Baseline & Pixel & Pyramid & Pixel AdaBelief \\
\end{tabular}
\caption{Visualizations of  pixel attacks using SGD on different pre-trainings: baseline, pixel, pyramid, and pixel Adabelief.}
\label{fig:adabelief}
\end{figure}

As shown in Figure~\ref{fig:adabelief}, we also observe significant visual difference in the pixel attacks on the pixel-trained model with AdaBelief.

\begin{figure}\centering
\setlength{\tabcolsep}{.12em}
\begin{tabular}{ccccc}
\includegraphics[width=0.19\columnwidth]{SupplementalFigures/attack_viz/original_0.png} &
\includegraphics[width=0.19\columnwidth]{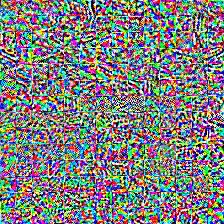} &
\includegraphics[width=0.19\columnwidth]{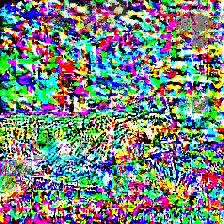} &
\includegraphics[width=0.19\columnwidth]{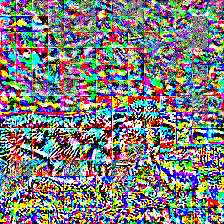} &
\includegraphics[width=0.19\columnwidth]{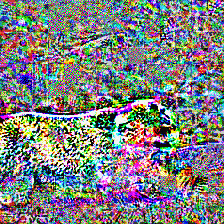}  \\
\includegraphics[width=0.19\columnwidth]{SupplementalFigures/attack_viz/original_1.png} &
\includegraphics[width=0.19\columnwidth]{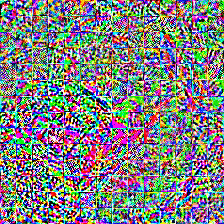} &
\includegraphics[width=0.19\columnwidth]{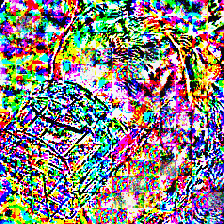} &
\includegraphics[width=0.19\columnwidth]{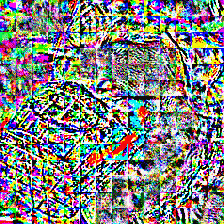} &
\includegraphics[width=0.19\columnwidth]{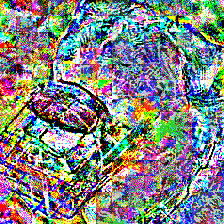}  \\
\includegraphics[width=0.19\columnwidth]{SupplementalFigures/attack_viz/original_2.png} &
\includegraphics[width=0.19\columnwidth]{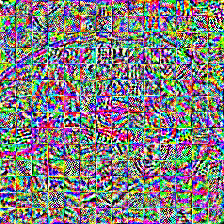} &
\includegraphics[width=0.19\columnwidth]{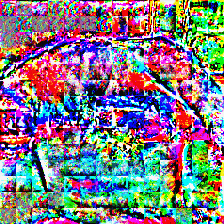} &
\includegraphics[width=0.19\columnwidth]{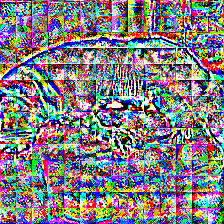} &
\includegraphics[width=0.19\columnwidth]{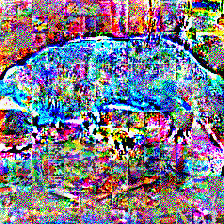}  \\
\includegraphics[width=0.19\columnwidth]{SupplementalFigures/attack_viz/original_3.png} &
\includegraphics[width=0.19\columnwidth]{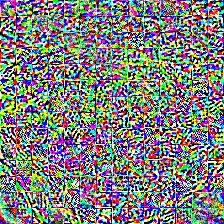} &
\includegraphics[width=0.19\columnwidth]{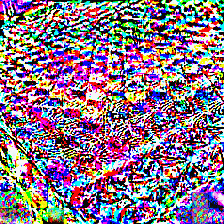} &
\includegraphics[width=0.19\columnwidth]{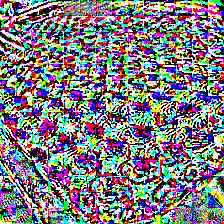} &
\includegraphics[width=0.19\columnwidth]{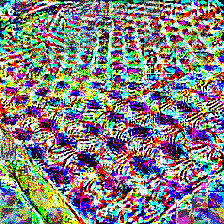}  \\
Original & Baseline & PixelAT SGD & PyramidAT SGD & PixelAT AdaBelief \\
\end{tabular}
\caption{Visualizations of  pixel attacks using AdaBelief on different pre-trainings: baseline, PixelAT SGD, PyramidAT SGD, and PixelAT Adabelief.}
\label{fig:adabelief_on_adabelief}
\end{figure}

Shown in Figure \ref{fig:adabelief_on_adabelief}, this visual difference is more apparent when looking at  pixel attacks using AdaBelief on these four different pre-trainings. In the pixel attacks using AdaBelief on AdaBelief pixel-trained model, contours and edges are clearly visible and the edits to the texture are smoother and more consistent. Even beyond classification, this may provide a way to do semi-supervised segmentation (with only the class label). Currently, AdaBelief does not provide such visible changes or improvements to pyramid. We leave this adaptation to future work.

\clearpage
\end{document}